\documentclass{article}

\usepackage{microtype}
\usepackage{graphicx}
\usepackage{subfigure}
\usepackage{booktabs} 

\usepackage{hyperref}

\usepackage[accepted]{icml2025}

\usepackage{amsmath}
\usepackage{amssymb}
\usepackage{mathtools}
\usepackage{amsthm}

\usepackage[capitalize,noabbrev]{cleveref}

\theoremstyle{plain}

\theoremstyle{definition}

\theoremstyle{remark}

\usepackage[textsize=tiny]{todonotes}

\newenvironment{allintypewriter}{\ttfamily}{\par}

\icmltitlerunning{Evaluating LLM Systematic Reasoning through Graph Coloring}

\begin{document}

\twocolumn[
\icmltitle{Evaluating the Systematic Reasoning Abilities \\ of Large Language Models through Graph Coloring}



\icmlsetsymbol{equal}{*}

\begin{icmlauthorlist}
\icmlauthor{Alex Heyman}{affil1}
\icmlauthor{Joel Zylberberg}{affil2,affil3}
\end{icmlauthorlist}

\icmlaffiliation{affil1}{Department of Electrical Engineering and Computer Science, York University, Toronto, ON, Canada}
\icmlaffiliation{affil2}{Jules Stein Eye Institute, University of California, Los Angeles, CA, USA}
\icmlaffiliation{affil3}{Learning in Machines and Brains Program, CIFAR, Toronto, ON, Canada}

\icmlcorrespondingauthor{Alex Heyman}{aheyman@yorku.ca}

\icmlkeywords{Machine Learning, Deep Learning, Large Language Model, Large Language Model Reasoning, Large Language Model Benchmarking}

\vskip 0.3in
]



\printAffiliationsAndNotice{} 

\begin{abstract}
Contemporary large language models are powerful problem-solving tools, but they exhibit weaknesses in their reasoning abilities which ongoing research seeks to mitigate. We investigate graph coloring as a means of evaluating an LLM's capacities for systematic step-by-step reasoning and possibility space exploration, as well as effects of semantic problem framing. We test Claude 3.5 Sonnet, Llama 3.1 405B, Gemini 1.5 Pro, GPT-4o, o1-mini, and DeepSeek-R1 on a dataset of $k$-coloring problems with $2 \leq k \leq 4$ and vertex count $4 \leq n \leq 8$, using partial algorithmic solvers to further categorize problems by difficulty. In addition to substantial but varying framing effects, we find that all models except o1-mini and R1 exhibit $>60\%$ error rates on difficult problem types in all frames ($>15\%$ for o1-mini and $>10\%$ for R1), and no model achieves perfect accuracy even in the simple domain of 2-coloring 4-vertex graphs. Our results highlight both the considerable recent progress in LLM systematic reasoning and the limits of its reliability, especially in relation to increasing computational costs. We expect that more complex graph coloring problems, and procedural generation of arbitrary-complexity reasoning problems more broadly, offer further untapped potential for LLM benchmarking.
\end{abstract}

\section{Introduction}

Contemporary large language models (LLMs) function as problem-solving machines by receiving a problem statement as a prompt and autoregressively generating a solution, using approaches learned from large-scale next-token-prediction pretraining typically followed by supervised fine-tuning (SFT) on smaller well-curated datasets and reinforcement learning from human feedback (RLHF). LLMs are capable of generating appropriate solutions to a wide variety of problems, including those that do not appear in their training datasets, indicating that their learned problem-solving approaches go beyond simple memorization and sparking interest in what has been conceptualized as LLM reasoning. As LLM parameter counts and training dataset sizes have grown by orders of magnitude over the past several years, their demonstrated reasoning abilities have grown as well \cite{naveed2024comprehensive}.

However, scientific study of LLM reasoning has accumulated evidence that it lacks the human-like robustness suggested by LLMs' anthropomorphic linguistic fluency. Of primary interest here is that LLMs seem to struggle with step-by-step compositional reasoning; they can sharply fail to generalize to problems of higher complexity than those seen in training \cite{dziri2024faith} and to facts not involved in appropriate reasoning examples seen in training \cite{wang2024grokked}, and they can be distracted by the inclusion of irrelevant information in problem statements, suggesting fragility in their approaches to composing facts \cite{shi2023large, mirzadeh2024gsm}. Even when a model gives a correct answer to a problem of higher complexity than it was trained on, its explanation of its own reasoning steps may contain errors, indicating that the explanation is a confabulation and the model's true approach is shallower \cite{dziri2024faith}. On tasks including but not limited to those involving step-by-step reasoning, models can fail to reach perfect accuracy on simple instances even as they sometimes correctly solve very difficult ones; also, as models grow larger and more extensively fine-tuned, their failures more often take the form of confidently presented incorrect answers as opposed to declining to answer, a behavior which some developers have incentivized by penalizing model ``evasiveness" \cite{zhou2024larger}.

In the past several months, a new class of LLM has emerged that integrates reinforcement learning into the training procedure in some capacity, with an eye toward teaching the model better step-by-step reasoning via autoregressive chains of thought; these ``large reasoning models" (LRMs) include OpenAI's o1 \cite{openai2024o1}, DeepSeek-R1 \cite{deepseek2025r1}, and MoonshotAI's Kimi k1.5 \cite{moonshotai2025kimik1.5}. First-party evaluation of LRMs has shown them decisively outperforming state-of-the-art standard LLMs on numerous reasoning benchmarks, but scientific evaluation and study of LRMs is overall in its infancy due to their novelty. Results so far suggest that LRMs possess standard LLM weaknesses such as unreliability on simple problems \cite{valmeekam2024llms} and out-of-distribution performance drops and distractibility \cite{mirzadeh2024gsm}, but to a lesser degree than standard LLMs. However, more research is needed to paint a clear picture of LRM reasoning's nature, capabilities, and limitations.

\subsection{Graph Coloring and LLM Reasoning}

\begin{figure*}[t]
\begin{centering}
\includegraphics[width=0.8\textwidth]{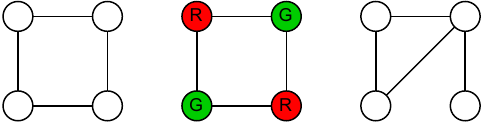}
\caption{\textbf{Left:} A simple undirected graph with 4 vertices and 4 edges. \textbf{Center:} A valid 2-coloring of the graph. \textbf{Right:} A different 4-vertex, 4-edge graph that cannot be 2-colored.}
\label{fig:graph_example}
\end{centering}
\end{figure*}

Graph coloring is the notion in graph theory of assigning labels (``colors") to elements of a graph subject to certain constraints. Here, we are concerned with a specific simple and well-known form of it: coloring the vertices of an undirected graph such that no two adjacent vertices (sharing an edge) also share a color. Whether or not a given graph can be colored with at most $k$ colors (``$k$-colored" for short) depends on the graph's precise connectivity and is computationally an NP-complete problem. (See Figure \ref{fig:graph_example} for a simple example.) The same is true of determining the minimum $k$ required to color a given graph.

Some previous work on LLM reasoning such as \citet{stechly2024self} has explored graph coloring in a limited capacity as one reasoning task for LLMs among several. However, we believe graph coloring can act as an uncommonly powerful tool for LLM reasoning evaluation, in the following ways:

\begin{itemize}
\item The space of possible graph coloring problems is large, with the number of possible graphs (counting vertex-order permutations separately) increasing superexponentially as vertex count increases. Sampling from the space of graphs to procedurally generate new graph coloring problems is computationally straightforward, and the complexity of the generated problems can be controlled via parameters such as number of vertices, edges, and available colors in the case of $k$-coloring for fixed $k$, limited only by the computational resources available. This makes graph coloring as an LLM evaluation tool robust for studying the effects of problem complexity on performance and gives it a way to resist interference from data contamination.
\item Reliably solving graph coloring problems requires systematic reasoning in the form of checking whether a candidate coloring satisfies each edge constraint, usually as part of the step-by-step compositional process of coloring vertices sequentially based on the coloring choices previously made for neighboring vertices. Beyond this, however, it is not always easy to find a valid $k$-coloring, show that one cannot exist, or find a graph's minimum $k$ ``on the first try"; solvers may need to go back on previously made coloring choices and systematically explore multiple possibilities to reach a correct answer. Recognizing one's own missteps and trying other options in response is a critical skill for robust reasoning agents, and graph coloring can evaluate LLMs on it.
\item Graphs are abstract mathematical entities that can be mapped onto many different real-world situations, and graph coloring problems can correspondingly be \textit{framed} in various terms other than those of graph coloring itself. Prior work has found that LLMs (like humans) change their performance on logic problems depending on the problems' semantic content apart from their abstract structure \cite{lampinen2024language}, and graph coloring problems repeated across multiple frames can be used to probe the magnitude and nature of an LLM's semantic biases to gain insight into its approaches to reasoning and their degree of robustness.
\end{itemize}

In this work, we generate a dataset of graph $k$-coloring problems with a range of (relatively small) vertex counts and values of $k$, categorized both by these parameters and by algorithmic measures of expected difficulty, and test four recent standard LLMs and two LRMs on these problems repeated across two to four semantic frames. We find framing effects that are substantial in all models but vary in nature between models, as well as generally higher error rates in more difficult problem categories as expected, with all four standard LLMs exceeding $60\%$ error for all frames in some categories. LRM error rates are much lower, but still increase drastically in more difficult categories, exceeding $10$-$15\%$ error at worst. Furthermore, no model (LRM or otherwise) fully reaches $0\%$ error for any combination of frame and vertex-count category. Our results reinforce existing evidence that LRMs decisively outperform standard LLMs at reasoning tasks, but also that they share the weaknesses of standard LLMs to some degree -- in this case, particularly in possibility space exploration.

\section{Methods}

\subsection{Problems}

In each of the problems in our dataset, the model is prompted with a description of an undirected graph with a specified number of vertices $n$ and set of edges that each connect two distinct vertices. The model is told to color each vertex with one of a set of $k$ colors so that no two adjacent vertices share a color, or state that this is impossible if it is. (More details on our prompts are given in Sections \ref{frames} and \ref{prompting}.)

The dataset is divided into five problem sets with varying $n$ and $k$: one with 4-vertex graphs that must be 2-colored (``4v2c" for short), followed in increasing complexity by 5v2c, 6v3c, 7v3c, and 8v4c. 4v2c consists of all $2^{C(4, 2)} = 64$ possible 4-vertex graphs, minus the graph with no edges. Similarly, 5v2c consists of all $2^{C(5, 2)} - 1 = 1023$ non-edgeless 5-vertex graphs. For 6v3c, 7v3c, and 8v4c, for each number of edges for which more than 50 $n$-vertex graphs exist, we randomly sample only 50 of them for the problem set; this yields sets of 631, 943, and 1307 problems respectively. Each of the five problem sets contains some graphs that are possible to $k$-color and some that are impossible. Note that unlike \citet{stechly2024self}, not all of our graphs are \textit{planar}; that is, it is not necessarily possible to draw the graph on a 2D plane such that no edges cross. This means that although our graphs have fewer vertices than theirs, some of ours are more densely connected, and while all planar graphs are 4-colorable, not all of our graphs are.

\subsection{Difficulty Categorization} \label{methods_difficulty}

Within each problem set, we categorize problems into different types based on algorithmic measures of difficulty. Problems where the graph is possible to color are categorized by how feasible they are to solve just by ``greedily" coloring one vertex at a time, without needing to explore multiple possibilities for any vertices' colors. We subject each problem to 10,000 trials of a simple randomized greedy algorithm that gives up if it reaches a contradiction (see Appendix \ref{greedy_score} for details), and the empirical success rate is called the problem's ``greedy score" $g$. We divide the problems into three types: $g \geq 0.9$, $0.5 \leq g < 0.9$, and $g < 0.5$. Our problem sets with higher $n$ and $k$ tend to contain proportionally more colorable problems with low $g$. (Note that a sufficiently careful greedy algorithm can solve 2-coloring problems without fail, but the algorithm we use merely fails \textit{infrequently} on colorable 4v2c and 5v2c problems.)

Meanwhile, uncolorable problems are categorized by whether or not they contain a complete subgraph of size $k + 1$ (that is, a subset of $k + 1$ vertices that each share edges with all of the others). Problems that contain such a subgraph are called ``complete-uncolorable" (C-uncolorable for short), and problems that do not are called ``diffuse-uncolorable" (D-uncolorable); the latter's uncolorability is expected to be less obvious. In our problem sets, almost all uncolorable problems are C-uncolorable; only 5v2c and 7v3c contain meaningful numbers of D-uncolorable problems (which in 5v2c's case are graphs with a 5-cycle but no 3-cycles).

\subsection{Frames} \label{frames}

We generate prompts based on the problems in our dataset using four different semantic frames (see Appendix \ref{example_prompts} for examples):

\begin{itemize}
\item The ``Math" frame presents the problem explicitly in terms of a graph with vertices and edges, where the vertices must be colored with $k$ colors so that no two adjacent vertices receive the same color. This frame is based on the graph coloring prompt from \citet{arkoudas2023gpt}.
\item The ``Cities" frame presents the problem in terms of a network of cities connected by highways; a monument to one of $k$ people must be built in each city so that no two cities directly connected by a highway have a monument to the same person. This frame presents the problem spatially, but not explicitly mathematically.
\item The ``Friends" frame presents the problem in terms of a social network of friendships, where everyone wears a shirt with one of $k$ colors and no one wants to wear the same color shirt as any of their friends. In this frame, the graph structure is neither explicit nor spatial.
\item The ``Math (demanding)" frame is identical to the Math frame except for a small difference in prompt phrasing. While the other three frames \textit{ask} the model whether satisfying the constraints is possible, Math (demanding) \textit{tells} the model to specify a valid coloring or to say that this is impossible. We include this frame as an example test of the effects of small prompt perturbations -- see e.g. \citet{salinas2024butterfly} -- in the context of graph coloring. We hypothesize that, if and when the phrasing difference from the Math frame has an effect, it will be to bias the model away from ``disappointingly" answering that the graph is uncolorable.
\end{itemize}

\subsection{Models}

We test the following six models on our problems:

\begin{itemize}
\item Meta's Llama 3.1 405B Instruct \cite{meta2024llama3.1}, hosted on Fireworks AI.
\item OpenAI's GPT-4o \cite{openai2024gpt4o}, specifically \texttt{gpt-4o-2024-08-06}, via the OpenAI API.
\item Anthropic's Claude 3.5 Sonnet \cite{anthropic2024claude3.5s}, specifically \texttt{claude-3-5-sonnet-20241022}, via the Anthropic API.
\item Google's Gemini 1.5 Pro \cite{google2024gemini1.5} via the Gemini Developer API.
\item OpenAI's LRM o1-mini \cite{openai2024o1mini}, specifically \texttt{o1-mini-2024-09-12}, via the OpenAI API. We chose o1-mini as the o1 version to test for cost-effectiveness reasons and because OpenAI reports that its mathematical reasoning performance is superior to o1-preview and comparable to the full o1.
\item DeepSeek's LRM R1 \cite{deepseek2025r1}, hosted on Fireworks AI.
\end{itemize}

Due to the high monetary costs of using o1-mini and of using DeepSeek-R1 through Fireworks AI shortly after its release, we only test the two LRMs on the Math and Friends frames, rather than all four. Also, for all problem sets except 4v2c, we do not test the LRMs on problems with the lowest edge counts (low enough that all problems are colorable with $g \geq 0.9$) or the highest edge counts (high enough that all problems are uncolorable). This leaves 5v2c, 6v3c, 7v3c, and 8v4c with 792, 300, 450, and 500 problems respectively in the ``interesting" region with medium edge counts. Finally, for 5v2c only, we randomly sample half of the problems from each ``interesting" edge count (3 through 6 edges) to bring the number of selected problems down to 396.

\subsection{Prompting and Response Evaluation} \label{prompting}

We prompt each of our models with each problem in our dataset presented in each frame (with certain problems and frames excluded for o1-mini and DeepSeek-R1 as described above), repeated five times each to account for model stochasticity and increase statistical power. We use a temperature of 0 for all models to minimize stochasticity for this low-creativity deductive reasoning task -- except for o1-mini, which only supports a temperature of 1, and DeepSeek-R1, whose developers recommend a temperature no lower than 0.5 \cite{deepseek2025r1modelcard}, which is what we use. We do not find endlessly repeating output to be a significant problem for any of our models at these temperatures.

Aside from setting model temperature, we do not use any techniques to guide or assist the models' approaches to our problems, such as non-default system prompts, in-context learning, or a tree-of-thought framework as in \citet{yao2024tree}. Our goal here is not to build a maximally effective LLM-based system for solving graph coloring problems, but to test the intelligence of LLMs apart from our intelligence as their human operators; a robust general-purpose AI reasoner must be able to autonomously solve problems even when its operators do not already know effective approaches to them.

Our evaluation of models' responses to our prompts is lenient; they are not penalized for not using the exact example answer format in the prompt, or for ignoring vertices with no adjoining edges. Any responses in which our automatic parser does not detect a coherent answer are reviewed and interpreted manually.

\subsection{Code and Data}

Code for reproducing our experiments and records of our problems, prompts, and responses can be found at \url{https://github.com/AlexHeyman/LLMGraphColoring}.

\section{Results}

\begin{figure*}[t]
\begin{centering}
\includegraphics[width=\textwidth]{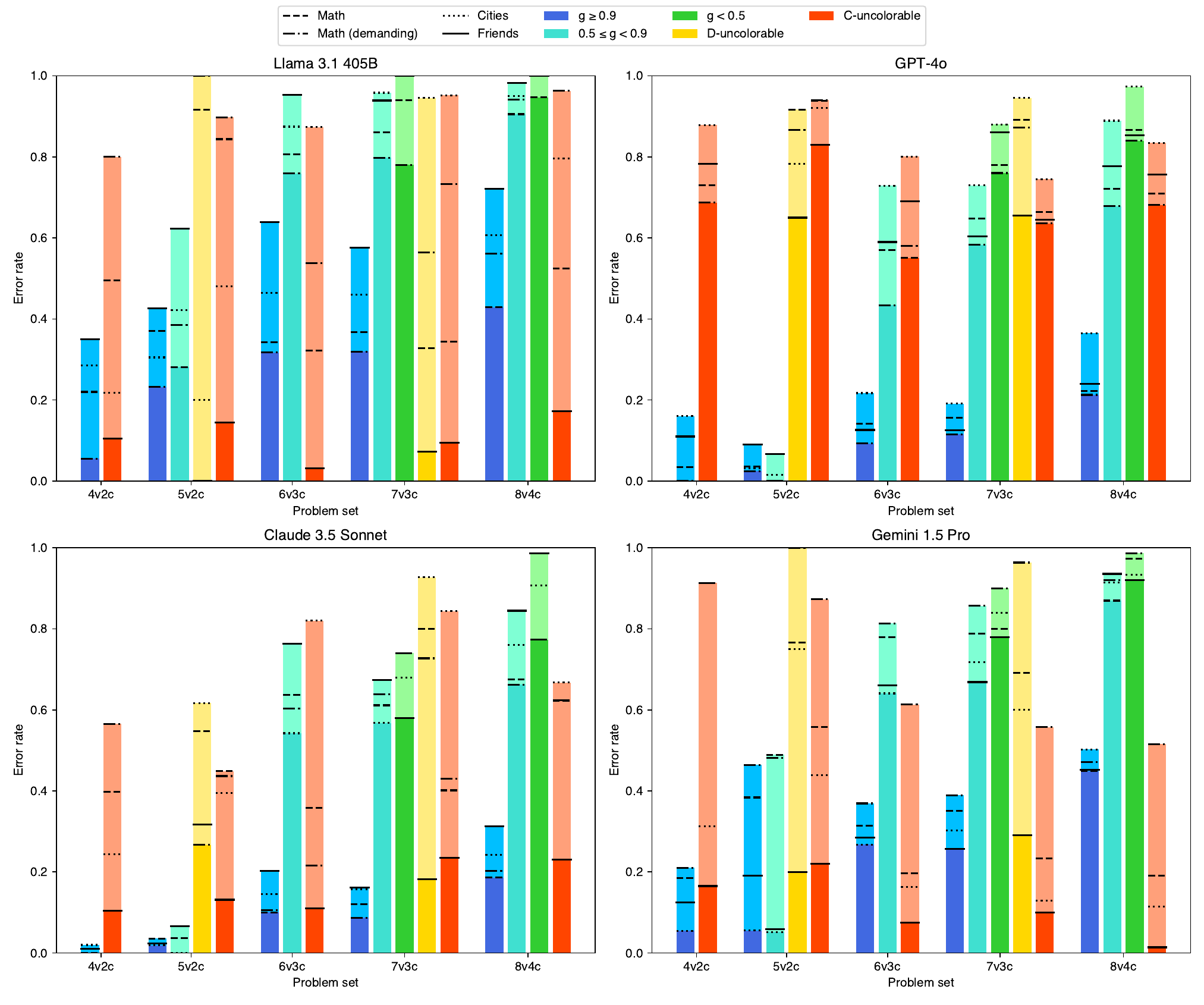}
\caption{Error rate for each standard LLM, problem set, problem type, and frame. The lighter-colored segment of each bar is the space between the lowest-error and highest-error frames.}
\label{fig:all_complete_models}
\end{centering}
\end{figure*}

\begin{figure*}[t]
\begin{centering}
\includegraphics[width=\textwidth]{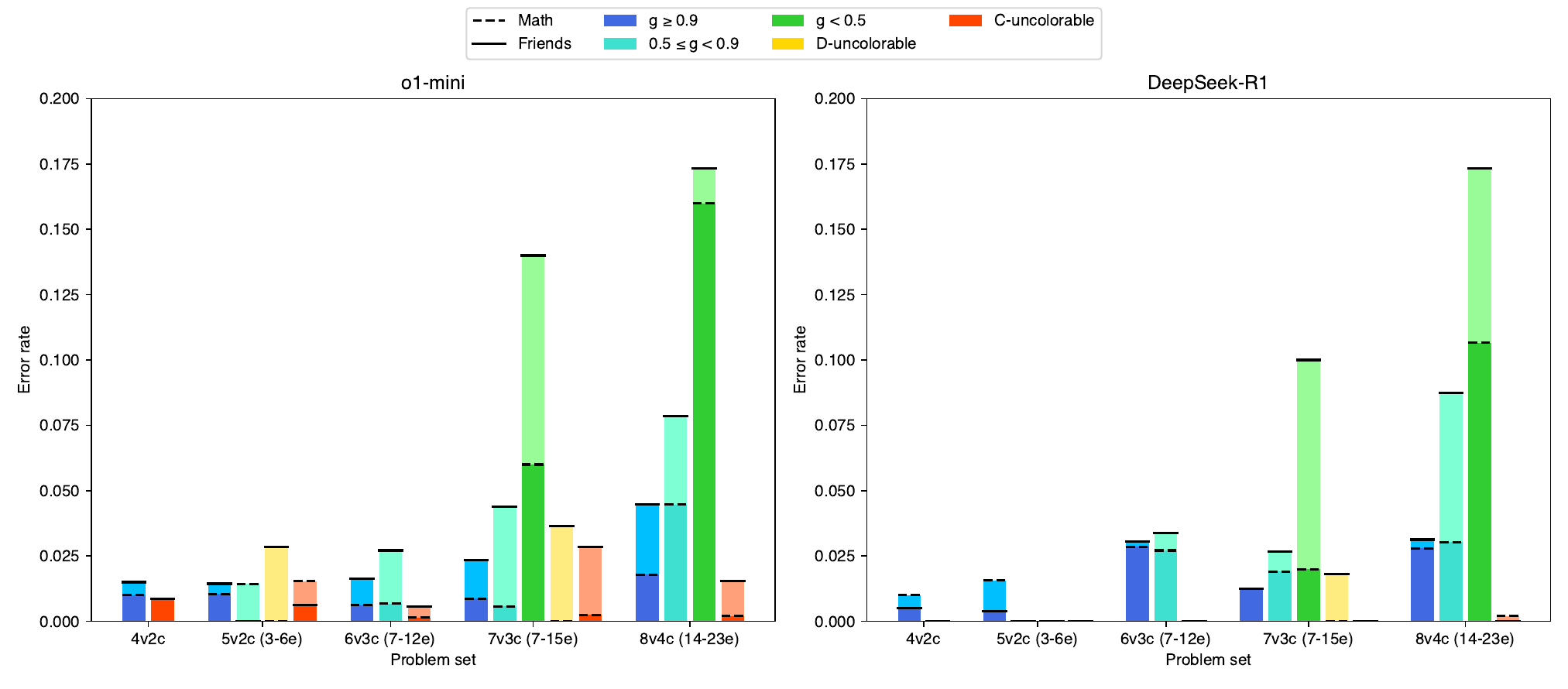}
\caption{Error rate for each LRM, edge-selected problem set, problem type, and frame. The lighter-colored segment of each bar is the space between the errors of the two tested frames. The y-axis is compressed in comparison to Figure \ref{fig:all_complete_models} to better show the differences between the smaller error rates.}
\label{fig:all_sel_models}
\end{centering}
\end{figure*}

\subsection{Standard LLMs}

Plots of the rates of incorrect answers from the four standard LLMs on each combination of problem set, problem type, and frame can be seen in Figure \ref{fig:all_complete_models}. A less concise but more detailed set of error rate plots can be seen in Appendix \ref{type_error_standard}, and plots of model accuracy vs. edge count can be seen in Appendix \ref{ec_accuracy_standard}. Note that 6v3c contains two D-uncolorable problems and 8v4c contains one, but these two categories are excluded from plots split by problem type due to low statistical power.

For all models, error rates trend upward with increasing problem set complexity -- at least for colorable problem types; uncolorable types show no consistent trend. As expected, error rates on colorable problems tend to increase as greedy scores decrease, though the increase from the $g \geq 0.9$ category to the $0.5 \leq g < 0.9$ category is much larger than the increase from the latter to $g < 0.5$. This is puzzling if we model the LLMs as approaching problems greedily but otherwise accurately, which suggests that some other form of difficulty increase correlated with $g$ substantially below 1 contributes to the error rate increase. Error rates can reach quite high; every model exceeds $60\%$ error on 8v4c $0.5 \leq g < 0.9$ and $g < 0.5$, $50\%$ error on 7v3c $0.5 \leq g < 0.9$ and $g < 0.5$, and $40\%$ error on 6v3c $0.5 \leq g < 0.9$, in all four frames. Additionally, no combination of model and frame reaches $0\%$ error on the combined set of 4v2c problems despite the simplicity of the domain.

These patterns can be seen from another angle in the aforementioned accuracy vs. edge count plots in Appendix \ref{ec_accuracy_standard}. For a given combination of model, problem set, and frame, as edge count increases from minimum to maximum, accuracy on problems of the given edge count typically starts at or near $100\%$ and then begins decreasing, with the start of the accuracy decrease either co-occurring with or preceding the start of the decrease in the proportion of $g \geq 0.9$ problems. As edge count approaches maximum and all problems become uncolorable, this decrease may completely reverse itself, continue until accuracy is near $0\%$, or anywhere in between. The minimum accuracy for any edge count trends downward as problem set complexity increases, always falling below $50\%$ on 7v3c and $30\%$ on 8v4c.

While patterns like these hold in general across models, the models also exhibit substantial differences; for instance, our plots split by problem type and by edge count both show Claude 3.5 Sonnet as overall the best-performing model despite the situationally poor performance it shares with the others. Additionally, the effects of frame choice are considerable for all models but vary in the specifics. For instance, for Llama and Claude, the Friends frame is below average at correctly solving colorable problems but above average at correctly identifying uncolorable problems as uncolorable; for GPT-4o, Friends is about average overall on both kinds of problems; and for Gemini, Friends is broadly above average. Cities is broadly below average for Llama and GPT-4o but broadly average for Gemini, and for Llama and Claude it seems to particularly struggle with uncolorable problems in 6v3c, 7v3c, and 8v4c. When Math (demanding) diverges significantly from normal Math, it tends to perform worse than Math particularly on uncolorable problems -- supporting our hypothesis that it biases models away from answering negatively -- although exceptions exist, like Claude on 6v3c.

Interestingly, we found that in very rare cases, Llama 3.1 405B outputted responses refusing to give a definitive answer, sometimes after trying to color the graph and reaching a contradiction, and usually invoking the problem's complexity to justify the refusal. This did not occur with any of the other models; it occurred almost exclusively in the Math frame with a few instances in Math (demanding) as well; and it occurred mainly in responses for 7v3c (accounting for about $1.5\%$ of 7v3c Math responses) and occasionally also 6v3c and 8v4c. We treated these responses as ``neither correct nor incorrect", not counting them negatively toward error rate nor positively toward accuracy; however, they were rare enough that this made little practical difference.

\subsection{LRMs -- o1-mini and DeepSeek-R1}

Plots of error rates for the two LRMs on each combination of edge-selected problem set, problem type, and frame can be seen in Figure \ref{fig:all_sel_models}. A less concise but more detailed set of error rate plots can be seen in Appendix \ref{type_error_lrm}, and plots of model accuracy vs. edge count can be seen in Appendix \ref{ec_accuracy_lrm}.

Error rates for the LRMs are much lower overall than for the standard LLMs; while every standard LLM exceeds $60\%$ error on certain 8v4c problem types in all frames, neither of the LRMs ever reaches $20\%$ error in either of their frames. As with the standard LLMs, LRM error rates for colorable problems increase as greedy scores decrease -- but unlike the standard LLMs, error rates spike upward drastically from the $0.5 \leq g < 0.9$ category to the $g < 0.5$ category, with 8v4c Math jumping from $\sim 4.5\%$ to $\sim 16\%$ for o1-mini and $\sim 3\%$ to $\sim 10.5\%$ for DeepSeek-R1. Meanwhile, error rates for uncolorable problems are generally low for o1-mini, and almost zero for DeepSeek-R1 with the exception of 7v3c D-uncolorable Friends. For both LRMs, error rates for Friends are generally higher than those for Math; this may reflect the prominence of explicitly mathematical reasoning problems in their training. Finally, like the standard LLMs, no combination of LRM and frame quite reaches $0\%$ error even on 4v2c (o1-mini Math had 3 out of $63 \times 5 = 315$ responses incorrect, o1-mini Friends had 4, DeepSeek-R1 Math had 2, and DeepSeek-R1 Friends had 1).

\section{Discussion}

In this work, we investigated graph coloring as a tool for evaluation of LLM reasoning, specifically systematic compositional reasoning, possibility space exploration, and semantic frame effects. We generated a dataset of relatively small-scale graph $k$-coloring problems, presented in four different frames, and tested four recent standard LLMs and two large reasoning models (LRMs) on it.

We found that, while the standard LLMs varied in performance to some extent, they all exhibited distinctly imperfect performance on even the simplest problems and struggled to find valid colorings in problems where they exist but are non-obvious. The LRMs erred much less frequently, but still fell short of perfect accuracy on simple problems and showed drastic increases in error rate as problems became more complex and valid colorings became less feasible to find with a greedy strategy.

Our results align with prior work finding that standard LLMs lack robust compositional reasoning abilities, and reinforce that LRMs represent a considerable performance improvement in this domain. However, our results also contribute to the body of examples of LRMs sharing standard LLM weaknesses to some degree; most distinctively, our finding that $k$-coloring performance for LRMs (proportionally) drops from low greedy scores even more than it does for standard LLMs suggests that LRMs have not improved on standard LLMs at possibility space exploration as much as they have at linear step-by-step reasoning.

Since the range of problem complexity levels that we test is fairly restricted, it is unclear from our results whether LRM performance on reasoning problems like ours ever exhibits a rapid drop toward chance-level performance as complexity increases ala \citet{dziri2024faith}, or whether performance declines more gradually. If an LRM exhibits a rapid drop at a complexity level where all solution steps can be performed well within the LRM's memory constraints, that would suggest that its approach to the problems is improperly shallow in a similar way to the standard LLM reasoning that \citet{dziri2024faith} studies. This question about LRM reasoning would be straightforward to investigate in future work, however, owing to the ease of generating graph coloring problems of any desired complexity within a wide range.

We found that the several semantic frames we tested induced differences in both standard LLM and LRM performance, and that these differences exhibited perceptible patterns -- albeit not universal patterns, especially across different models. Even the small phrasing change between the Math and Math (demanding) frames led to differing performance in some situations, in an overall pattern matching what we hypothesized from the semantics of the phrasing change. This aligns with prior work finding that LLM performance responds to problem framing, and indicates that repeating LLM test problems across multiple frames and breaking the results down both by frame and by other problem features has potential as a method of probing LLM semantic biases in detail.

Our finding that none of the LLMs we tested achieved $0\%$ error even on 4-vertex 2-coloring problems has concerning implications for their \textit{reliability}; as \citet{zhou2024larger} might phrase it, when it comes to reasoning problems like ours, neither standard LLMs nor LRMs appear to have an \textit{operating range} of non-trivial complexity levels within which they can be trusted to provide a correct solution. This is particularly disappointing in LRMs' case given their reasoning-focused design priorities, and must be accounted for by anyone seeking to deploy LLMs to autonomously perform tasks involving compositional reasoning. Compounding this issue is the observation, made by \citet{zhou2024larger} and reinforced by our work, that large and thoroughly fine-tuned LLMs rarely decline to present a supposed solution to a problem even when the problem is of a sort that they are not likely to answer correctly; in other words, LLMs cannot be trusted to make their own limitations known to their operators. LLM developers must carefully consider whether they want their models to exhibit this behavior, and be transparent about their choices.

Finally, we wish to discuss the issue of the computational costs of training and deploying LLMs, and the implications of our findings for decision-making in this realm. Despite recent innovations in cost-efficiency from the likes of \citet{deepseek2025r1}, high-performing LLMs (including LRMs) remain massive neural networks with a minimum of tens of billions of parameters active at a time, and LRMs drive inference costs higher still to the extent that their behavior of autoregressively ``thinking" before responding leads them to take more total autoregressive steps for each response. Furthermore, OpenAI found that o1's accuracy on the American Invitational Mathematics Examination (AIME) benchmark increased only logarithmically with increasing test-time compute \cite{openai2024o1}, and DeepSeek similarly found that over the course of R1-Zero's training, its average response length grew linearly but its accuracy on AIME saw diminishing returns \cite{deepseek2025r1}. Given these observations, we advise LRM developers to resist the temptation to rely on ``scaling up" chain-of-thought length to combat the performance and reliability limitations that our work helps illuminate; we expect that it will not yield fundamental improvements in reasoning robustness, much as scaling up the architecture size of standard LLMs did not \cite{press2023measuring}, and neither will it yield large enough benefits in practice to make up for both the sustainability costs and the direct economic costs. We instead advise a focus on continuing to innovate in the design of architectures and training procedures.


\section*{Acknowledgements}

This work was supported by a Canada Research Chair Grant (CRC-2022-00277) to JZ.

\section*{Impact Statement}

Our goal in this work is to contribute to the project of accurately assessing the capabilities of large language models and using the resulting knowledge to inform both the deployment of existing LLMs and the development of future LLMs and other AI systems. We believe it is important for individuals and society to avoid overestimating the capabilities of AI systems and to avoid becoming reliant on them more than they merit. We also believe the AI research community has a collective responsibility to develop an accurate understanding of the behavior of the systems we develop and disseminate that understanding to those who use them. We hope to contribute to the fulfillment of that responsibility through this work.

To the extent that this work helps guide the development of future AI systems toward more powerful reasoning capabilities, it shares in the responsibility for the effects of such systems on the world. Since the potential effects of the use of powerful AI reasoners include dangerous and harmful ones alongside beneficial ones, we take this opportunity to remind readers of the importance of thorough safety testing and holistic ethical judgment prior to the deployment of any substantially autonomous AI system, and the moral responsibility that AI developers bear for the actions of the systems they design.

\bibliography{paper}
\bibliographystyle{icml2025}

\newpage
\appendix
\onecolumn
\section{Greedy Score Algorithm} \label{greedy_score}

The deliberately imperfect randomized greedy algorithm that we use to calculate the greedy score for each colorable problem in our dataset (see Section \ref{methods_difficulty}) proceeds as follows:

\begin{enumerate}
\item Initialize the graph with all vertices uncolored. Order the available colors arbitrarily from ``smallest" to ``largest".
\item Pick an uncolored vertex at random to be the ``current vertex" $c$. If this cannot be done because all vertices are colored, the algorithm ends in \textbf{success}.
\item Color $c$ with the smallest color that none of its colored neighbors share. If $c$ has a neighbor of every color, the algorithm ends in \textbf{failure}.
\item If $c$ has any uncolored neighbors, pick one of them at random to be the new $c$ and go to step 3. Otherwise, go to step 2.
\end{enumerate}

\section{Example Prompts} \label{example_prompts}

Following is one example prompt for each of our four frames, all based on the same problem from the 6v3c problem set.

\subsection{Math}

\noindent\fbox{
\parbox{\textwidth - 2\fboxsep}{
\begin{allintypewriter}
Consider an undirected graph with 6 vertices (numbered 0 through 5) and the following set of edges:\\

\{(0,4), (1,3), (1,4), (1,5), (2,4), (3,4), (4,5)\}\\

Suppose that we want to color every vertex either red, green, or blue so that no two adjacent vertices receive the same color. Is this possible? If it is impossible, write "Impossible" as the final line of your response. If it is possible, the final lines of your response should present a plan for it in a format like the following:\\

0 Red\\
1 Green\\
2 Blue\\
(etc.)
\end{allintypewriter}
}
}

\subsection{Math (demanding)}

\noindent\fbox{
\parbox{\textwidth - 2\fboxsep}{
\begin{allintypewriter}
Consider an undirected graph with 6 vertices (numbered 0 through 5) and the following set of edges:\\

\{(0,4), (1,3), (1,4), (1,5), (2,4), (3,4), (4,5)\}\\

Color every vertex either red, green, or blue so that no two adjacent vertices receive the same color, or if this is impossible, say so. If it is impossible, write "Impossible" as the final line of your response. If it is possible, the final lines of your response should present your vertex coloring in a format like the following:\\

0 Red\\
1 Green\\
2 Blue\\
(etc.)
\end{allintypewriter}
}
}

\subsection{Cities}

\noindent\fbox{
\parbox{\textwidth - 2\fboxsep}{
\begin{allintypewriter}
Suppose there is a country with 6 cities (numbered 1 through 6) and a highway system where each highway connects exactly two cities. The highways are: between city 1 and city 5, between city 2 and city 4, between city 2 and city 5, between city 2 and city 6, between city 3 and city 5, between city 4 and city 5, and between city 5 and city 6.\\

Suppose the government wants to erect a monument in each city. Each monument will be to either the first President, the first Vice President, or the first Secretary of State. The government wants it so, if any pair of cities are directly connected by a highway, the monuments in those cities cannot both be to the same person. Is this possible? If it is impossible, write "Impossible" as the final line of your response. If it is possible, the final lines of your response should present a plan for it in a format like the following:\\

1 President\\
2 VP\\
3 Secretary\\
(etc.)
\end{allintypewriter}
}
}

\subsection{Friends}

\noindent\fbox{
\parbox{\textwidth - 2\fboxsep}{
\begin{allintypewriter}
Imagine 6 people: Alice, Bob, Carol, Dave, Ethan, and Fran. Suppose that the following friendships exist between them: Alice is friends with Ethan, Bob is friends with Dave, Bob is friends with Ethan, Bob is friends with Fran, Carol is friends with Ethan, Dave is friends with Ethan, and Ethan is friends with Fran.\\

Suppose that the 6 people are all going to attend a party, and each of them is going to wear either a red shirt, a green shirt, or a blue shirt. Suppose that none of them want to wear the same color shirt as anyone they are friends with. Is this possible? If it is impossible, write "Impossible" as the final line of your response. If it is possible, the final lines of your response should present a plan for it in a format like the following:\\

Alice: Red\\
Bob: Green\\
Carol: Blue\\
(etc.)
\end{allintypewriter}
}
}

\section{Error Rates by Model, Problem Set, Frame, and Problem Type (Standard LLMs)} \label{type_error_standard}

Following are bar plots of error rates for each combination of standard LLM and problem set, split further by frame and problem type. These present the same information as Figure \ref{fig:all_complete_models} with additional detail. Error bars represent $95\%$ Clopper-Pearson binomial confidence intervals. For the three colorable problem types, a lighter-colored segment at the top of a bar represents trials where the model answered with an invalid coloring, while the normally colored bar segment represents trials where the model falsely declared the graph uncolorable. A black segment at the bottom of a bar represents trials where the model did not produce a coherent answer.

\begin{figure}[H]
\begin{subfigure}
\centering
\includegraphics[width=0.5\textwidth]{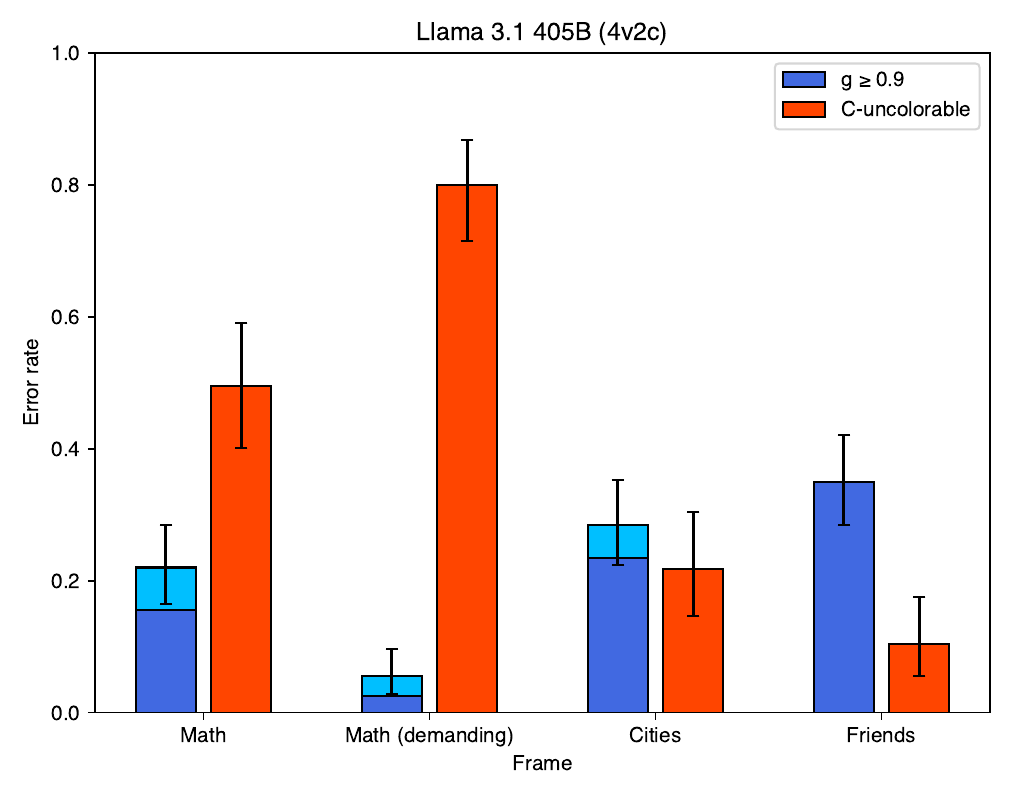}
\end{subfigure}
\hspace*{-0.9em}
\begin{subfigure}
\centering
\includegraphics[width=0.5\textwidth]{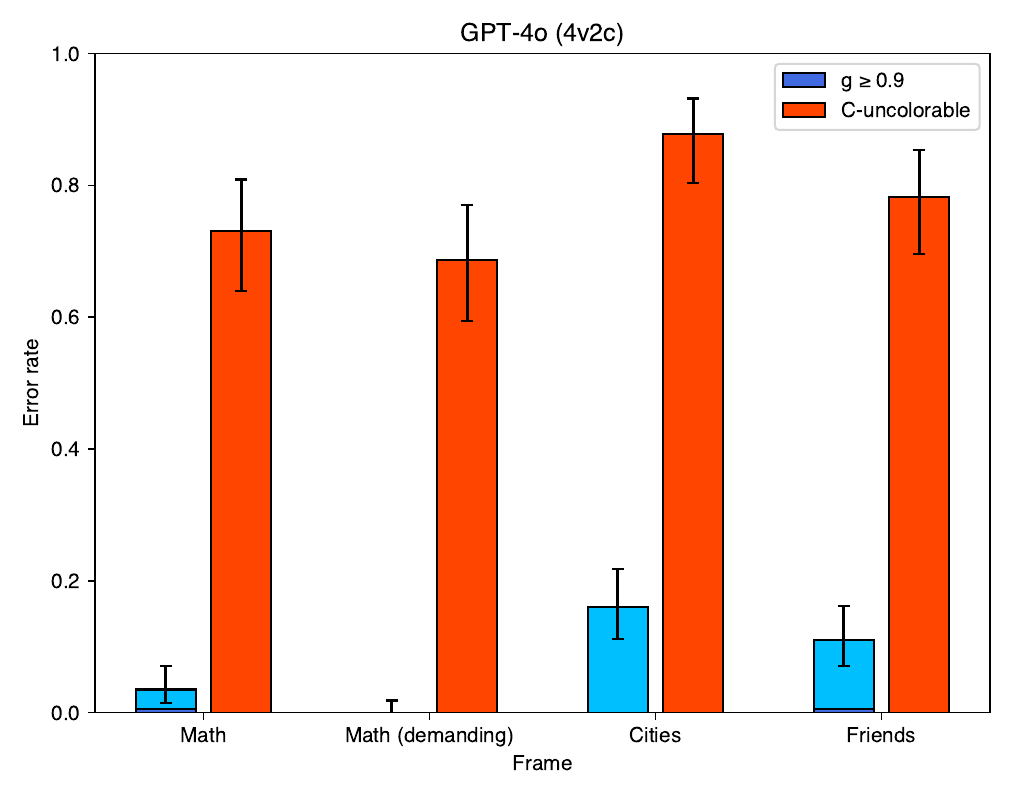}
\end{subfigure}
\end{figure}
\vspace{-0.45in}

\begin{figure}[H]
\begin{subfigure}
\centering
\includegraphics[width=0.5\textwidth]{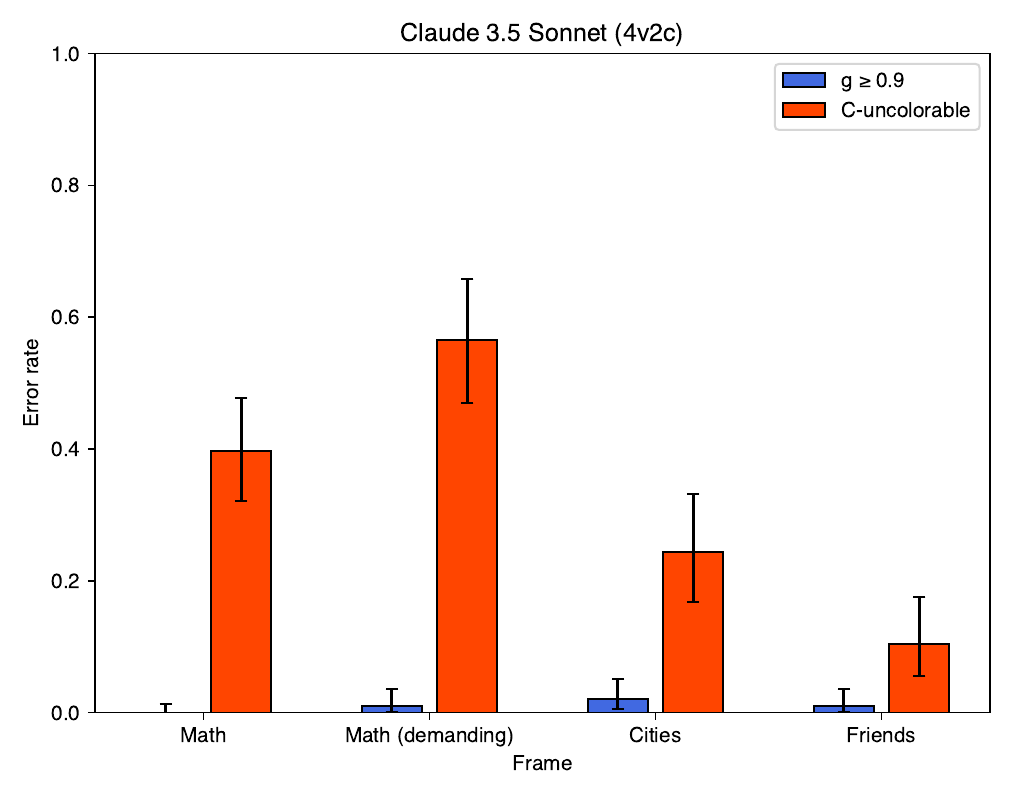}
\end{subfigure}
\hspace*{-0.9em}
\begin{subfigure}
\centering
\includegraphics[width=0.5\textwidth]{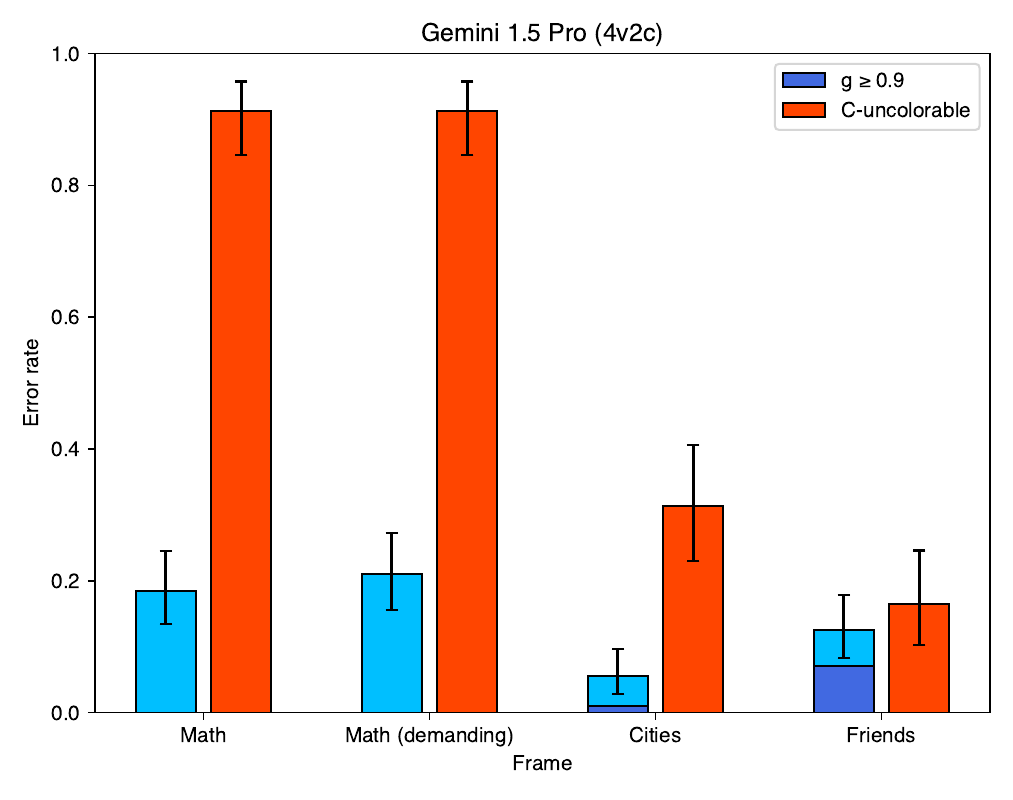}
\end{subfigure}
\end{figure}
\vspace{-0.45in}

\begin{figure}[H]
\begin{subfigure}
\centering
\includegraphics[width=0.5\textwidth]{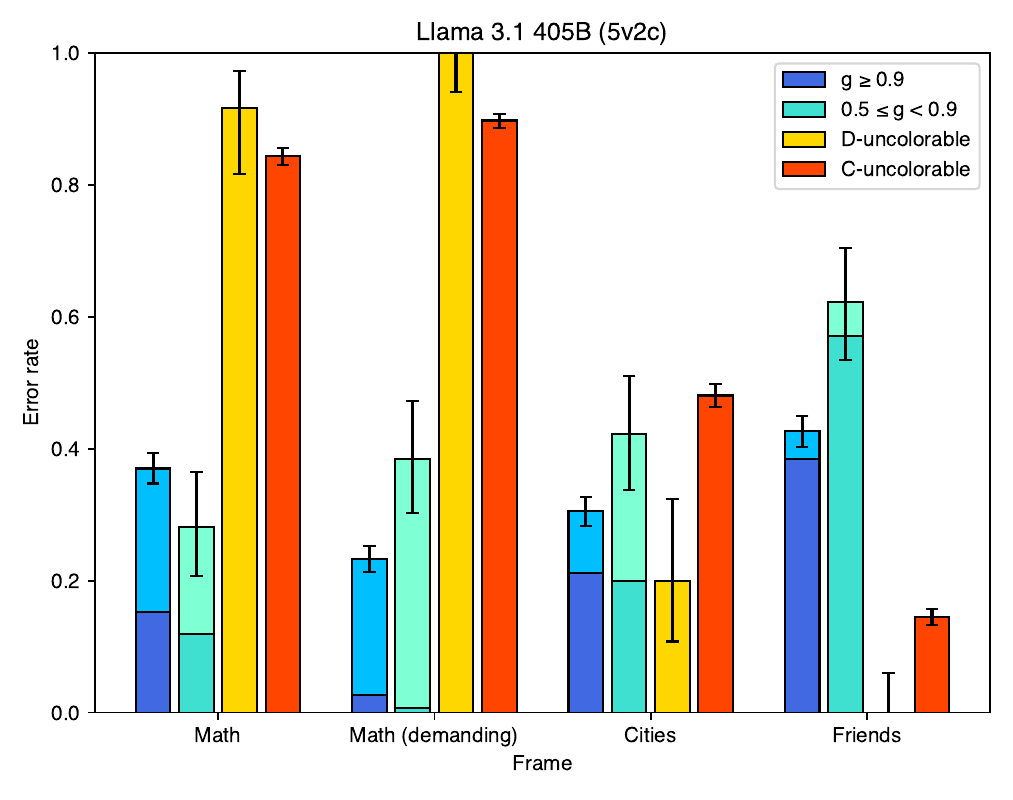}
\end{subfigure}
\hspace*{-0.9em}
\begin{subfigure}
\centering
\includegraphics[width=0.5\textwidth]{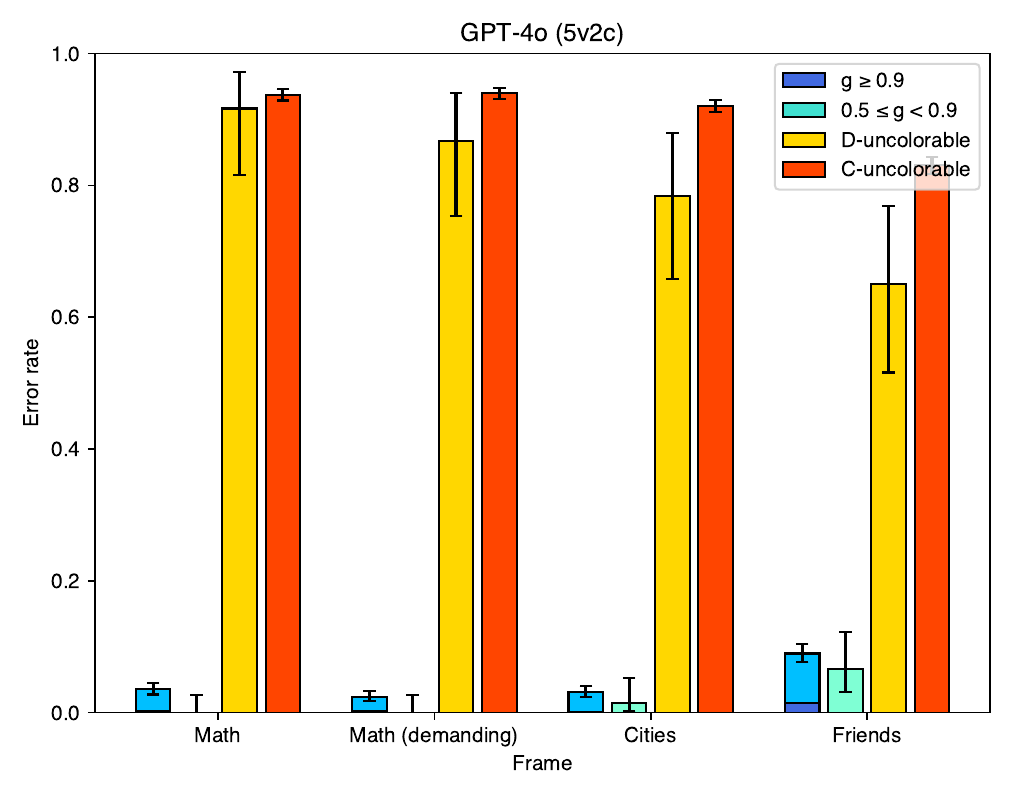}
\end{subfigure}
\end{figure}
\vspace{-0.45in}

\begin{figure}[H]
\begin{subfigure}
\centering
\includegraphics[width=0.5\textwidth]{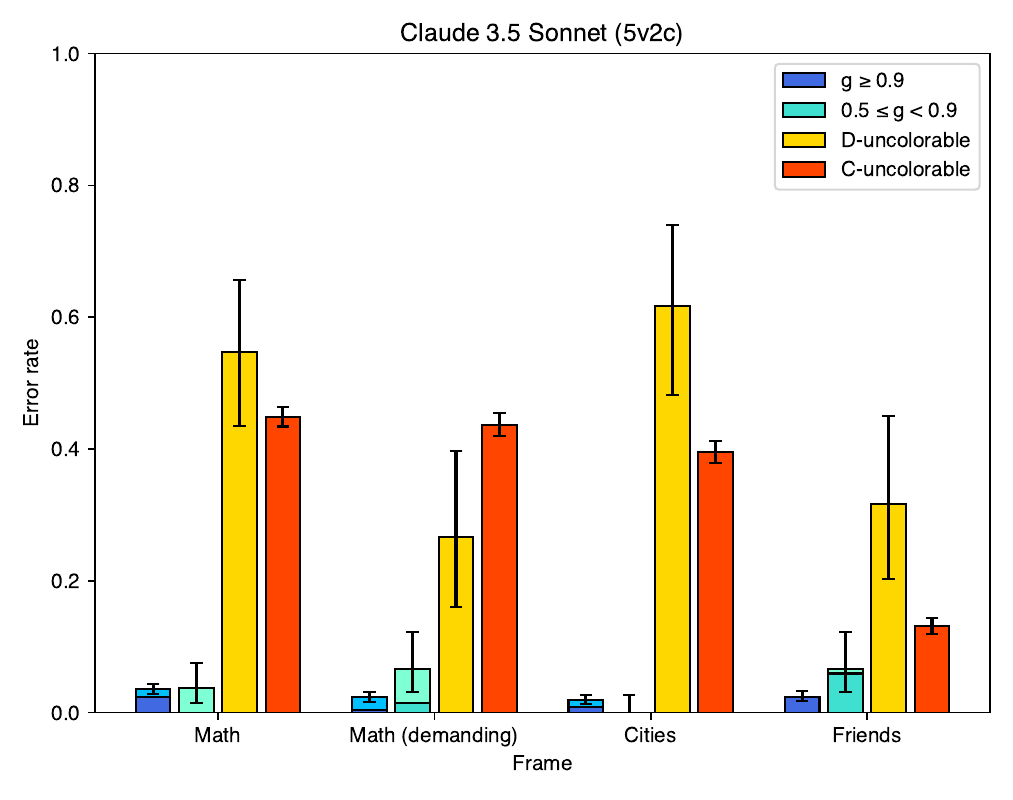}
\end{subfigure}
\hspace*{-0.9em}
\begin{subfigure}
\centering
\includegraphics[width=0.5\textwidth]{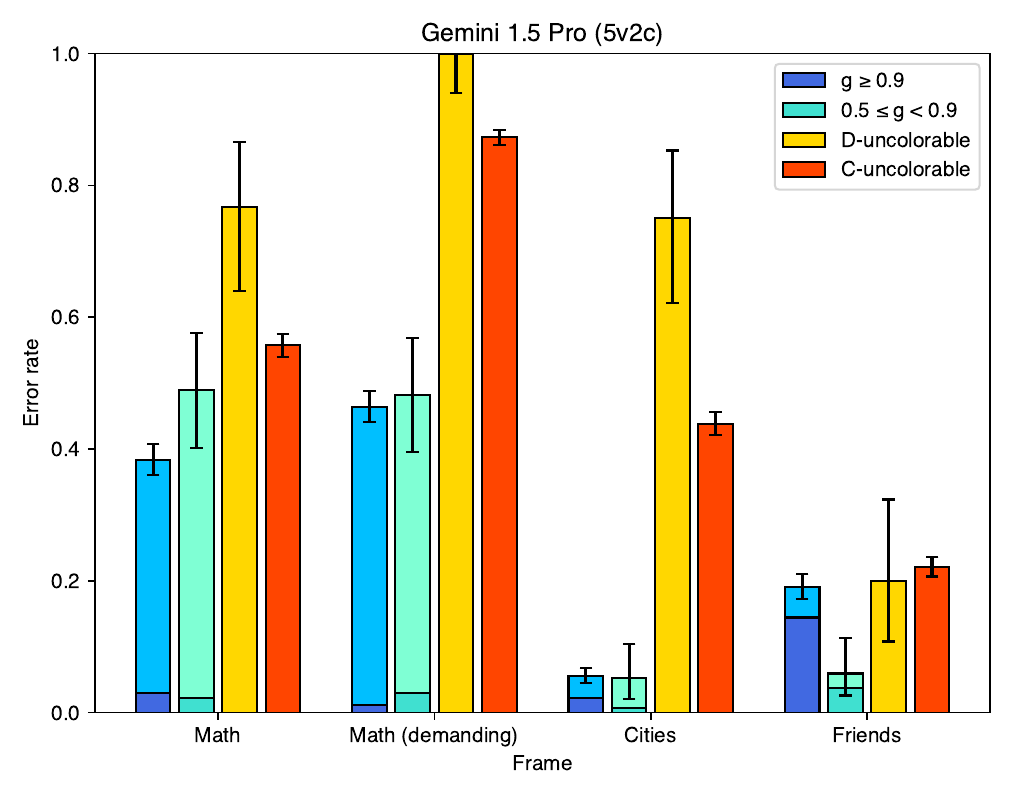}
\end{subfigure}
\end{figure}
\vspace{-0.45in}

\begin{figure}[H]
\begin{subfigure}
\centering
\includegraphics[width=0.5\textwidth]{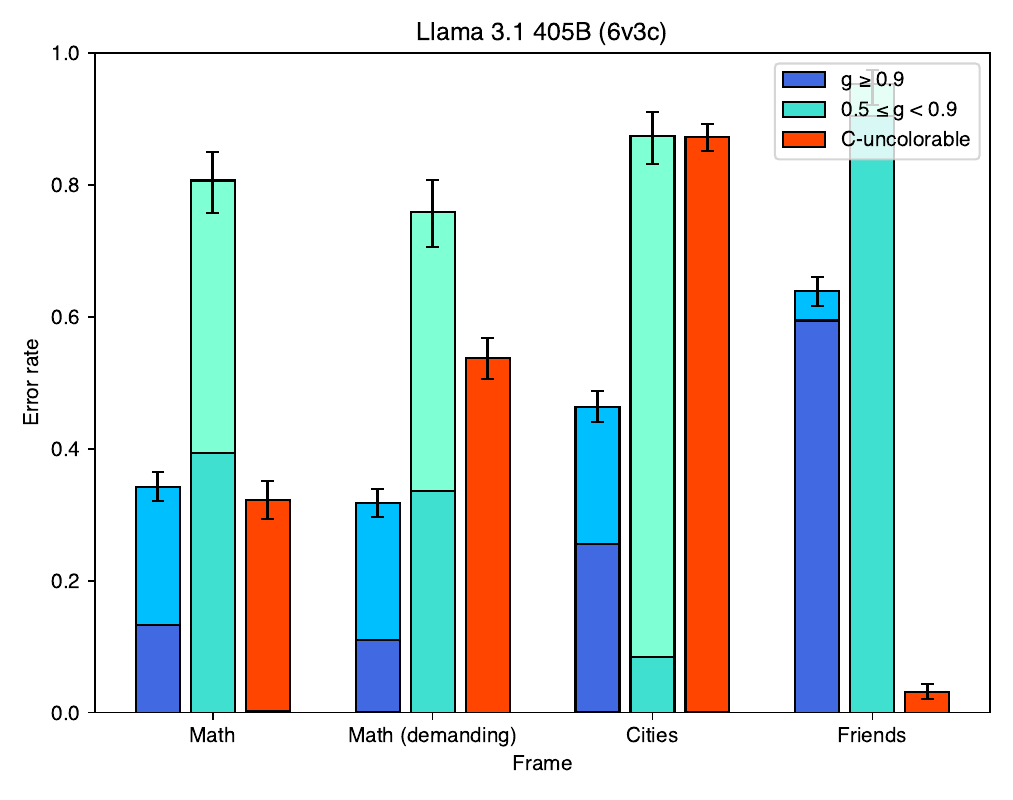}
\end{subfigure}
\hspace*{-0.9em}
\begin{subfigure}
\centering
\includegraphics[width=0.5\textwidth]{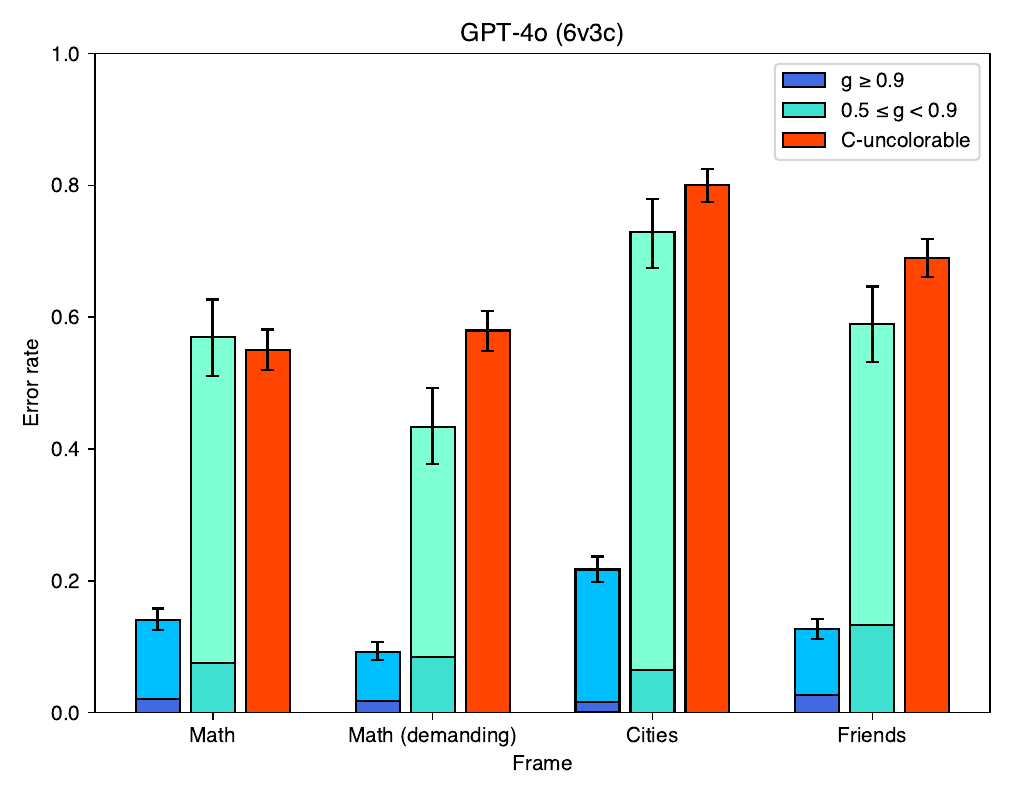}
\end{subfigure}
\end{figure}
\vspace{-0.45in}

\begin{figure}[H]
\begin{subfigure}
\centering
\includegraphics[width=0.5\textwidth]{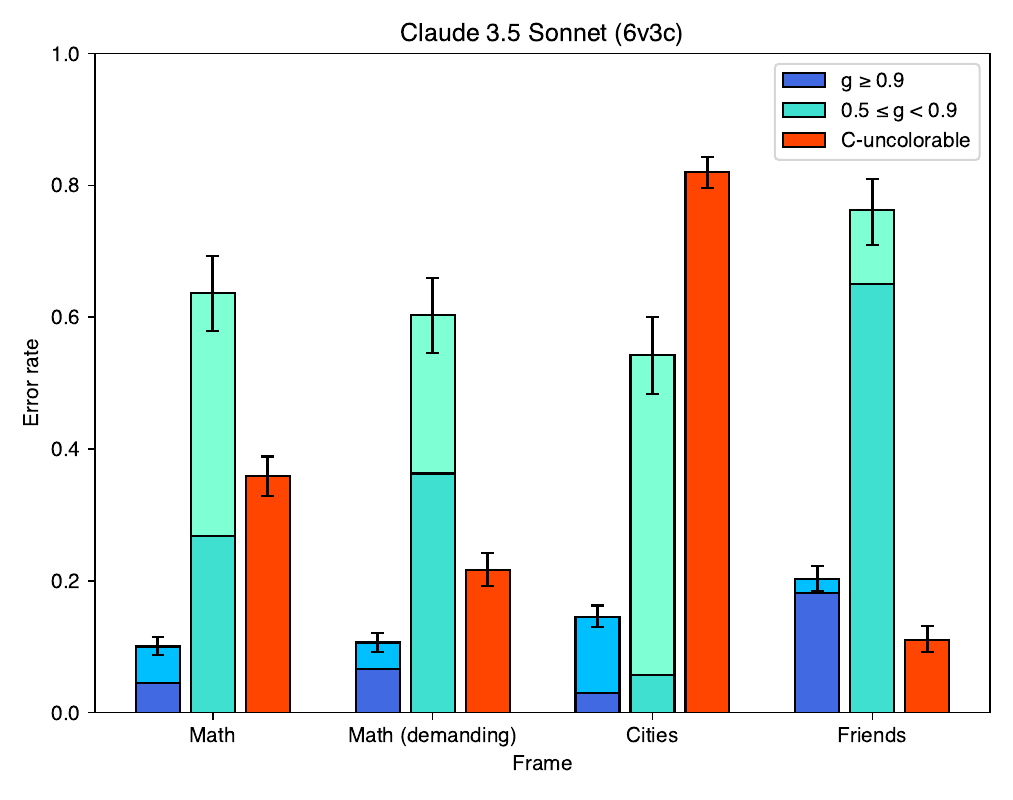}
\end{subfigure}
\hspace*{-0.9em}
\begin{subfigure}
\centering
\includegraphics[width=0.5\textwidth]{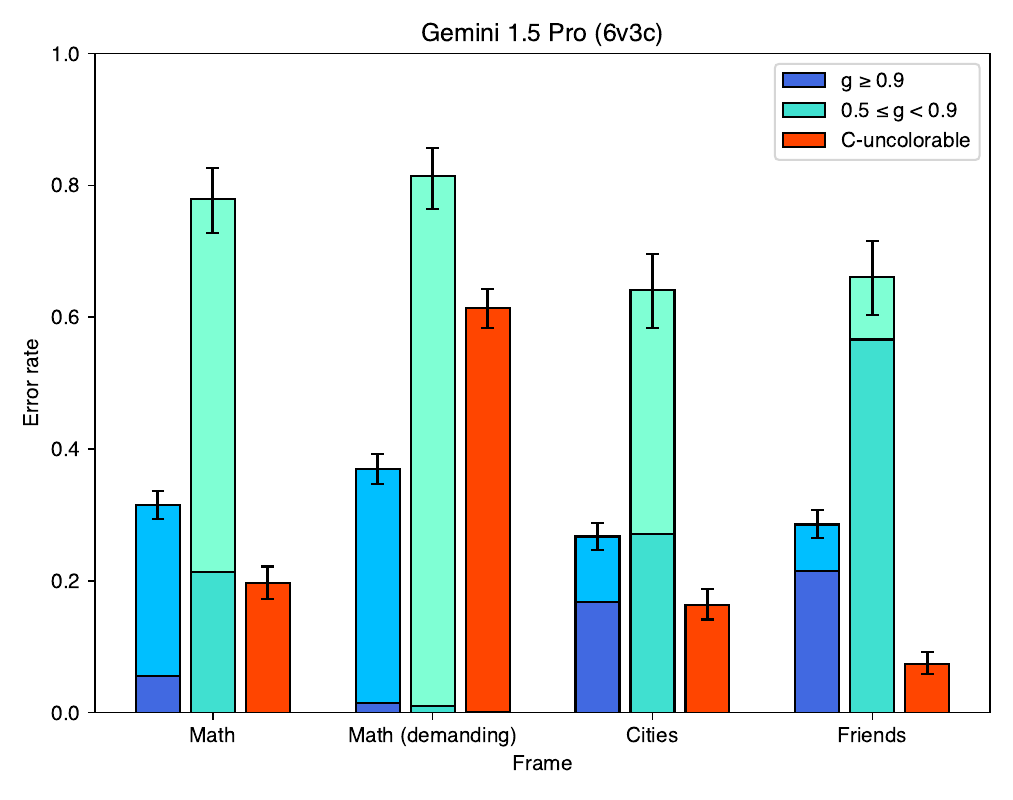}
\end{subfigure}
\end{figure}
\vspace{-0.45in}

\begin{figure}[H]
\begin{subfigure}
\centering
\includegraphics[width=0.5\textwidth]{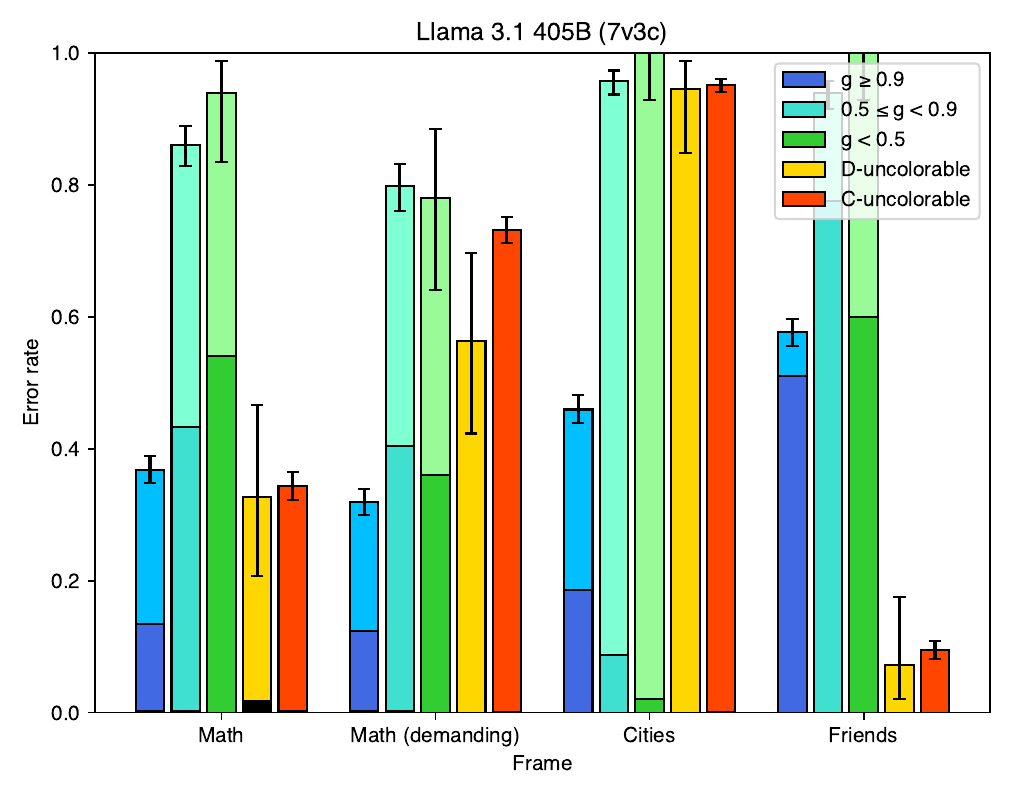}
\end{subfigure}
\hspace*{-0.9em}
\begin{subfigure}
\centering
\includegraphics[width=0.5\textwidth]{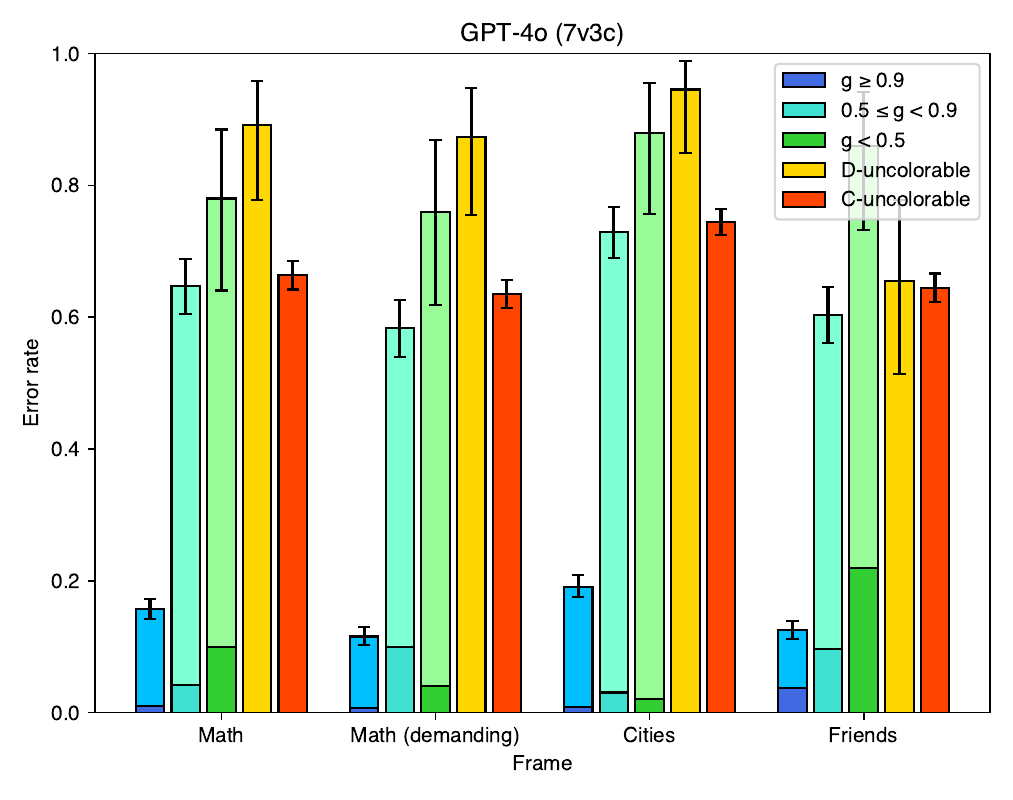}
\end{subfigure}
\end{figure}
\vspace{-0.45in}

\begin{figure}[H]
\begin{subfigure}
\centering
\includegraphics[width=0.5\textwidth]{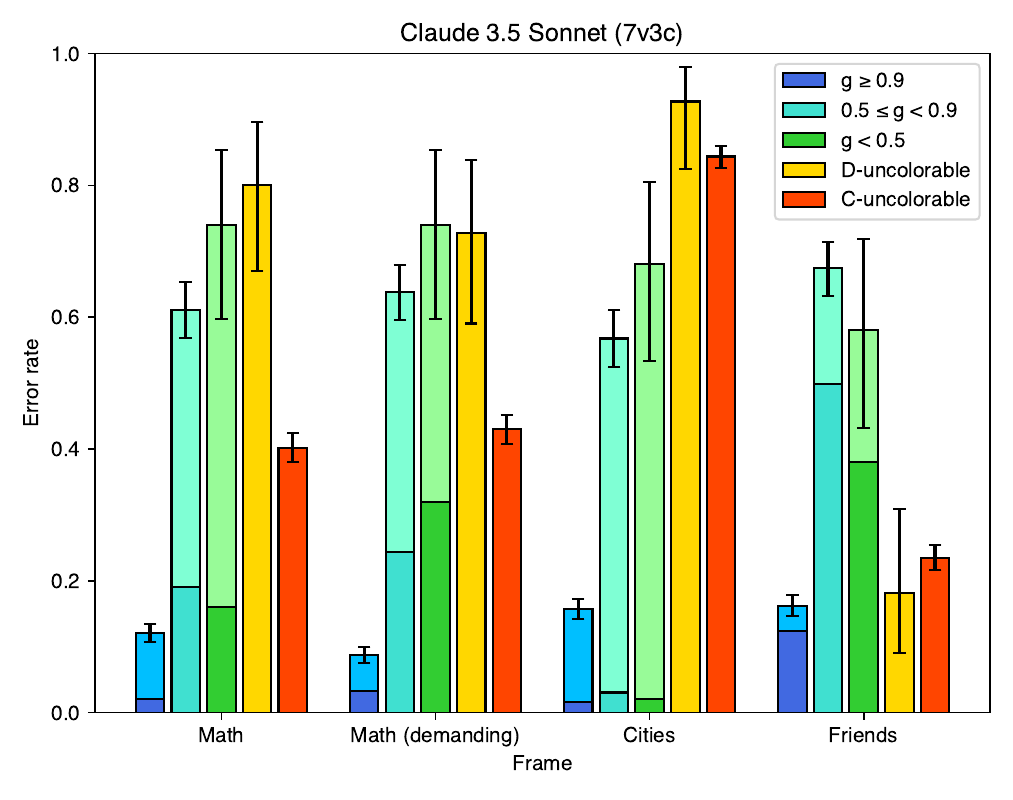}
\end{subfigure}
\hspace*{-0.9em}
\begin{subfigure}
\centering
\includegraphics[width=0.5\textwidth]{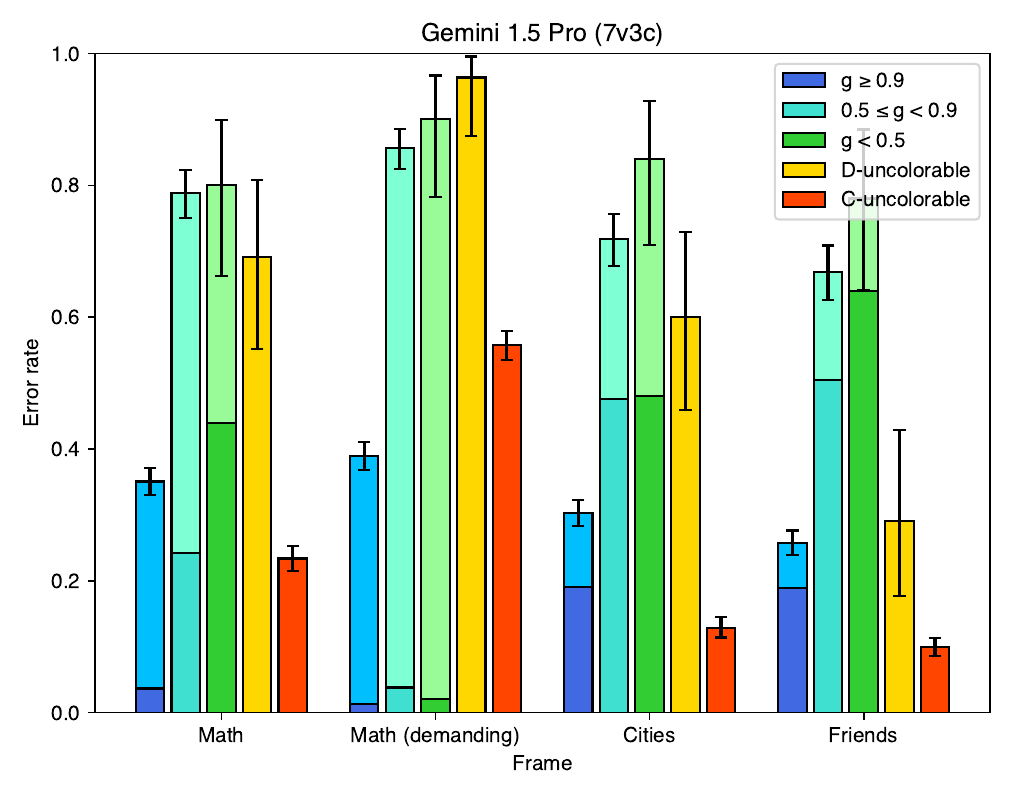}
\end{subfigure}
\end{figure}
\vspace{-0.45in}

\begin{figure}[H]
\begin{subfigure}
\centering
\includegraphics[width=0.5\textwidth]{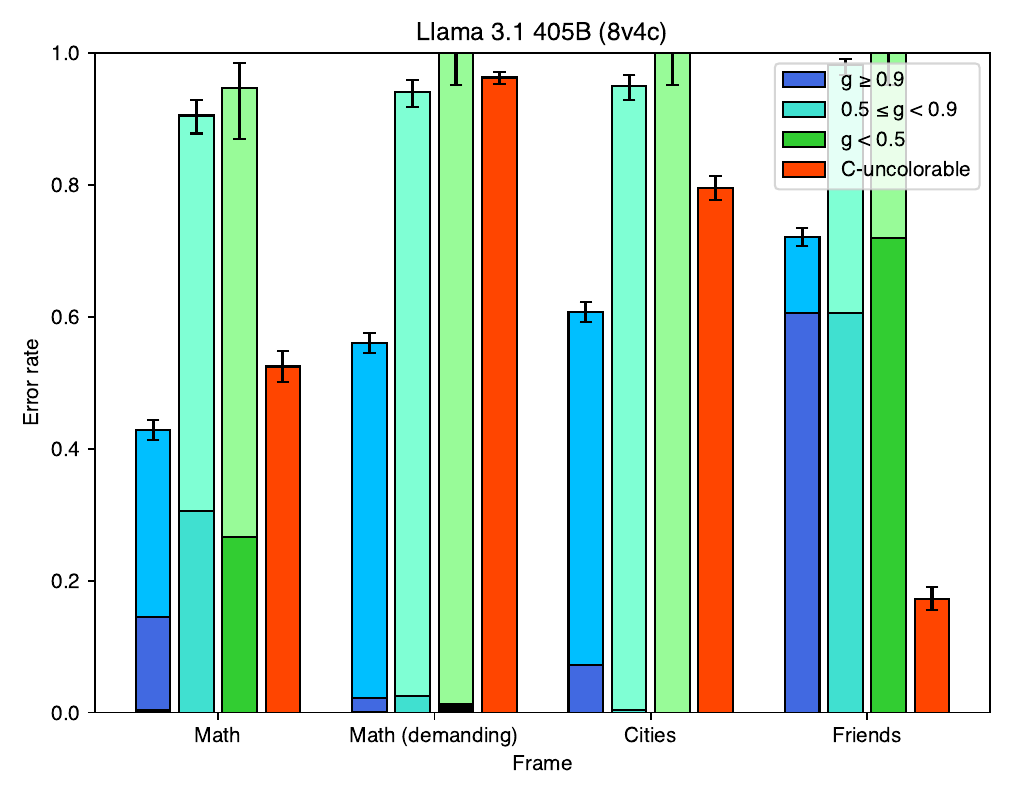}
\end{subfigure}
\hspace*{-0.9em}
\begin{subfigure}
\centering
\includegraphics[width=0.5\textwidth]{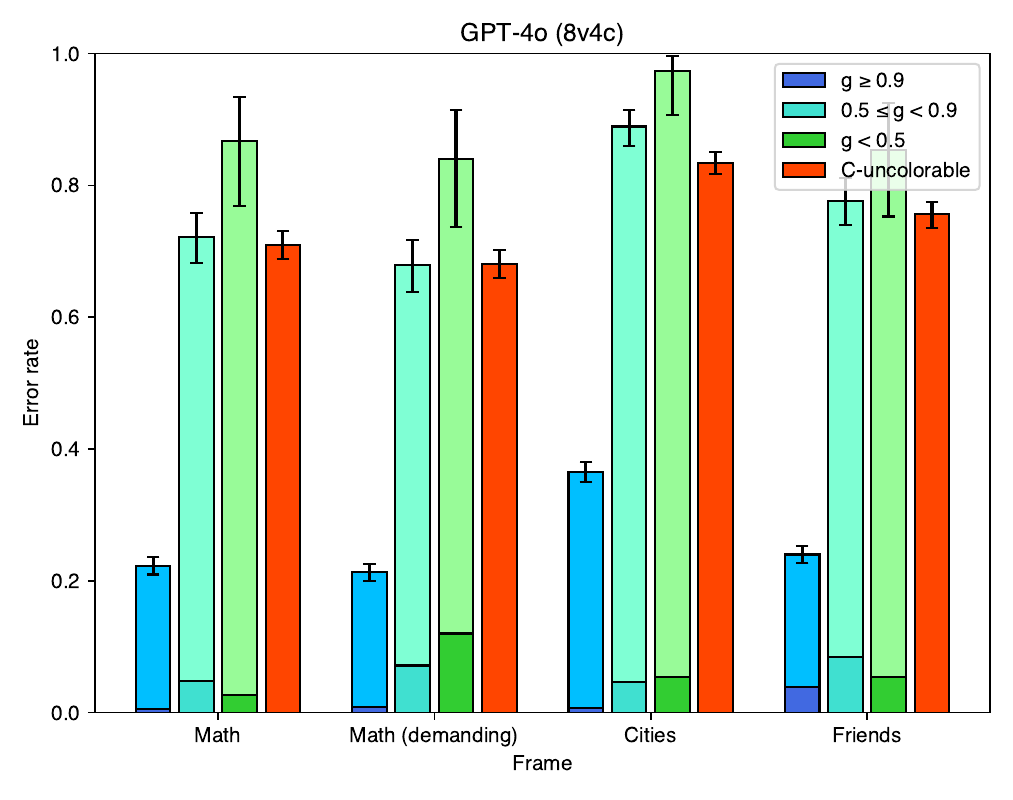}
\end{subfigure}
\end{figure}
\vspace{-0.45in}

\begin{figure}[H]
\begin{subfigure}
\centering
\includegraphics[width=0.5\textwidth]{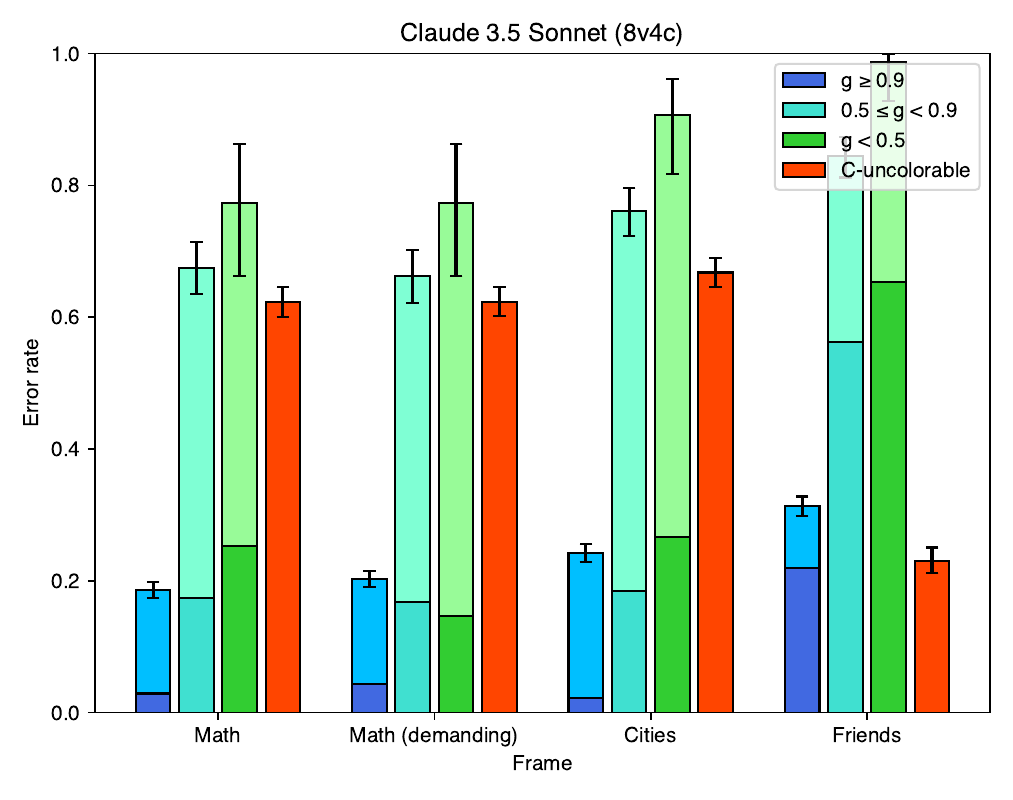}
\end{subfigure}
\hspace*{-0.9em}
\begin{subfigure}
\centering
\includegraphics[width=0.5\textwidth]{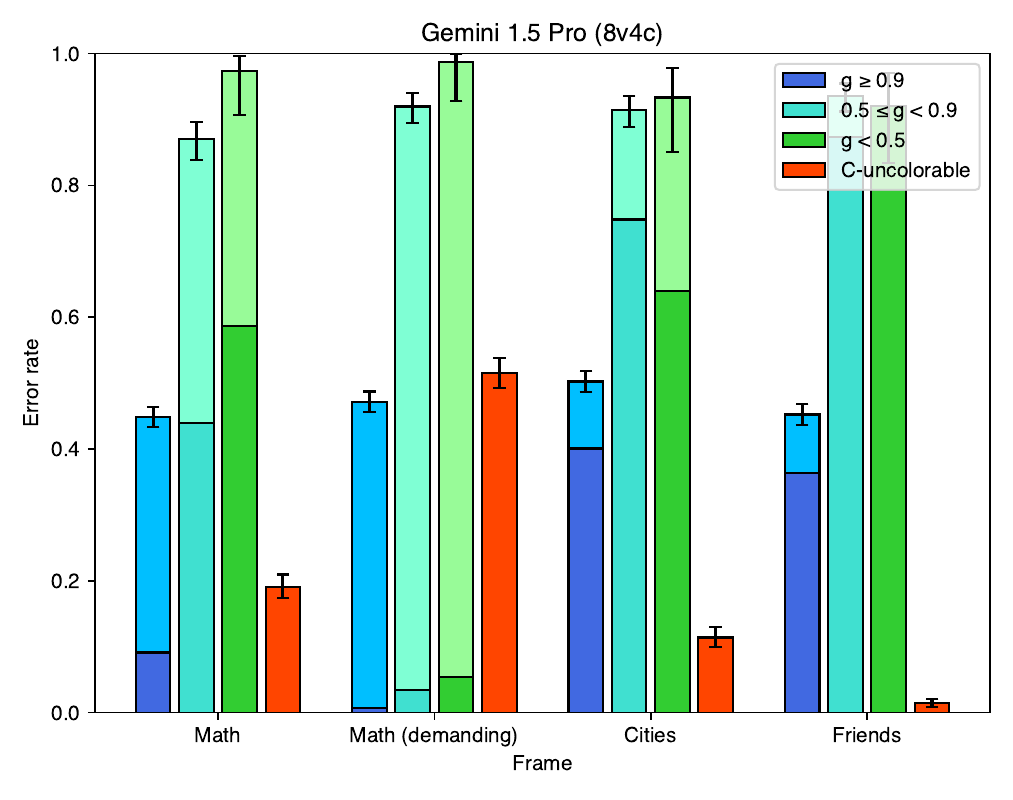}
\end{subfigure}
\end{figure}
\vspace{-0.45in}

\section{Accuracy by Model, Problem Set, Frame, and Edge Count (Standard LLMs)} \label{ec_accuracy_standard}

Following are line plots of accuracy (rate of correct answers; higher is better) for each combination of standard LLM and problem set, split further by frame and number of edges. Error bars represent $95\%$ Clopper-Pearson binomial confidence intervals. Also shown are the proportions of each colorable problem type at each edge count. Each problem set's maximum edge count is excluded due to low statistical power (1 problem, 5 responses).

\begin{figure}[H]
\begin{subfigure}
\centering
\includegraphics[width=0.5\textwidth]{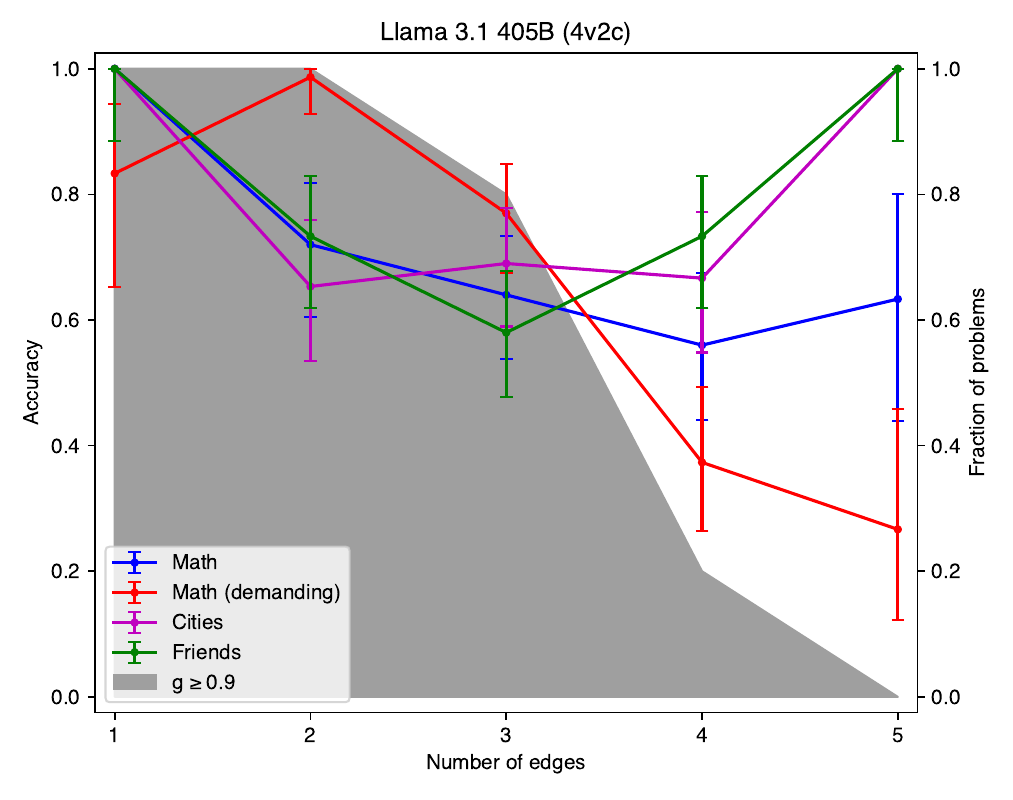}
\end{subfigure}
\hspace*{-0.9em}
\begin{subfigure}
\centering
\includegraphics[width=0.5\textwidth]{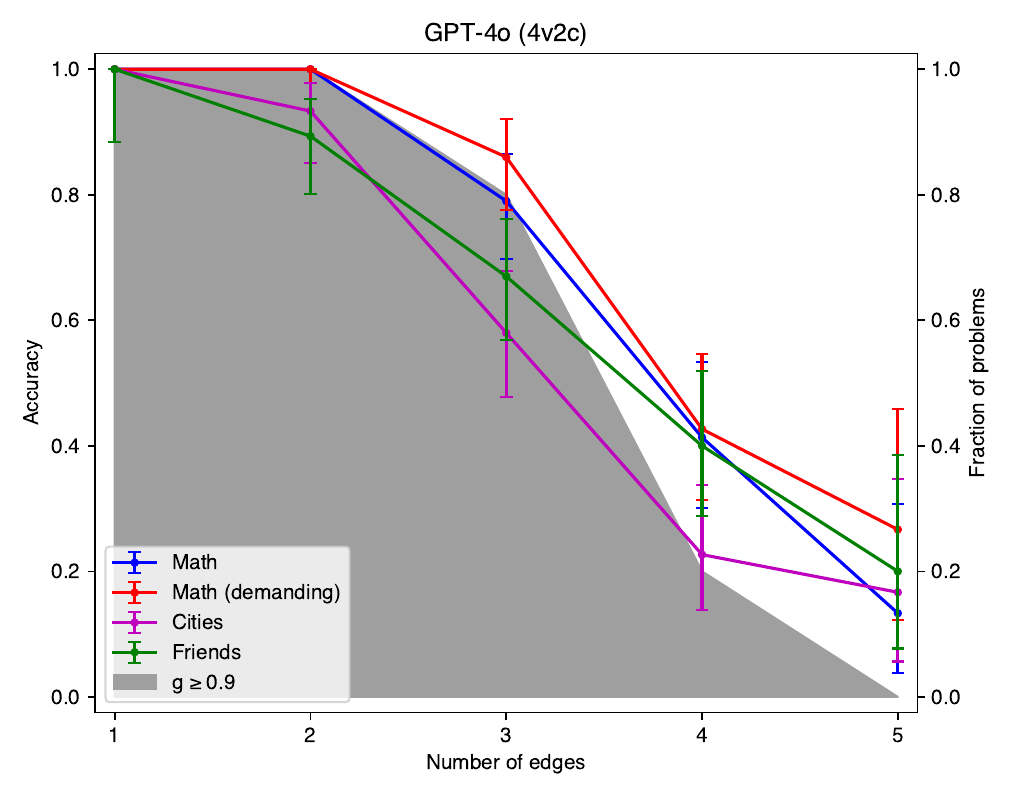}
\end{subfigure}
\end{figure}
\vspace{-0.45in}

\begin{figure}[H]
\begin{subfigure}
\centering
\includegraphics[width=0.5\textwidth]{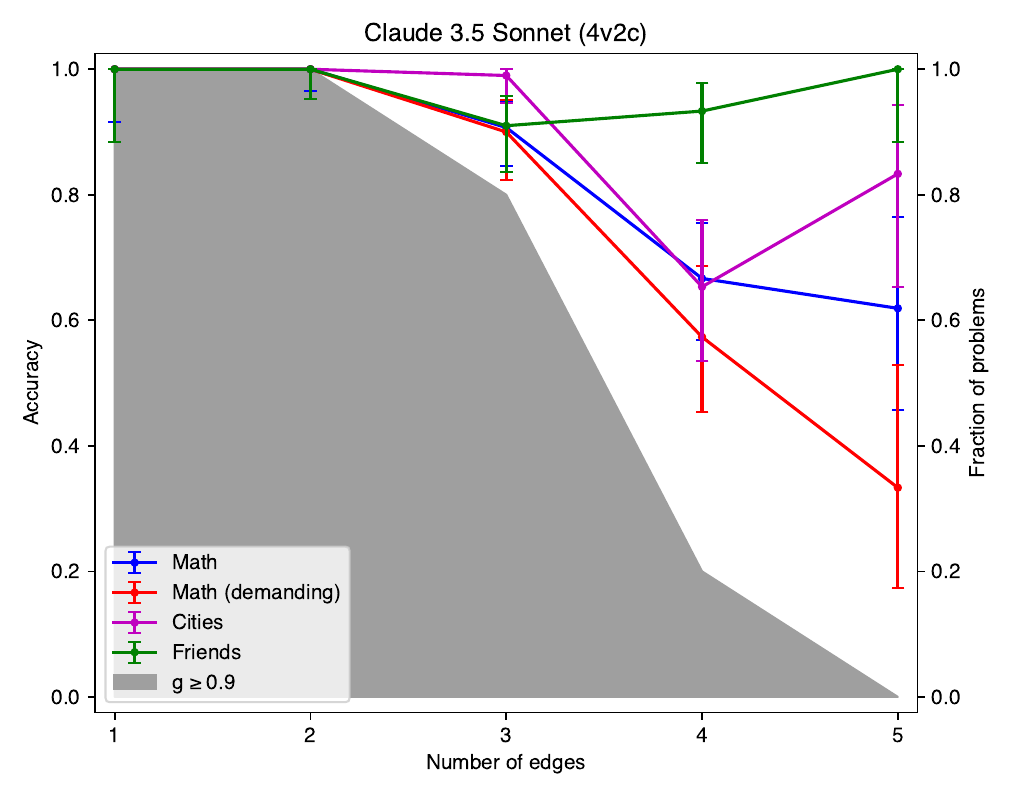}
\end{subfigure}
\hspace*{-0.9em}
\begin{subfigure}
\centering
\includegraphics[width=0.5\textwidth]{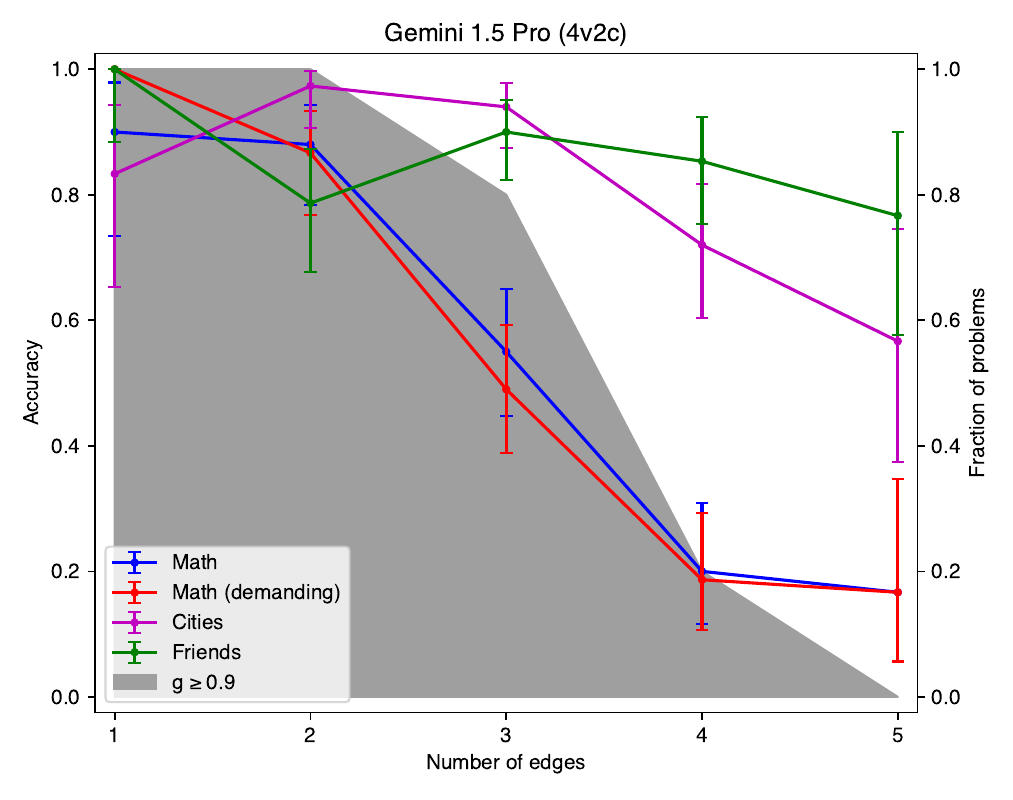}
\end{subfigure}
\end{figure}
\vspace{-0.45in}

\begin{figure}[H]
\begin{subfigure}
\centering
\includegraphics[width=0.5\textwidth]{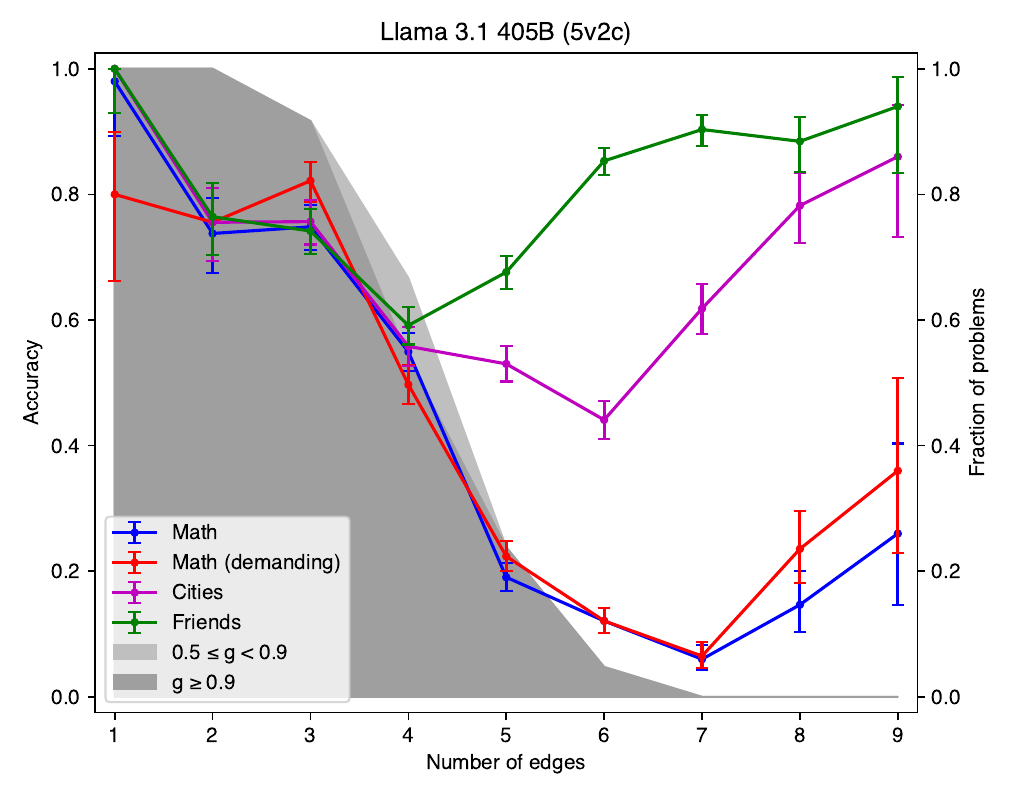}
\end{subfigure}
\hspace*{-0.9em}
\begin{subfigure}
\centering
\includegraphics[width=0.5\textwidth]{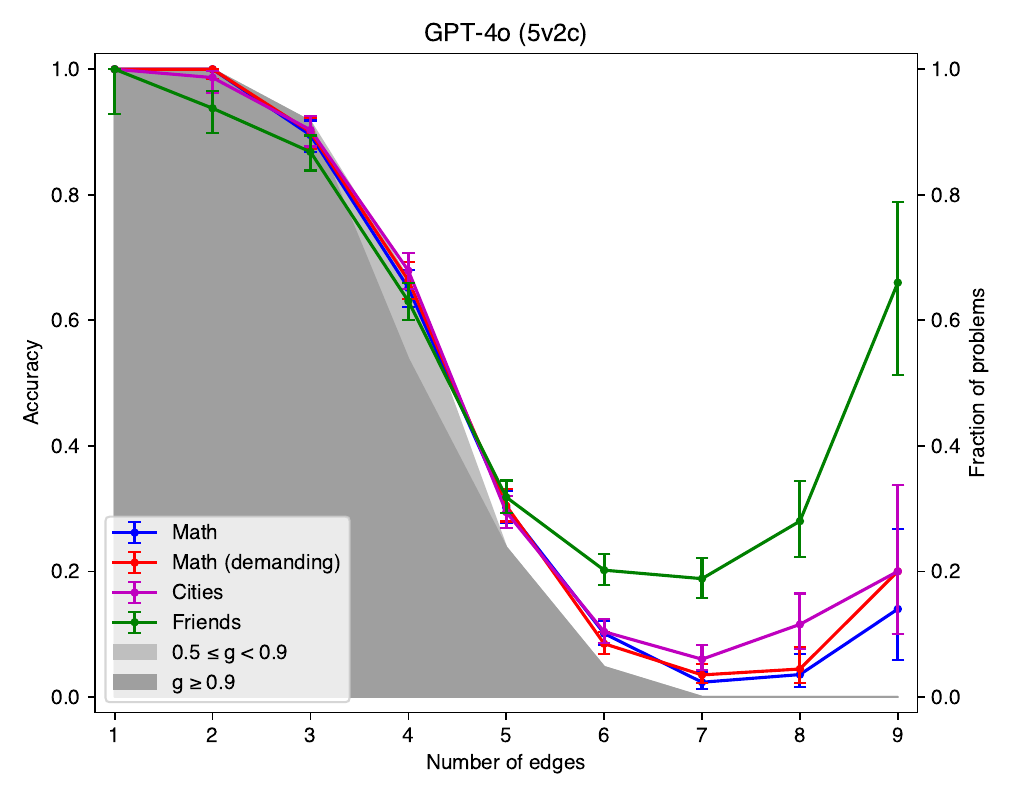}
\end{subfigure}
\end{figure}
\vspace{-0.45in}

\begin{figure}[H]
\begin{subfigure}
\centering
\includegraphics[width=0.5\textwidth]{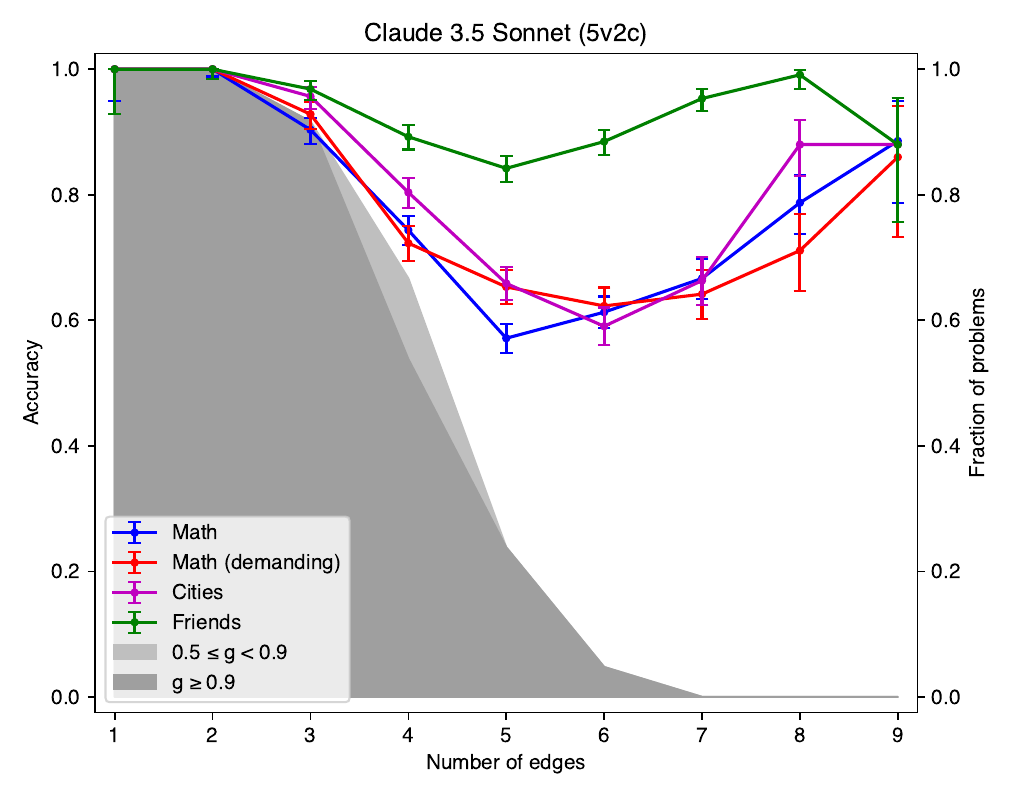}
\end{subfigure}
\hspace*{-0.9em}
\begin{subfigure}
\centering
\includegraphics[width=0.5\textwidth]{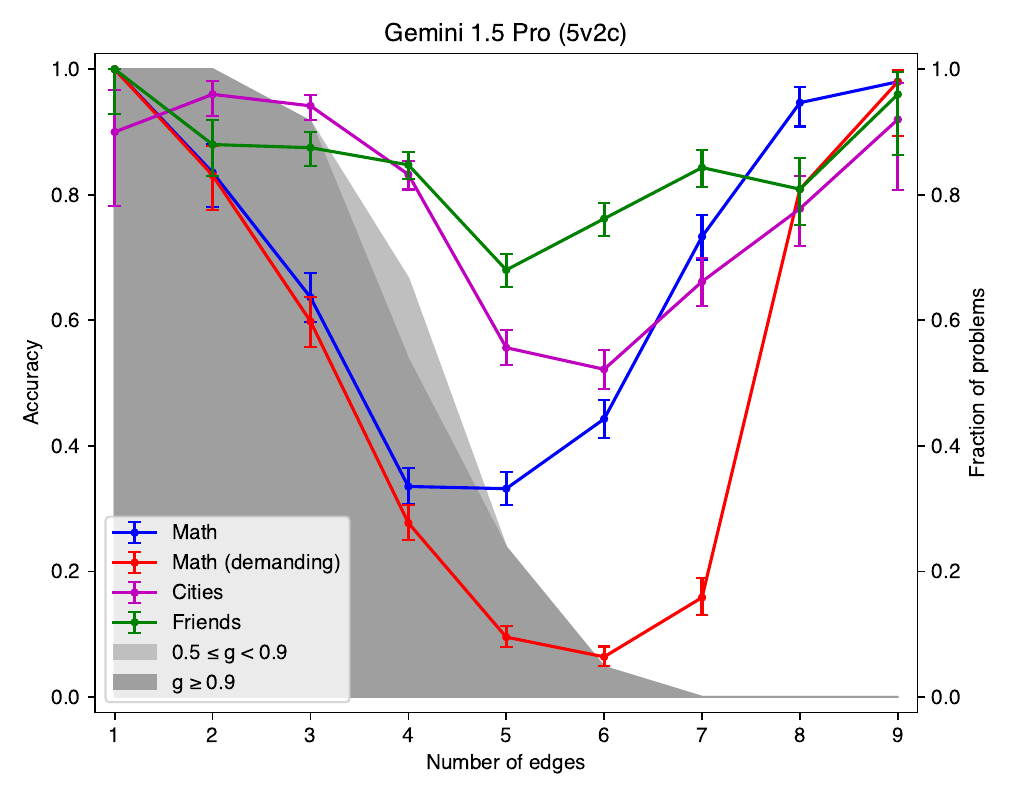}
\end{subfigure}
\end{figure}
\vspace{-0.45in}

\begin{figure}[H]
\begin{subfigure}
\centering
\includegraphics[width=0.5\textwidth]{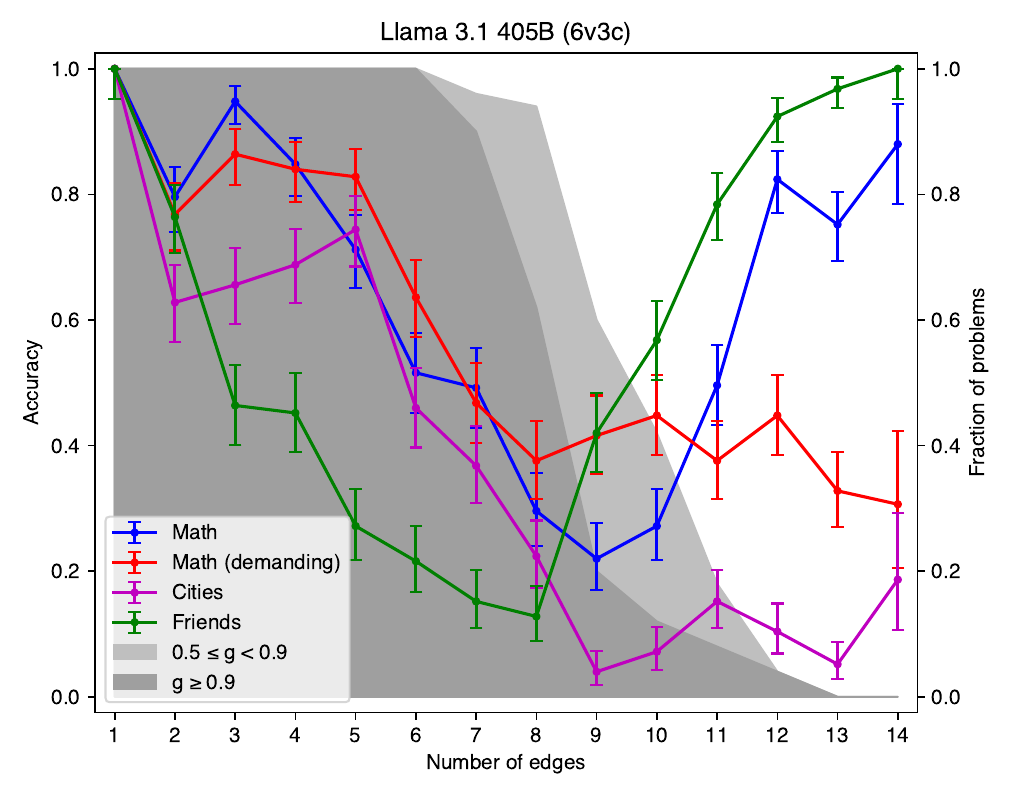}
\end{subfigure}
\hspace*{-0.9em}
\begin{subfigure}
\centering
\includegraphics[width=0.5\textwidth]{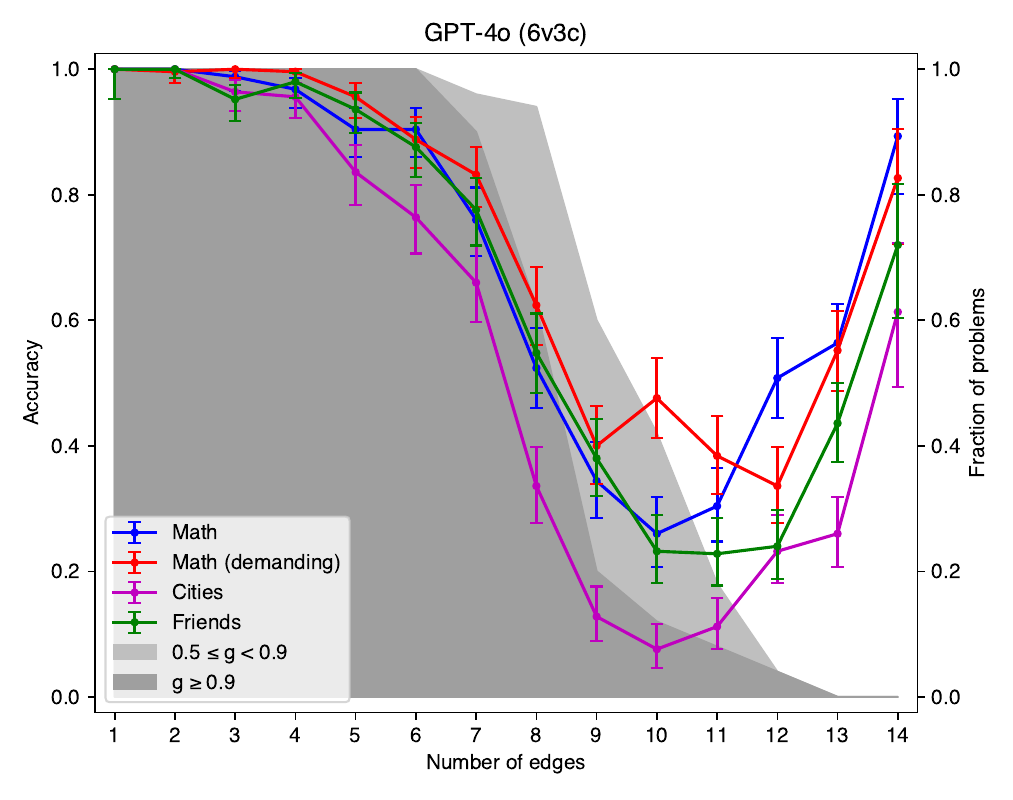}
\end{subfigure}
\end{figure}
\vspace{-0.45in}

\begin{figure}[H]
\begin{subfigure}
\centering
\includegraphics[width=0.5\textwidth]{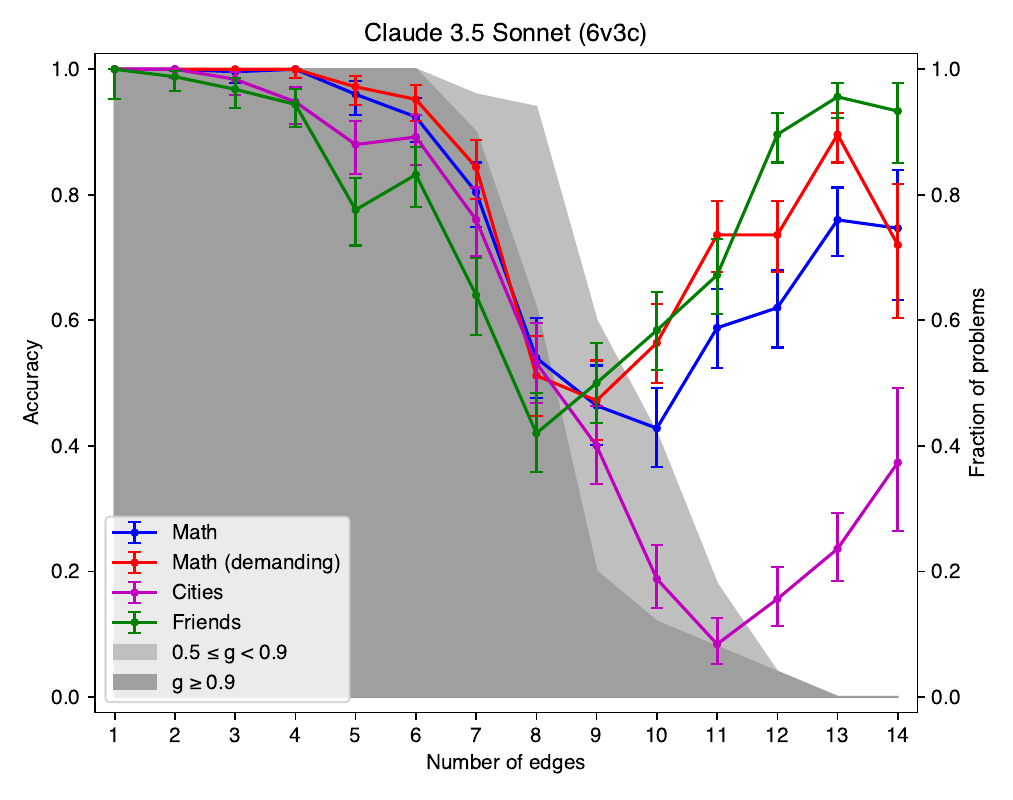}
\end{subfigure}
\hspace*{-0.9em}
\begin{subfigure}
\centering
\includegraphics[width=0.5\textwidth]{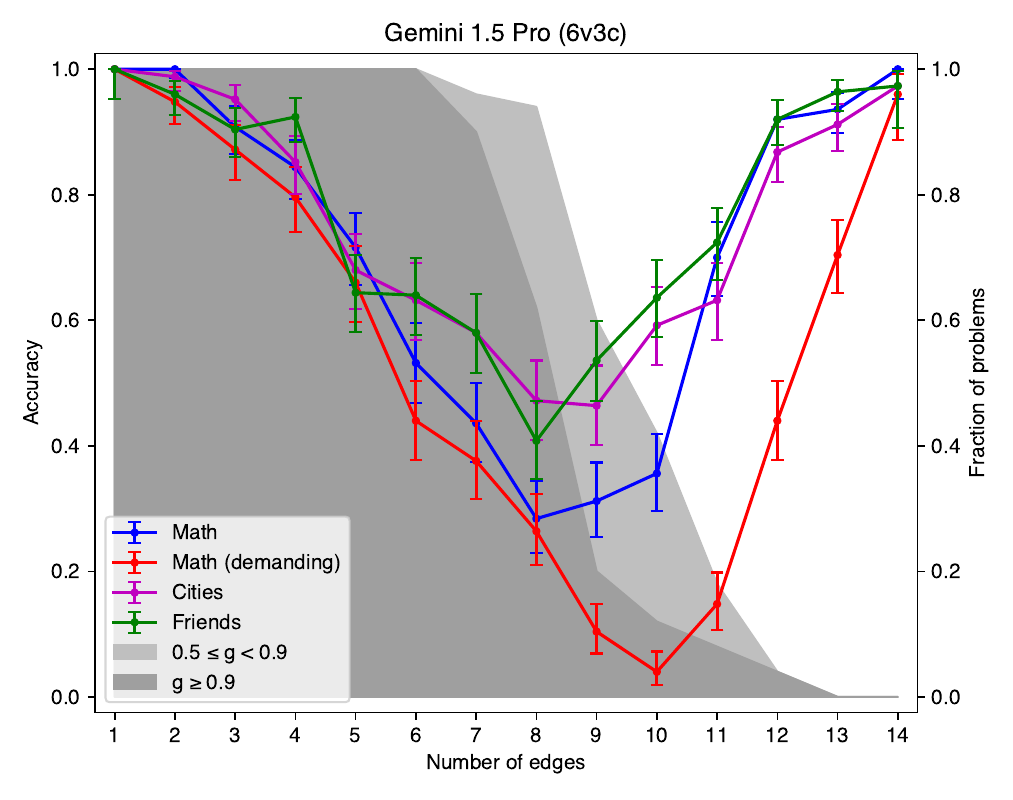}
\end{subfigure}
\end{figure}
\vspace{-0.45in}

\begin{figure}[H]
\begin{subfigure}
\centering
\includegraphics[width=0.5\textwidth]{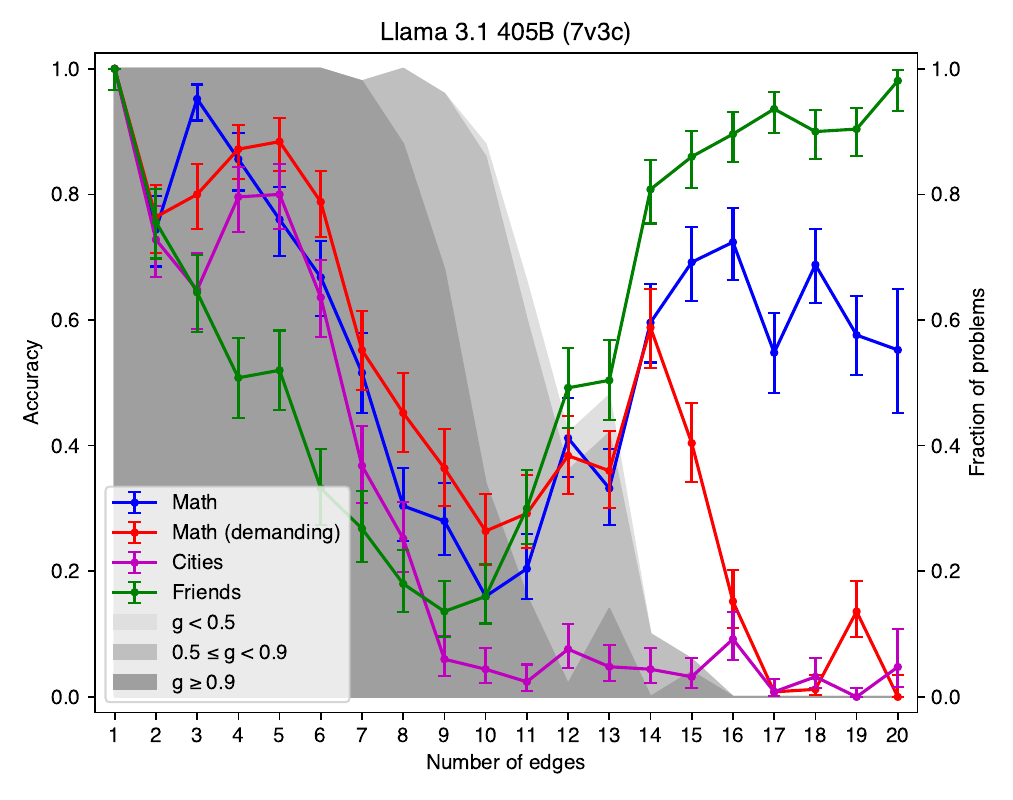}
\end{subfigure}
\hspace*{-0.9em}
\begin{subfigure}
\centering
\includegraphics[width=0.5\textwidth]{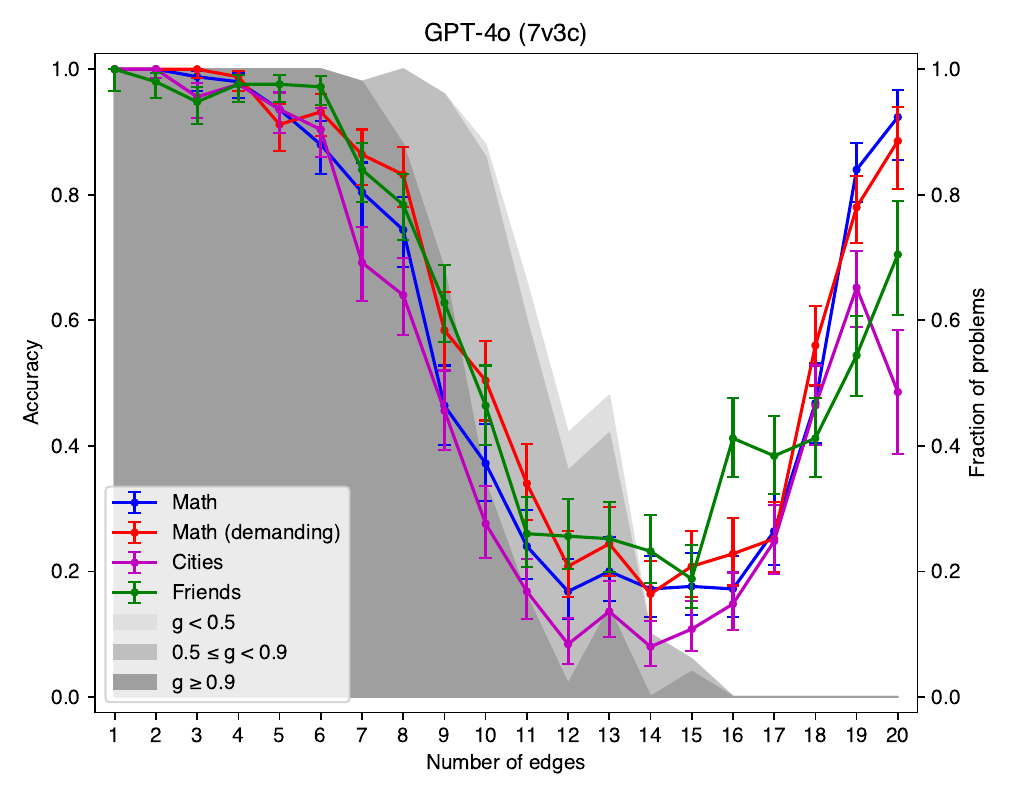}
\end{subfigure}
\end{figure}
\vspace{-0.45in}

\begin{figure}[H]
\begin{subfigure}
\centering
\includegraphics[width=0.5\textwidth]{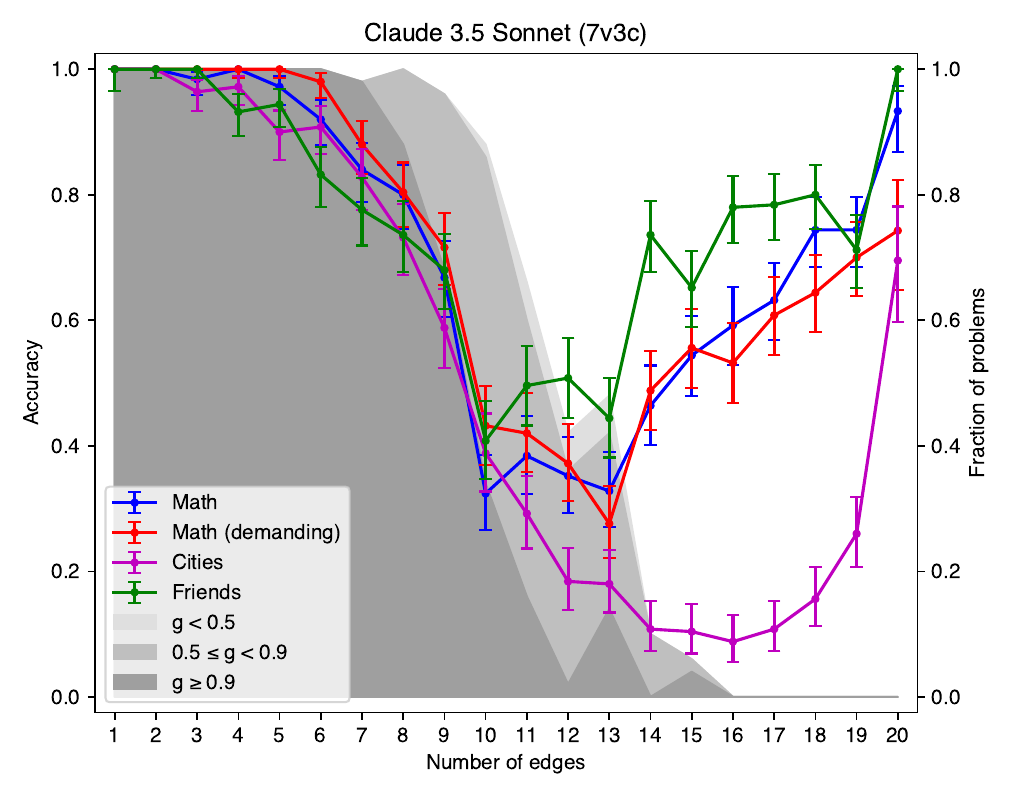}
\end{subfigure}
\hspace*{-0.9em}
\begin{subfigure}
\centering
\includegraphics[width=0.5\textwidth]{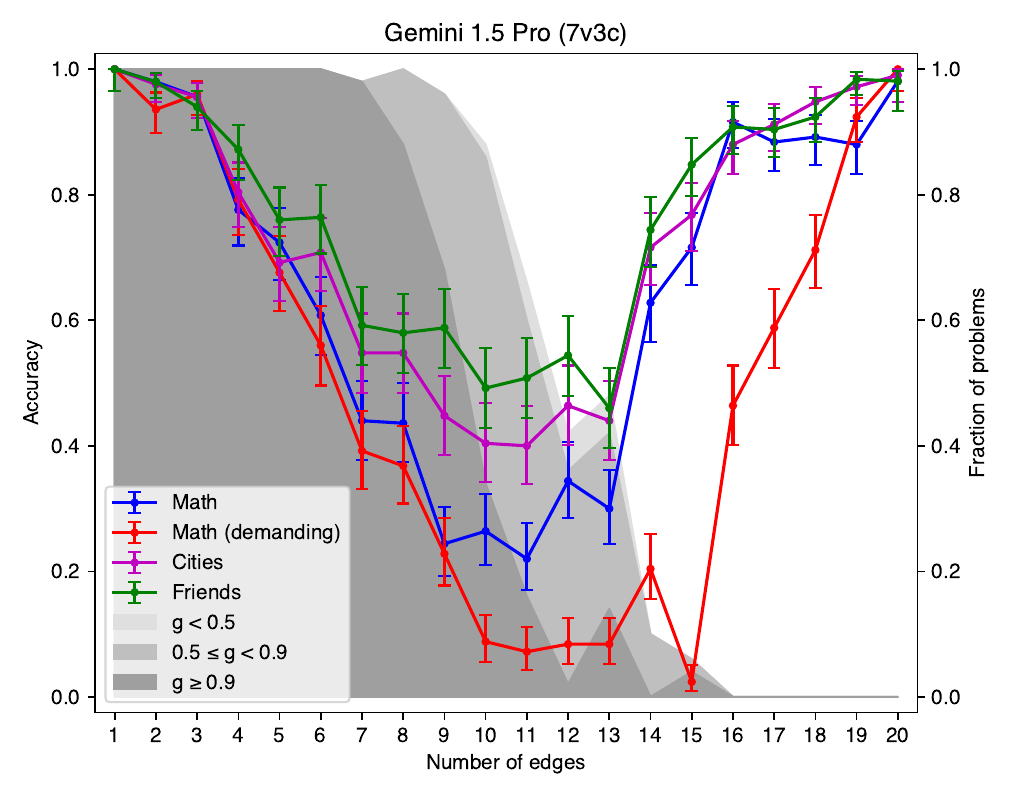}
\end{subfigure}
\end{figure}
\vspace{-0.45in}

\begin{figure}[H]
\begin{subfigure}
\centering
\includegraphics[width=0.5\textwidth]{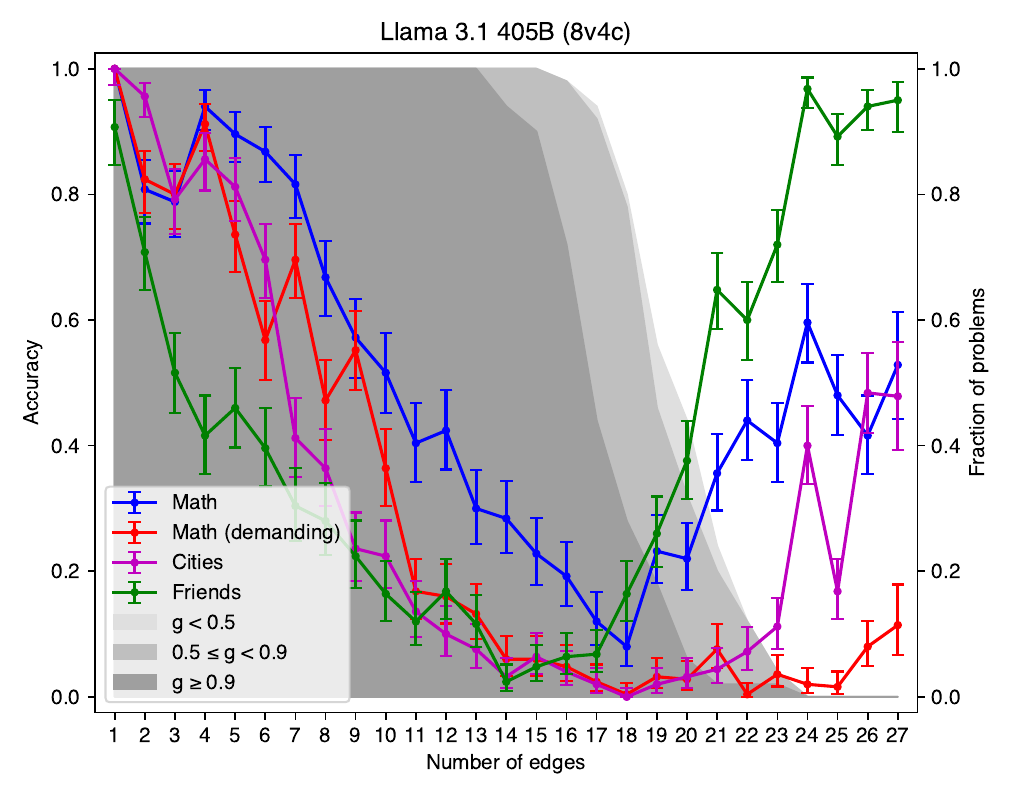}
\end{subfigure}
\hspace*{-0.9em}
\begin{subfigure}
\centering
\includegraphics[width=0.5\textwidth]{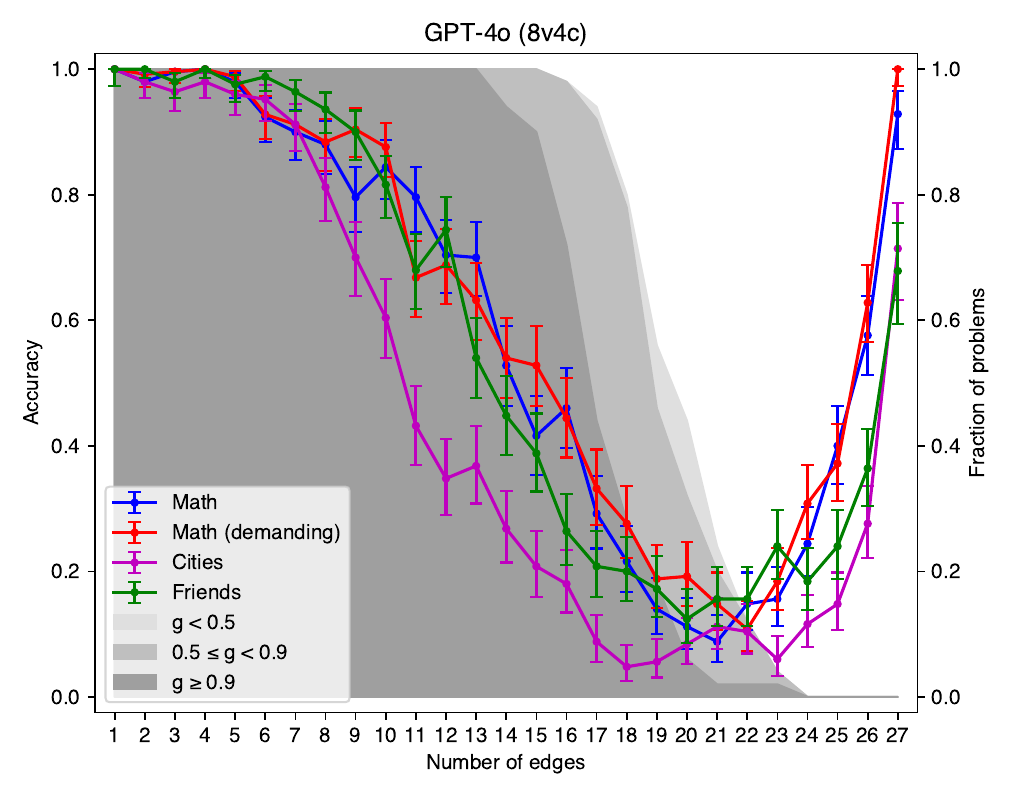}
\end{subfigure}
\end{figure}
\vspace{-0.45in}

\begin{figure}[H]
\begin{subfigure}
\centering
\includegraphics[width=0.5\textwidth]{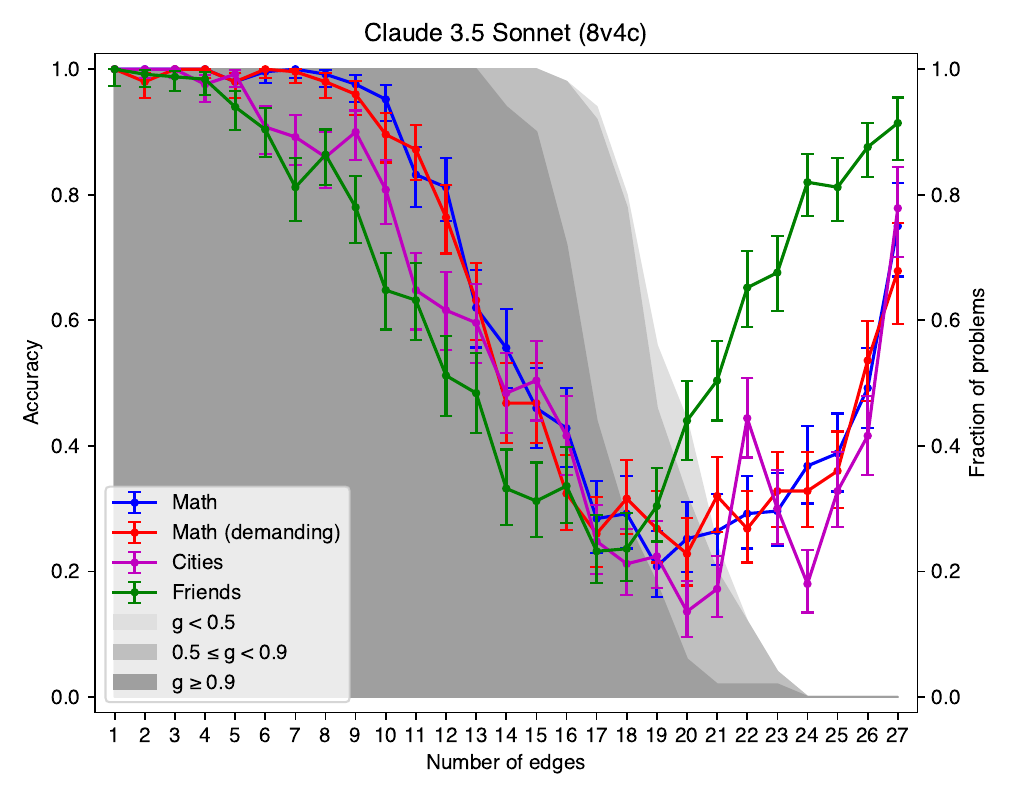}
\end{subfigure}
\hspace*{-0.9em}
\begin{subfigure}
\centering
\includegraphics[width=0.5\textwidth]{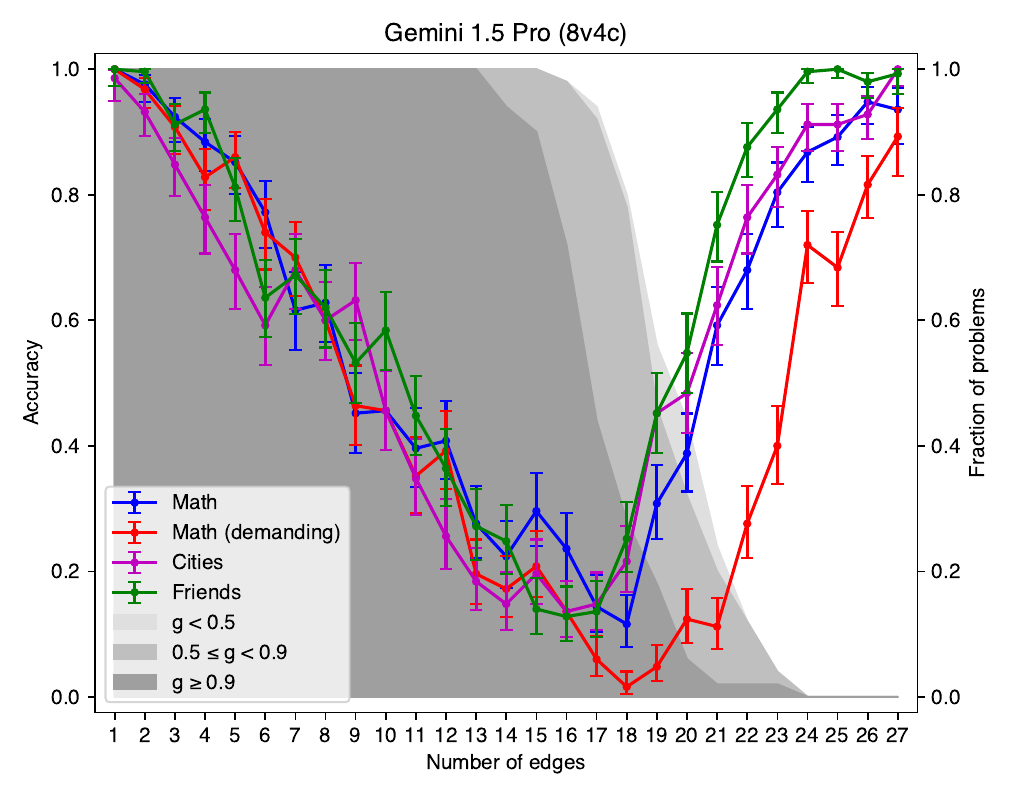}
\end{subfigure}
\end{figure}
\vspace{-0.45in}

\section{Error Rates by Model, Problem Set, Frame, and Problem Type (LRMs)} \label{type_error_lrm}

Following are bar plots of error rates for each combination of LRM and edge-selected problem set, split further by frame and problem type. These present the same information as Figure \ref{fig:all_sel_models} with additional detail. Error bars represent $95\%$ Clopper-Pearson binomial confidence intervals. For the three colorable problem types, a lighter-colored segment at the top of a bar represents trials where the model answered with an invalid coloring, while the normally colored bar segment represents trials where the model falsely declared the graph uncolorable. The y-axis is compressed in comparison to the plots in Appendix \ref{type_error_standard} to better show the differences between the smaller error rates.

\begin{figure}[H]
\begin{subfigure}
\centering
\includegraphics[width=0.5\textwidth]{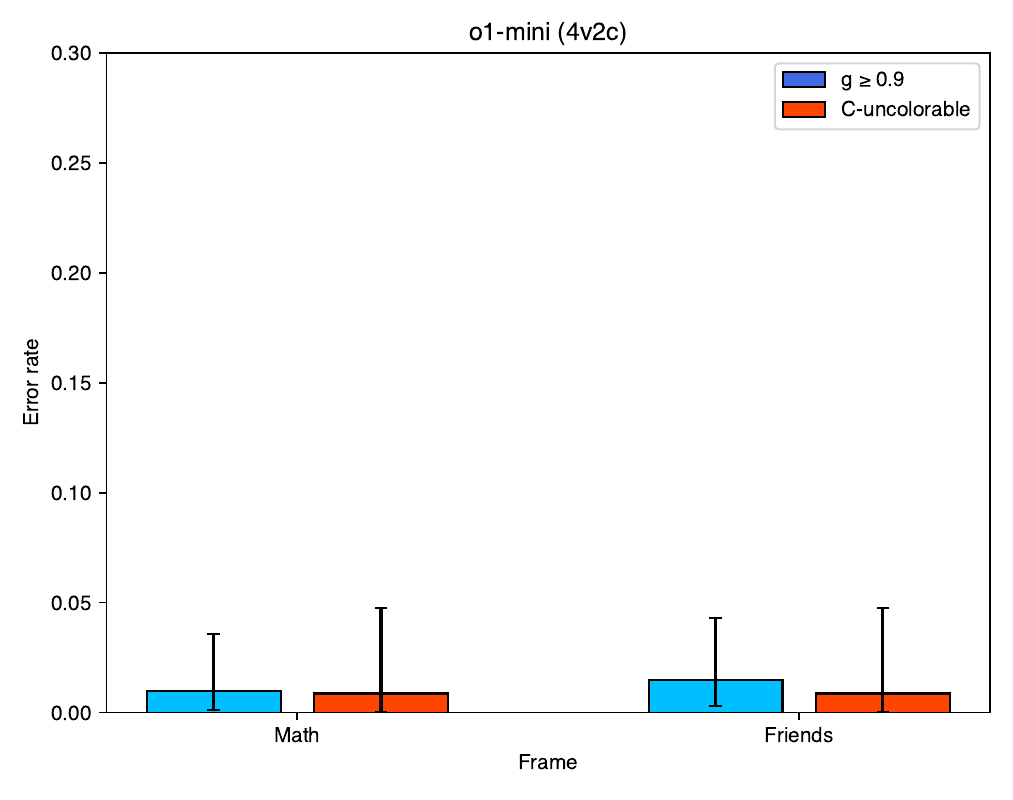}
\end{subfigure}
\hspace*{-0.9em}
\begin{subfigure}
\centering
\includegraphics[width=0.5\textwidth]{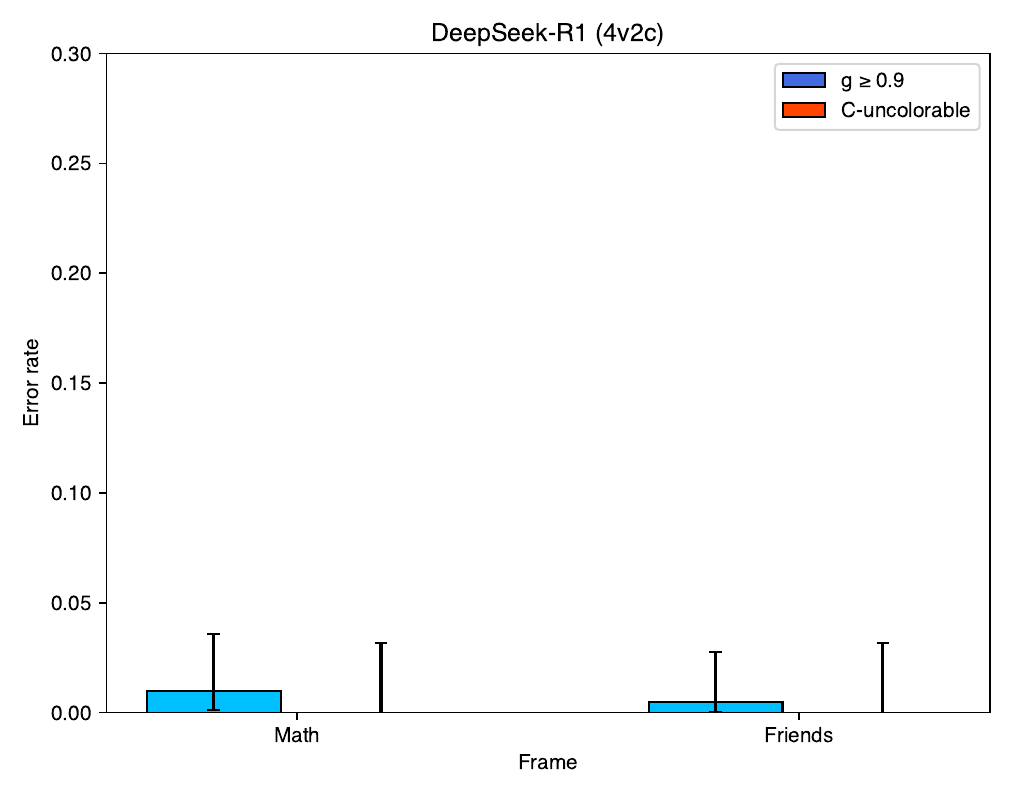}
\end{subfigure}
\end{figure}
\vspace{-0.45in}

\begin{figure}[H]
\begin{subfigure}
\centering
\includegraphics[width=0.5\textwidth]{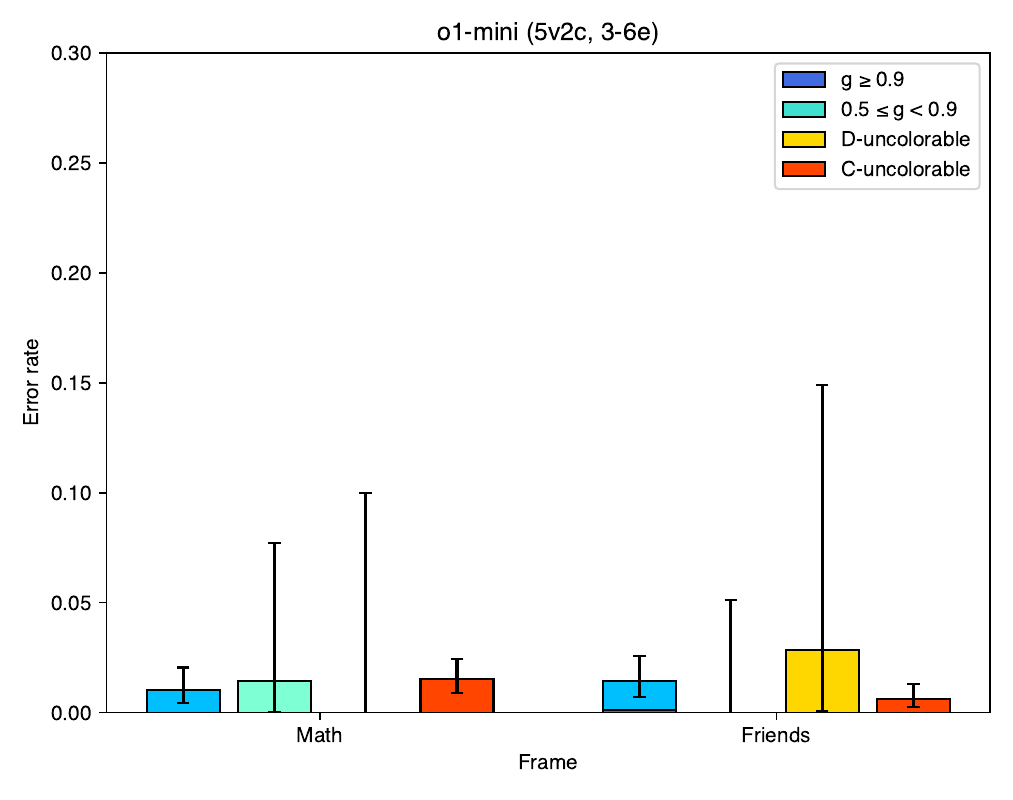}
\end{subfigure}
\hspace*{-0.9em}
\begin{subfigure}
\centering
\includegraphics[width=0.5\textwidth]{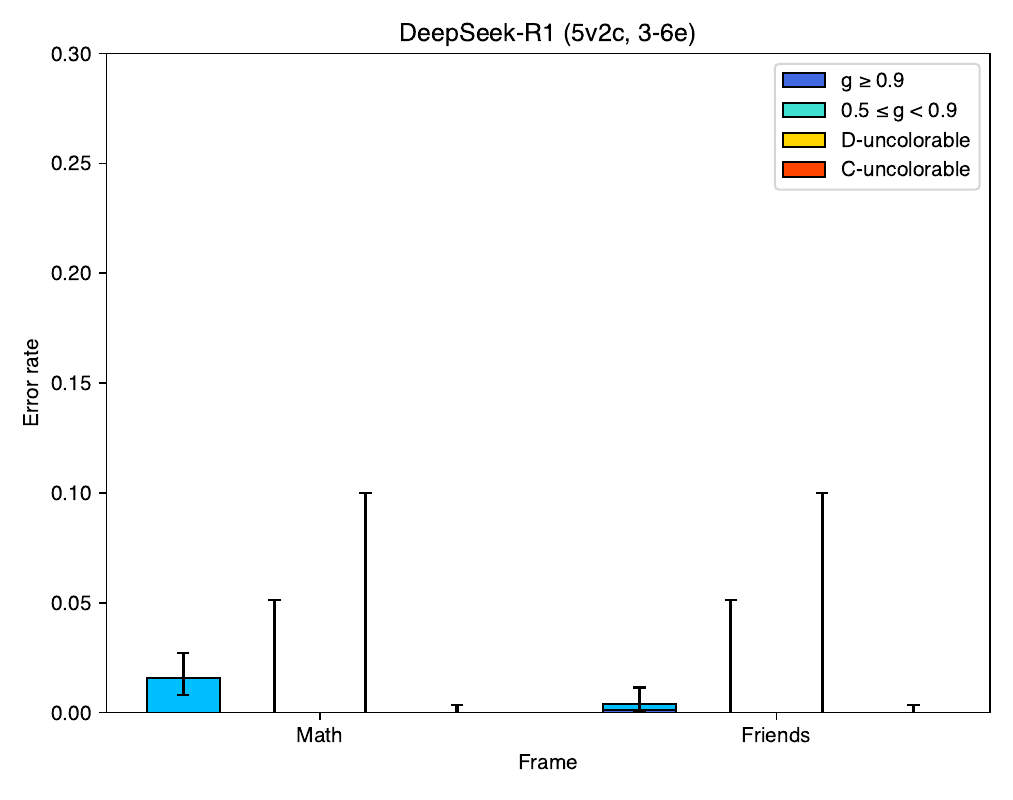}
\end{subfigure}
\end{figure}
\vspace{-0.45in}

\begin{figure}[H]
\begin{subfigure}
\centering
\includegraphics[width=0.5\textwidth]{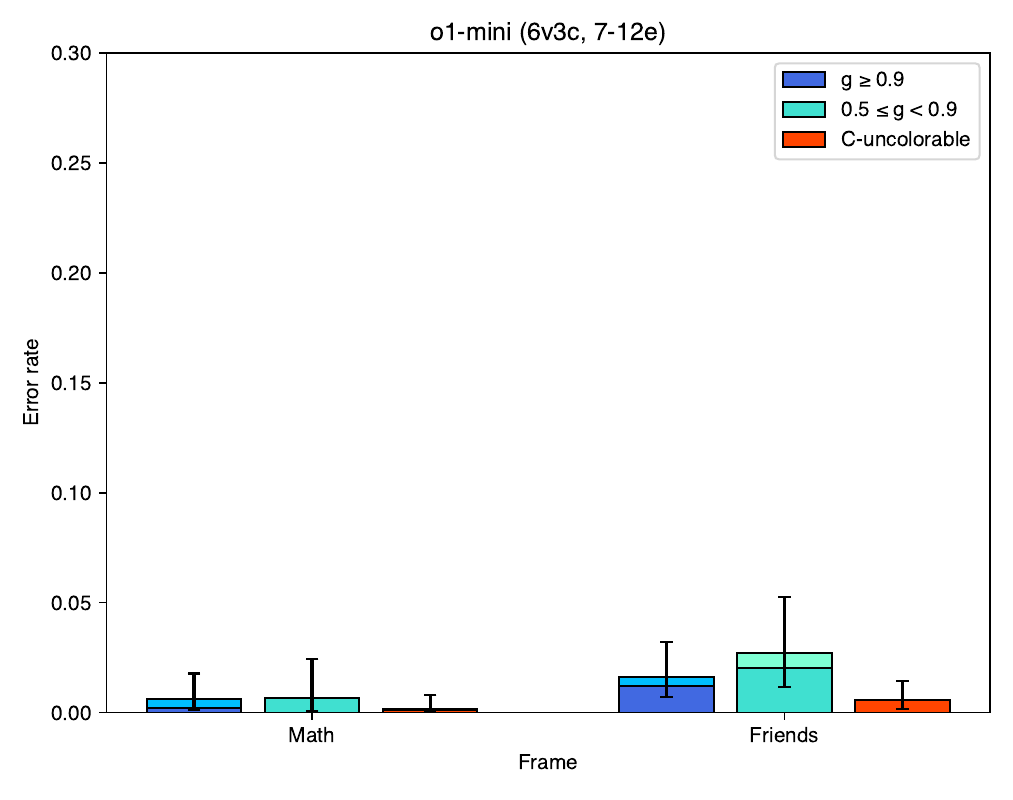}
\end{subfigure}
\hspace*{-0.9em}
\begin{subfigure}
\centering
\includegraphics[width=0.5\textwidth]{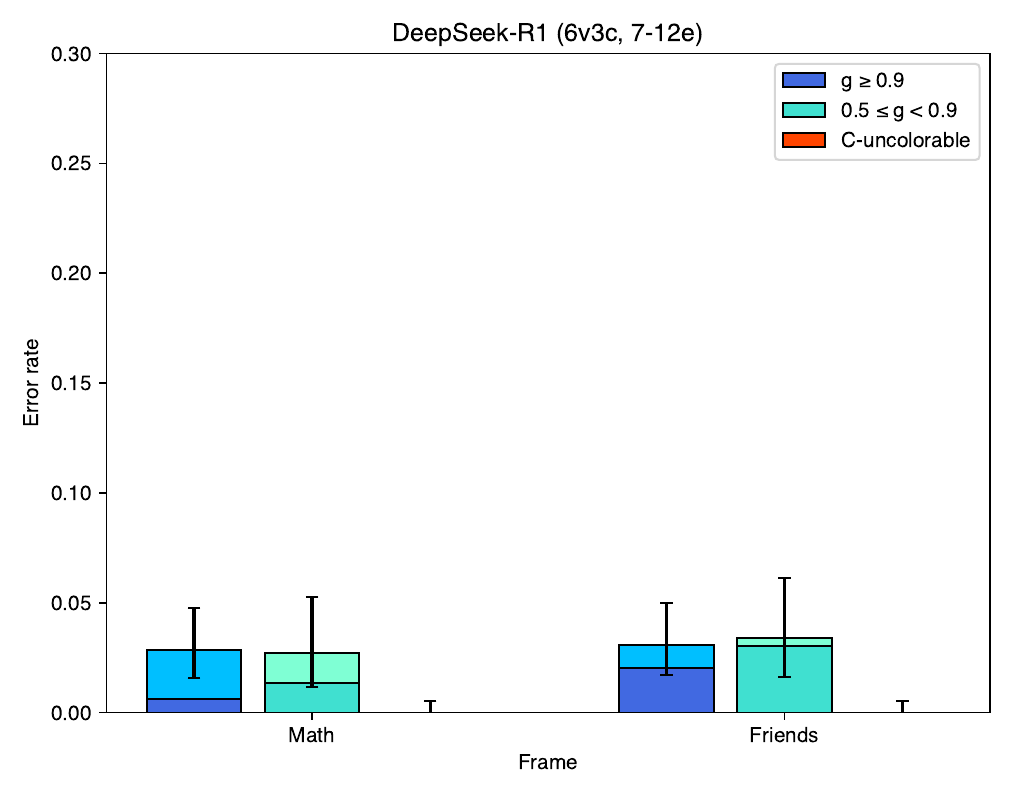}
\end{subfigure}
\end{figure}
\vspace{-0.45in}

\begin{figure}[H]
\begin{subfigure}
\centering
\includegraphics[width=0.5\textwidth]{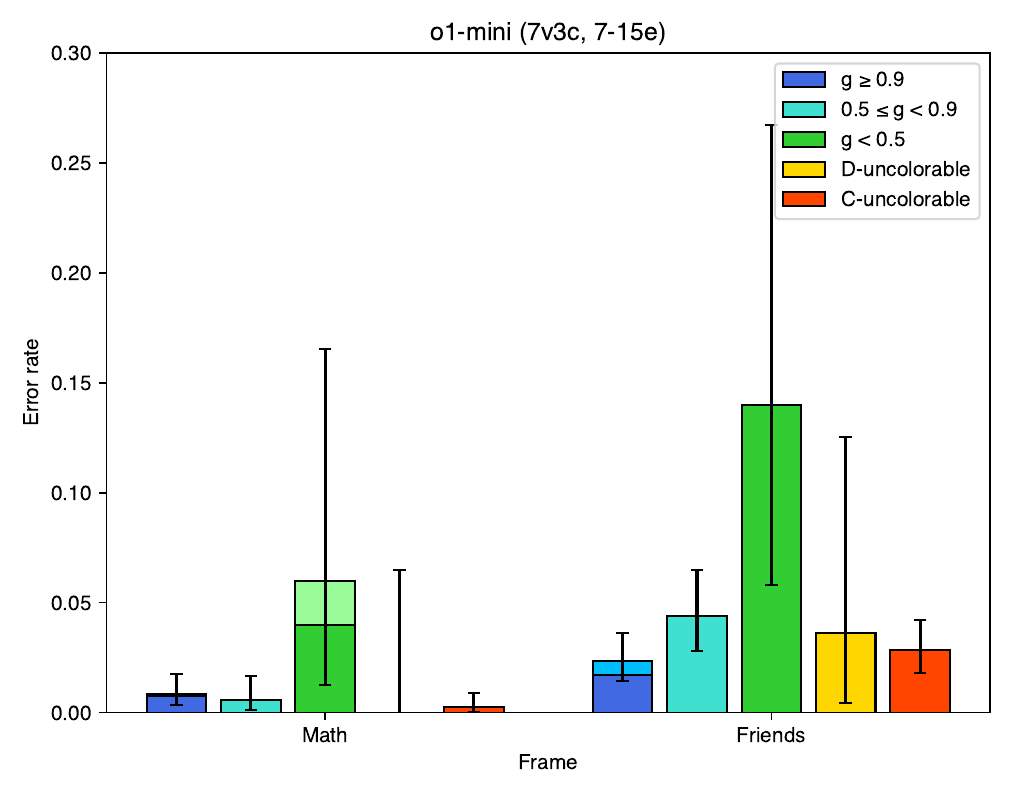}
\end{subfigure}
\hspace*{-0.9em}
\begin{subfigure}
\centering
\includegraphics[width=0.5\textwidth]{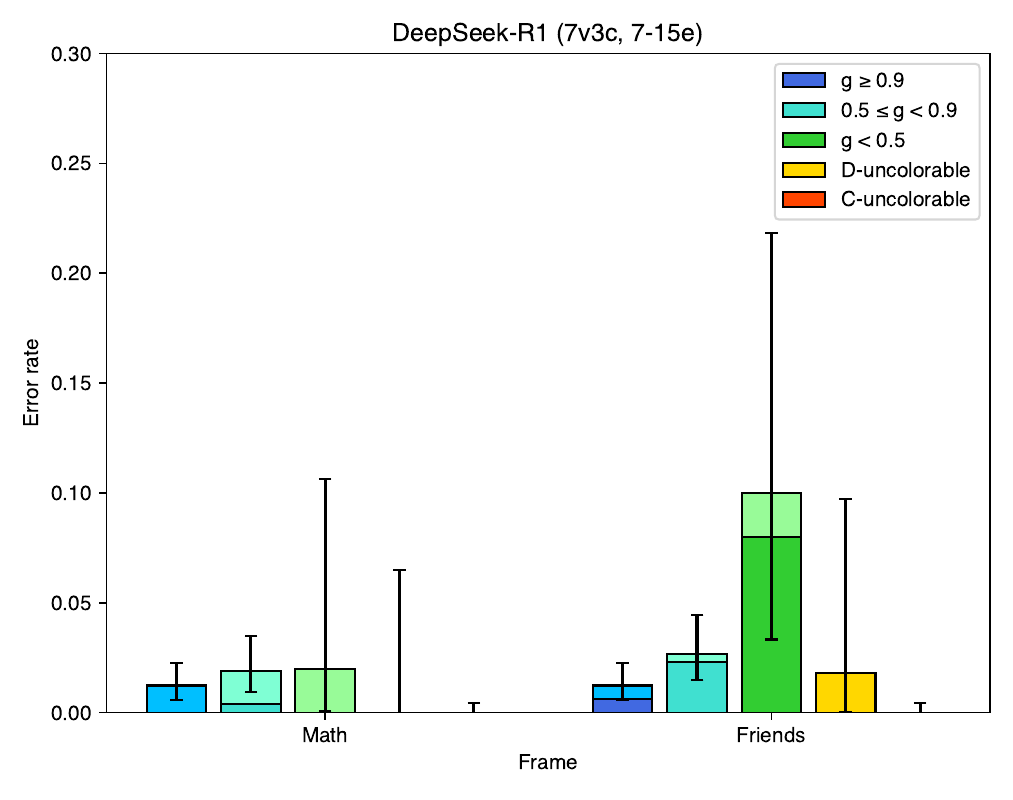}
\end{subfigure}
\end{figure}
\vspace{-0.45in}

\begin{figure}[H]
\begin{subfigure}
\centering
\includegraphics[width=0.5\textwidth]{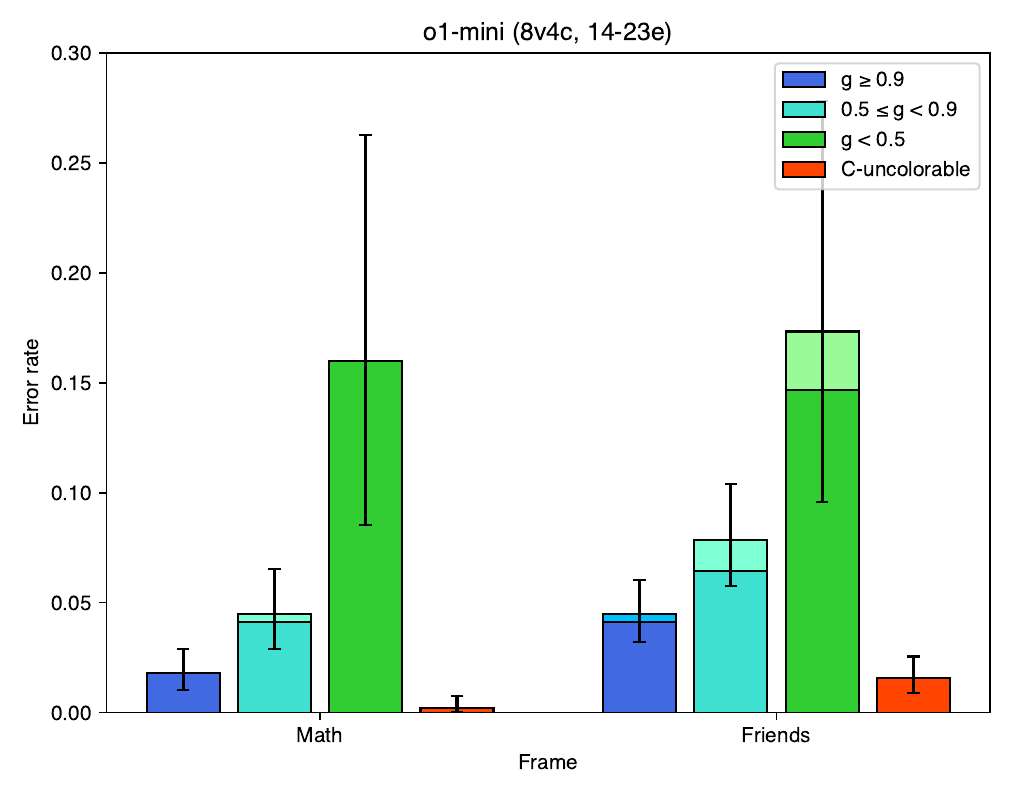}
\end{subfigure}
\hspace*{-0.9em}
\begin{subfigure}
\centering
\includegraphics[width=0.5\textwidth]{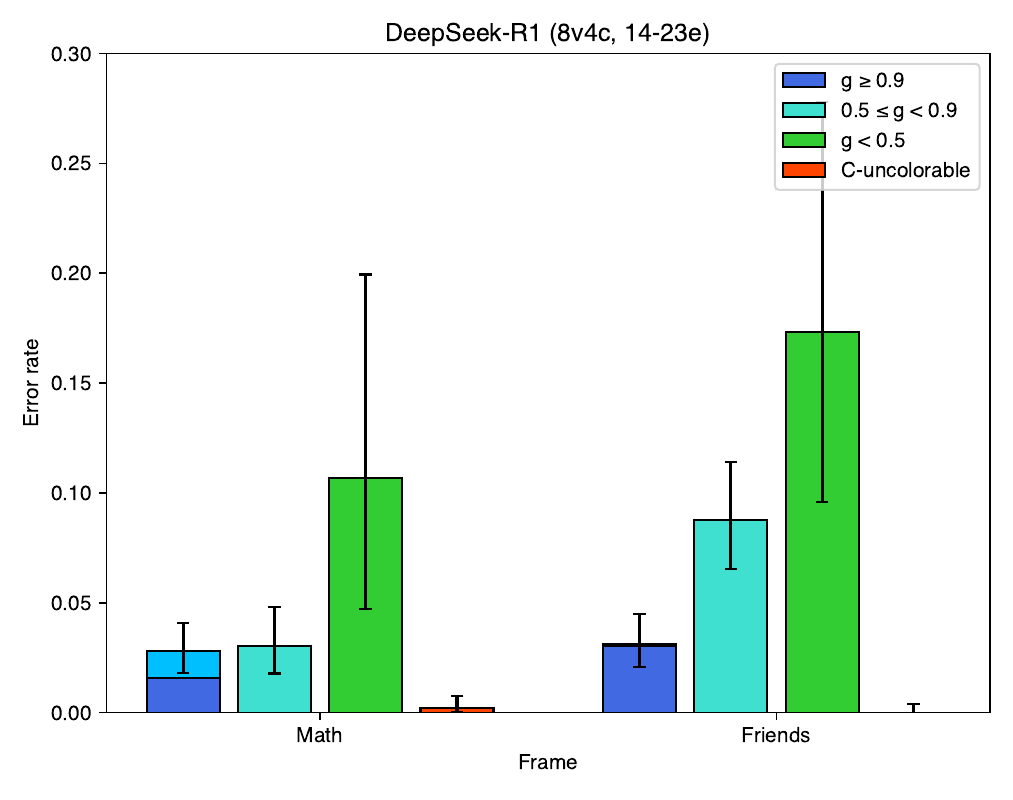}
\end{subfigure}
\end{figure}
\vspace{-0.45in}

\section{Accuracy by Model, Problem Set, Frame, and Edge Count (LRMs)} \label{ec_accuracy_lrm}

Following are line plots of accuracy (rate of correct answers; higher is better) for each combination of LRM and edge-selected problem set, split further by frame and number of edges. Error bars represent $95\%$ Clopper-Pearson binomial confidence intervals. Also shown are the proportions of each colorable problem type at each edge count.

\begin{figure}[H]
\begin{subfigure}
\centering
\includegraphics[width=0.5\textwidth]{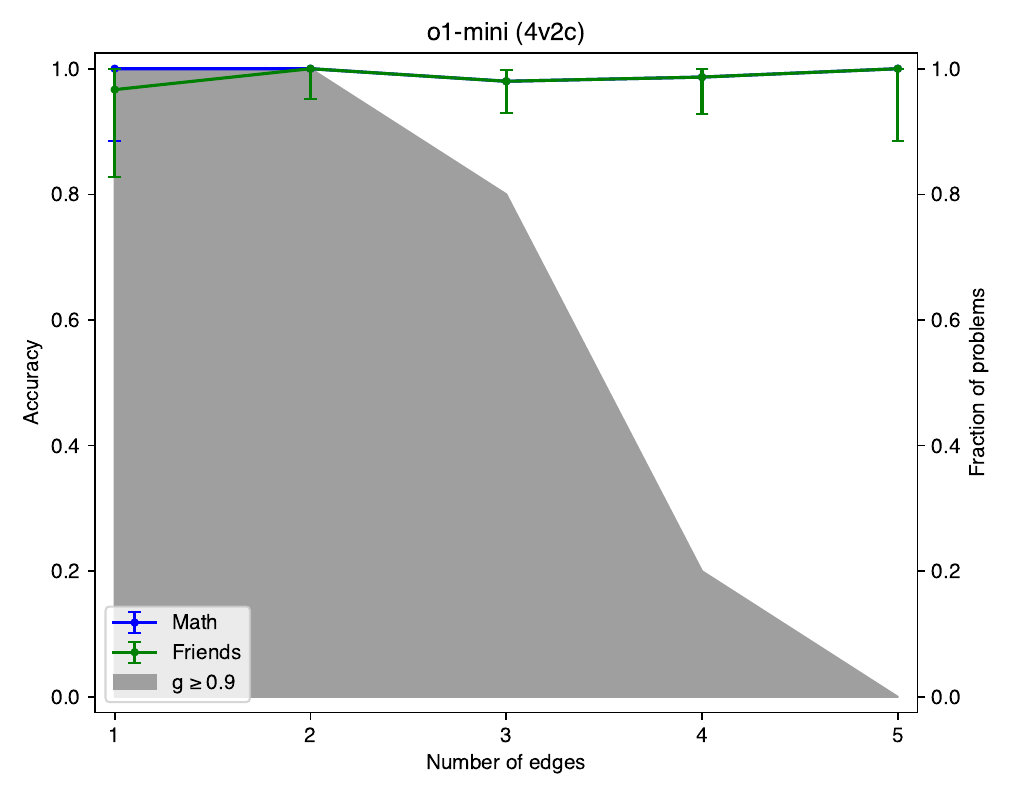}
\end{subfigure}
\hspace*{-0.9em}
\begin{subfigure}
\centering
\includegraphics[width=0.5\textwidth]{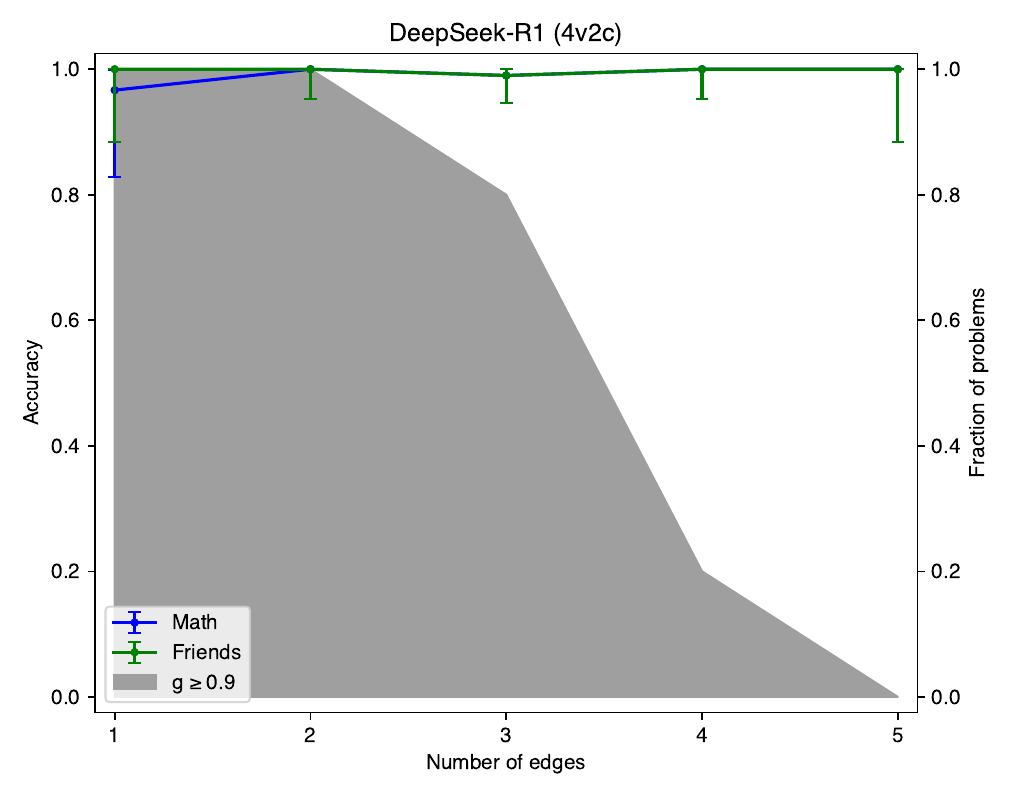}
\end{subfigure}
\end{figure}
\vspace{-0.45in}

\begin{figure}[H]
\begin{subfigure}
\centering
\includegraphics[width=0.5\textwidth]{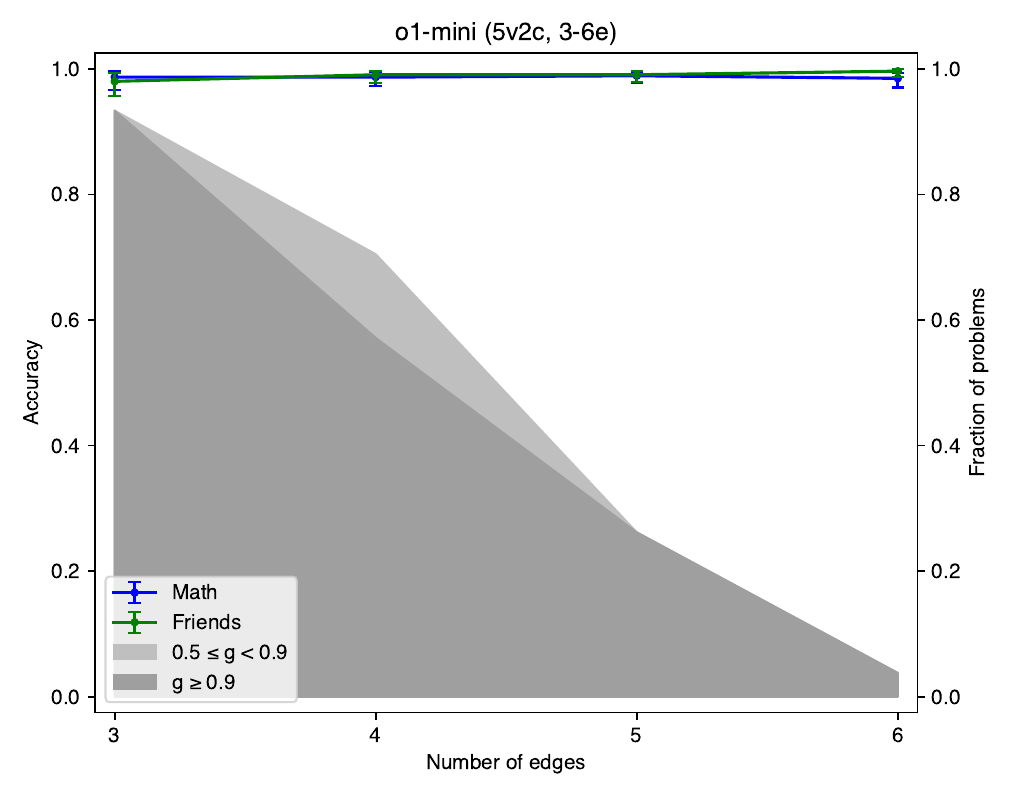}
\end{subfigure}
\hspace*{-0.9em}
\begin{subfigure}
\centering
\includegraphics[width=0.5\textwidth]{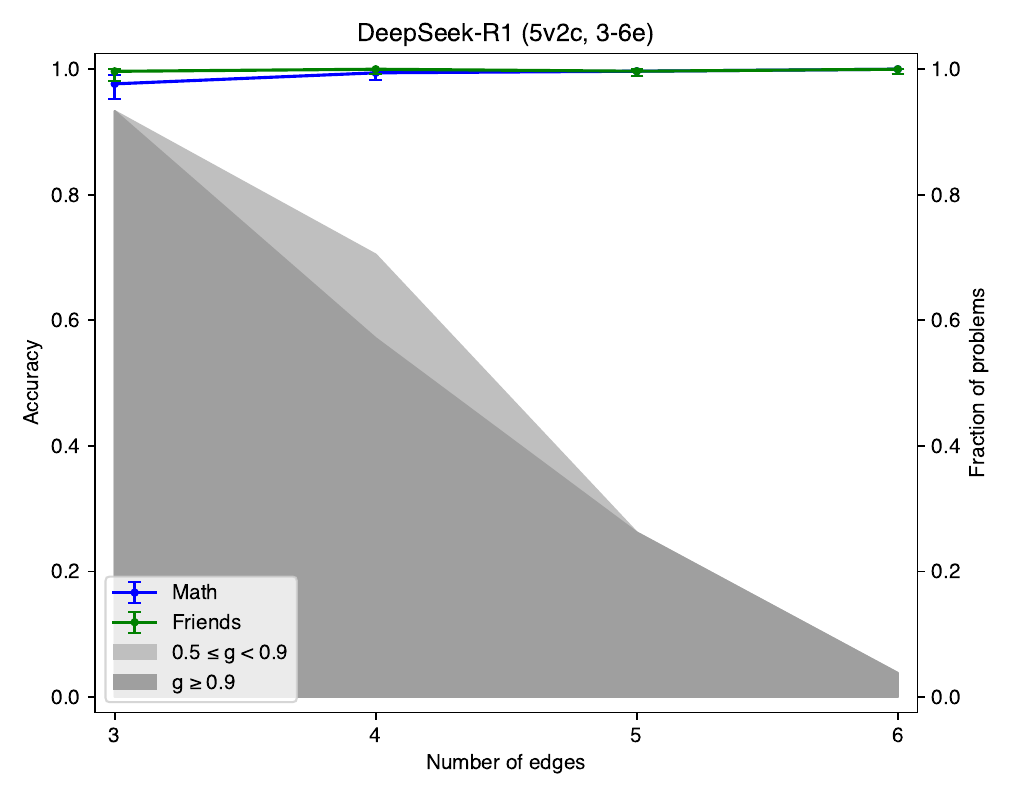}
\end{subfigure}
\end{figure}
\vspace{-0.45in}

\begin{figure}[H]
\begin{subfigure}
\centering
\includegraphics[width=0.5\textwidth]{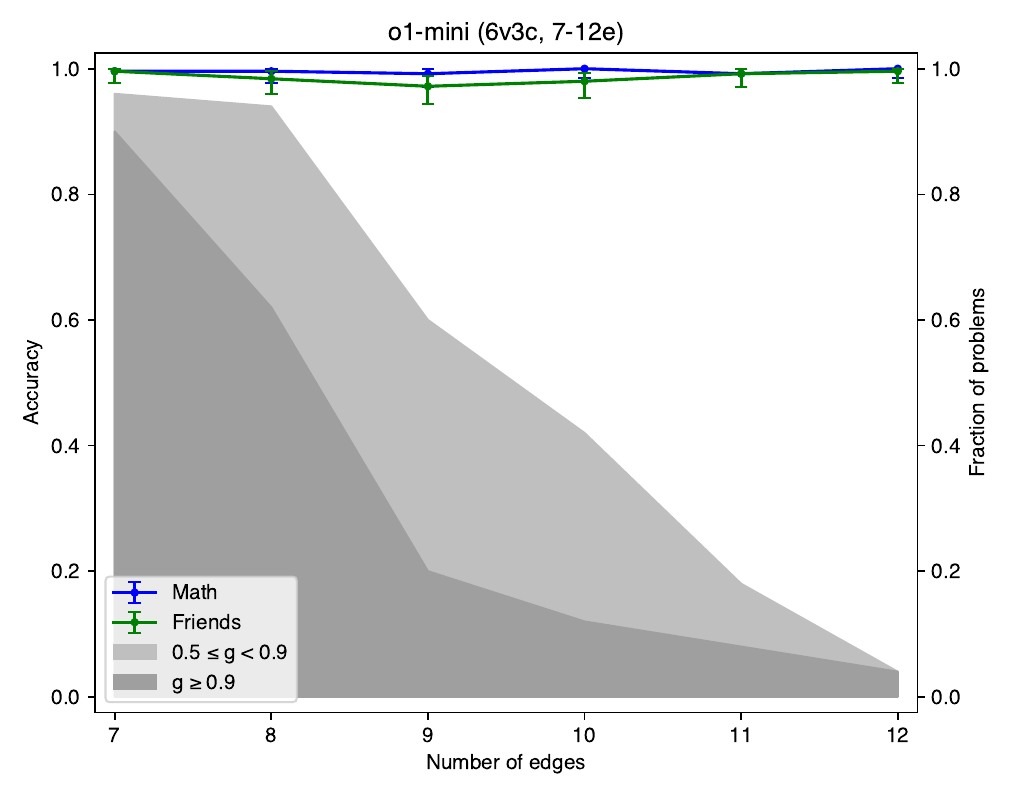}
\end{subfigure}
\hspace*{-0.9em}
\begin{subfigure}
\centering
\includegraphics[width=0.5\textwidth]{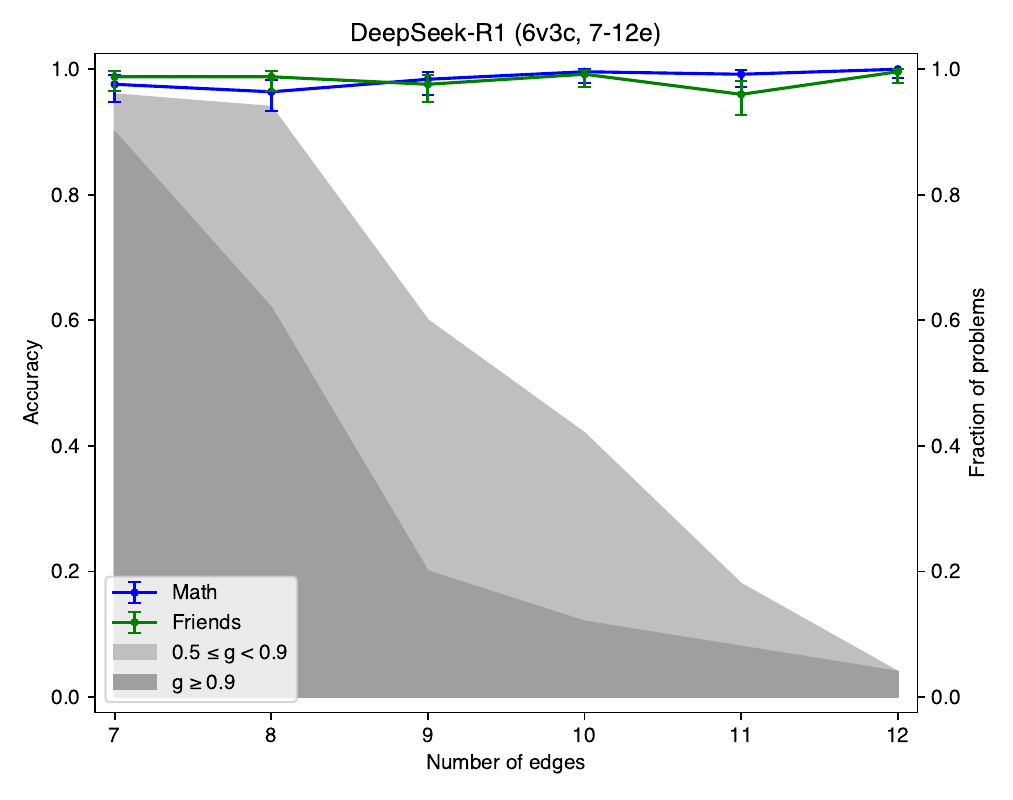}
\end{subfigure}
\end{figure}
\vspace{-0.45in}

\begin{figure}[H]
\begin{subfigure}
\centering
\includegraphics[width=0.5\textwidth]{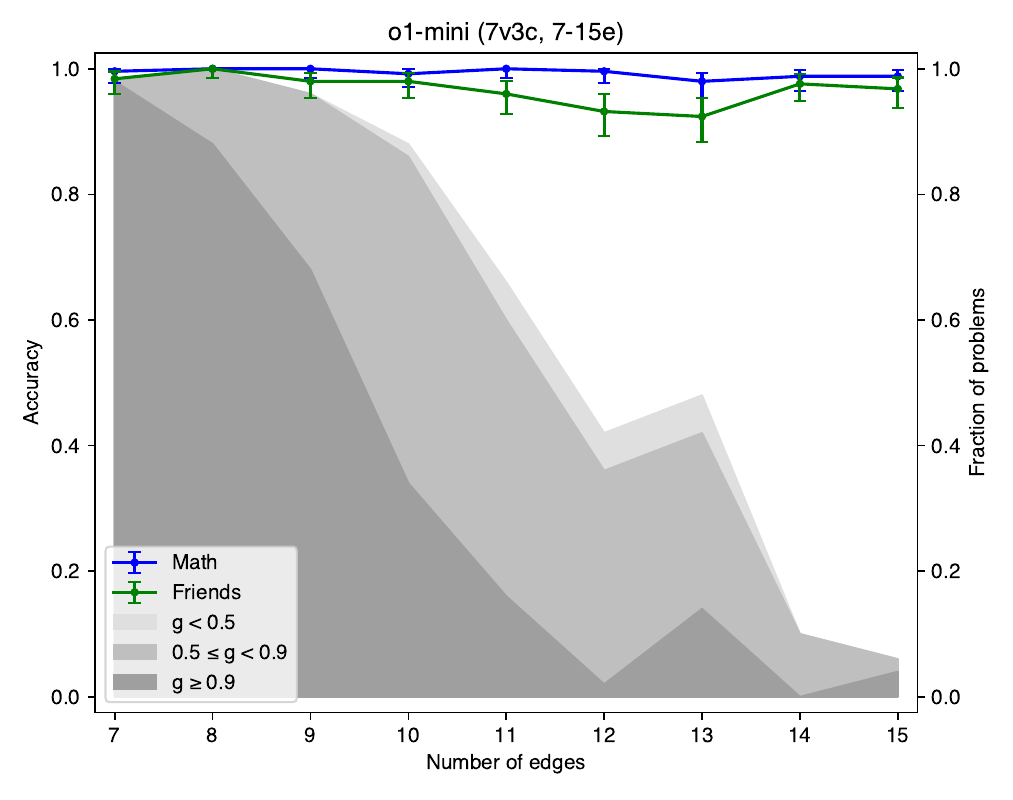}
\end{subfigure}
\hspace*{-0.9em}
\begin{subfigure}
\centering
\includegraphics[width=0.5\textwidth]{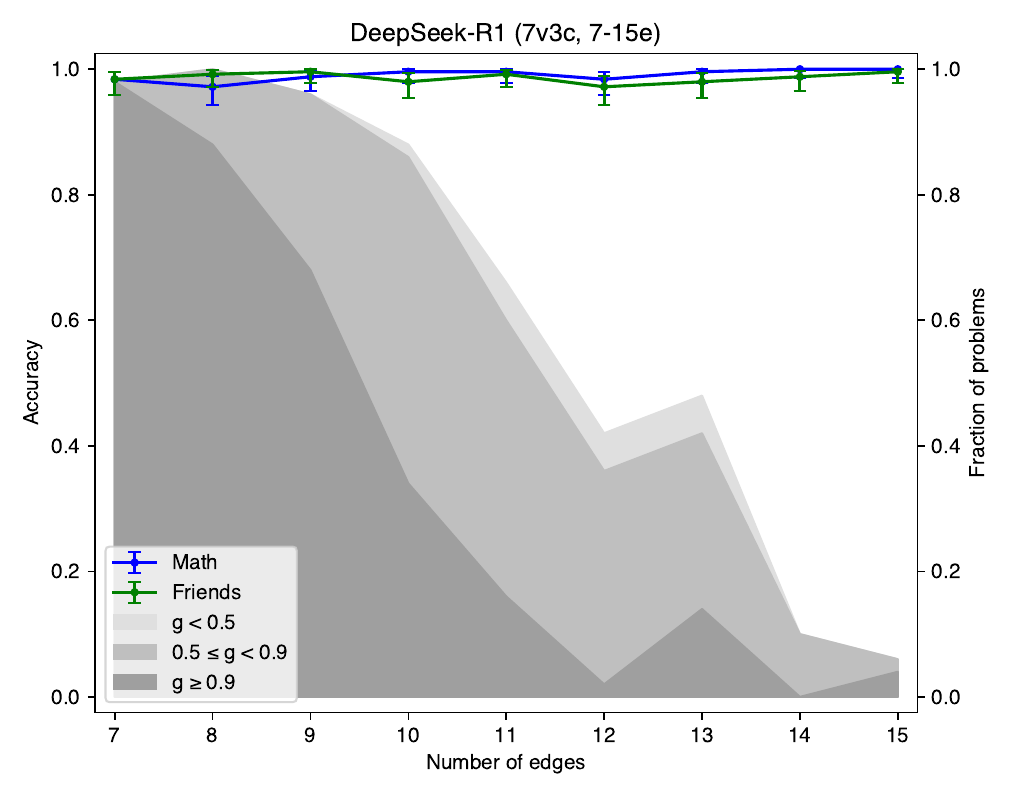}
\end{subfigure}
\end{figure}
\vspace{-0.45in}

\begin{figure}[H]
\begin{subfigure}
\centering
\includegraphics[width=0.5\textwidth]{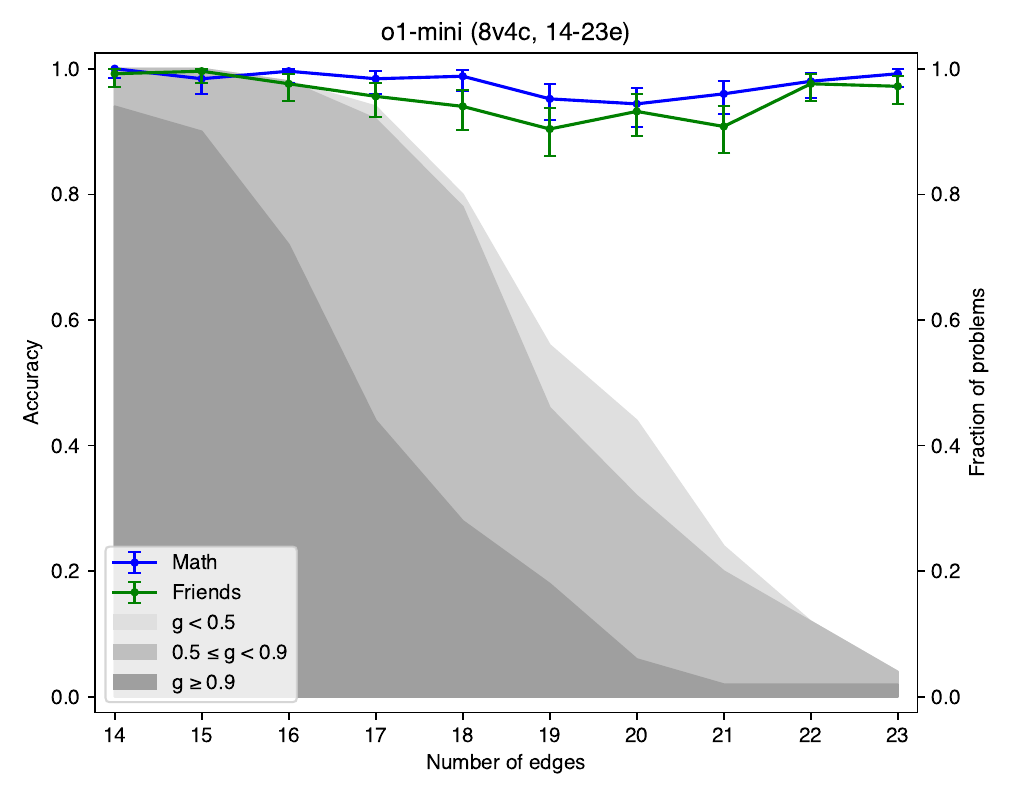}
\end{subfigure}
\hspace*{-0.9em}
\begin{subfigure}
\centering
\includegraphics[width=0.5\textwidth]{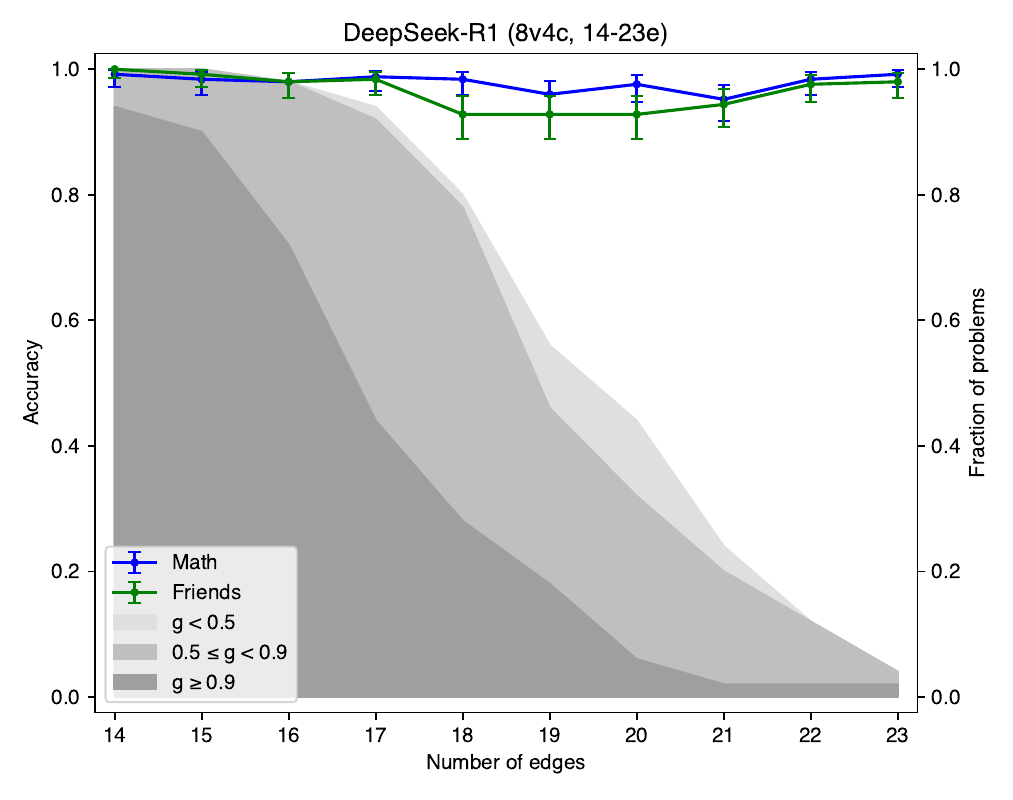}
\end{subfigure}
\end{figure}
\vspace{-0.45in}

\section{Standard LLM vs. LRM Error Rates on LRM-Selected Problems} \label{comparison}

Following are bar plots of error rates on each of the edge-selected problem sets created to evaluate o1-mini and DeepSeek-R1, showing results for those models alongside the standard LLMs split by problem type and LRM-tested frame. The lighter-colored segment of each bar is the space between the errors of the two frames. These plots directly compare all six tested models on identical sets of inputs, particularly highlighting the improvement of the LRMs over the standard LLMs.

\begin{figure}[H]
\begin{subfigure}
\centering
\includegraphics[width=0.5\textwidth]{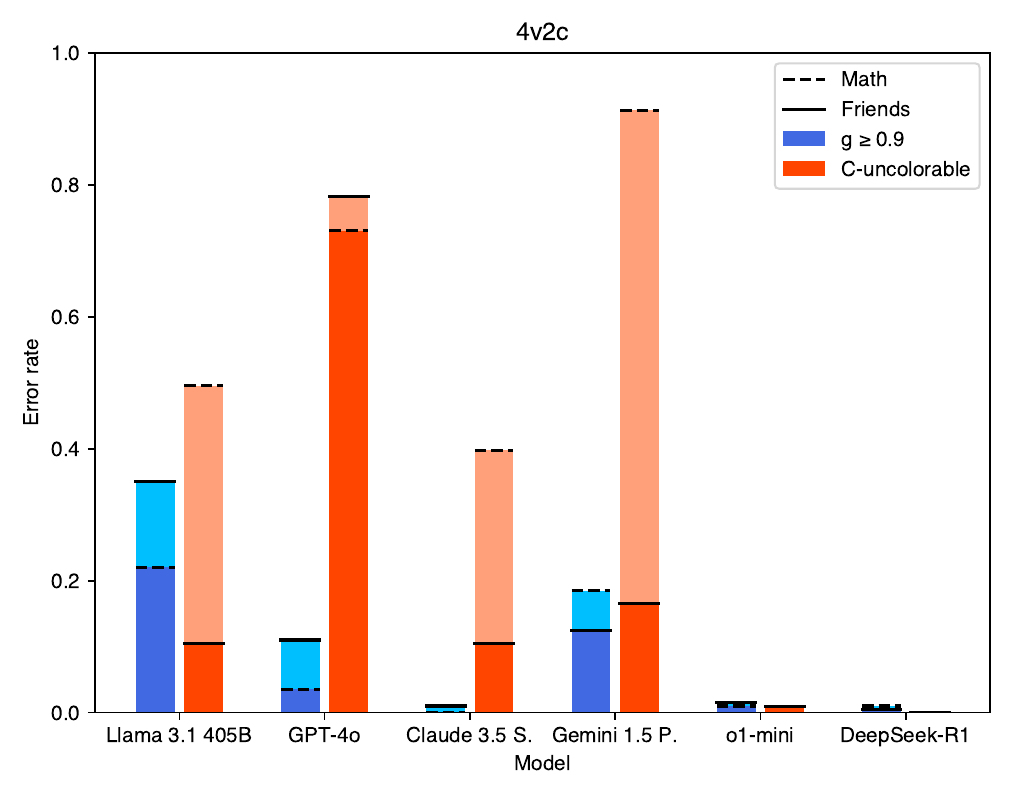}
\end{subfigure}
\hspace*{-0.9em}
\begin{subfigure}
\centering
\includegraphics[width=0.5\textwidth]{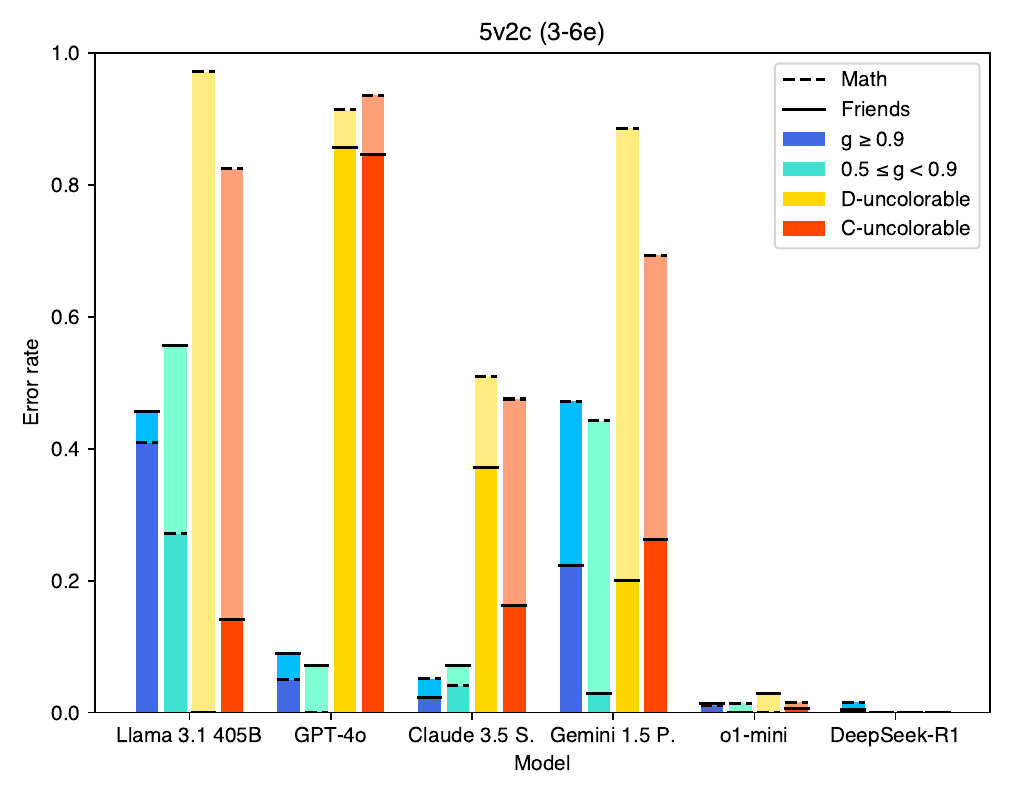}
\end{subfigure}
\end{figure}
\vspace{-0.45in}

\begin{figure}[H]
\begin{subfigure}
\centering
\includegraphics[width=0.5\textwidth]{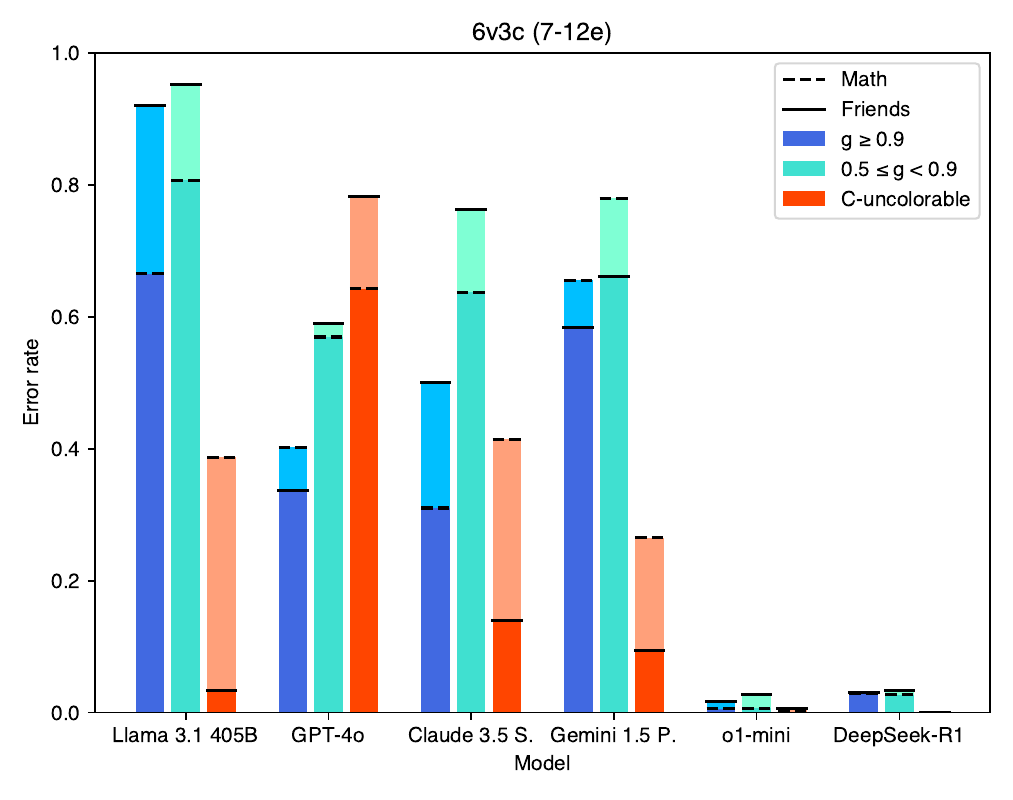}
\end{subfigure}
\hspace*{-0.9em}
\begin{subfigure}
\centering
\includegraphics[width=0.5\textwidth]{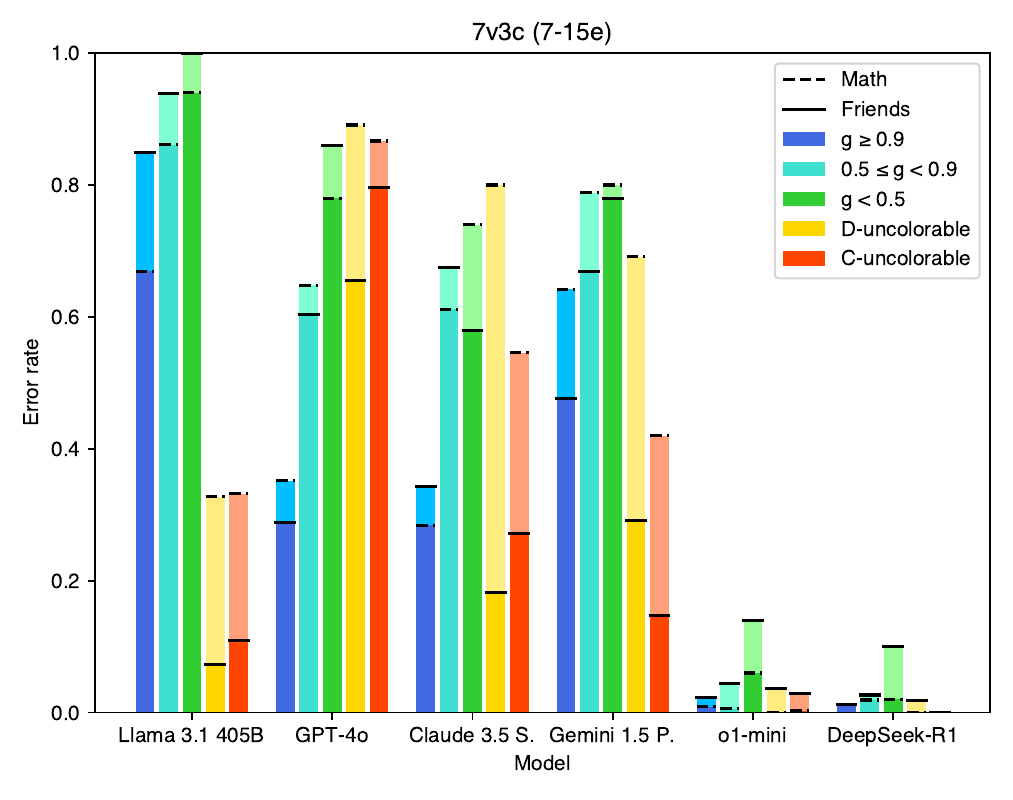}
\end{subfigure}
\end{figure}
\vspace{-0.45in}

\begin{figure}[H]
\centering
\begin{subfigure}
\centering
\includegraphics[width=0.5\textwidth]{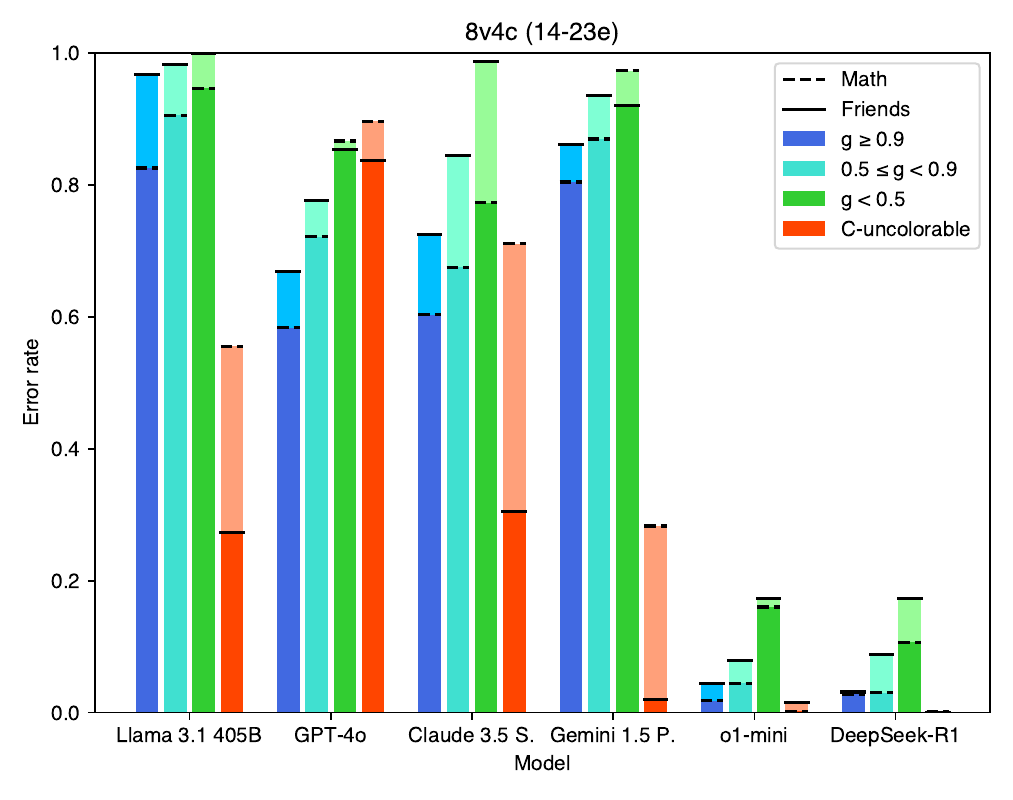}
\end{subfigure}
\end{figure}
\vspace{-0.45in}

\end{document}